\documentclass{article}

\usepackage[margin=1in]{geometry}
\usepackage[T1]{fontenc}
\usepackage{lmodern}
\usepackage{microtype}
\microtypesetup{expansion=false}
\usepackage{mathtools,amssymb,amsthm}
\usepackage{bm,dsfont,mleftright}
\usepackage{nicefrac}

\usepackage[dvipsnames]{xcolor}
\usepackage[normalem]{ulem}

\usepackage{graphicx}
\usepackage{subcaption}
\usepackage{wrapfig}
\usepackage{placeins}
\usepackage{tikz}
\usetikzlibrary{arrows.meta,calc}
\usepackage{float}

\usepackage{booktabs,diagbox,nicematrix}

\usepackage{enumitem}
\usepackage{csquotes}
\usepackage{pifont}

\usepackage[authoryear,round]{natbib}
\usepackage{hyperref}
\hypersetup{
    colorlinks=true,
    linkcolor=blue,
    citecolor=blue,
    urlcolor=blue
}
\usepackage[capitalize,noabbrev]{cleveref}

\definecolor{gold(metallic)}{rgb}{0.83,0.69,0.22}
\definecolor{AdamRed}{HTML}{D55E00}
\definecolor{Ink}{HTML}{1F1F1F}
\definecolor{ReachGray}{HTML}{767676}
\definecolor{FrameGray}{HTML}{D9DDE2}
\definecolor{CanvasLabel}{HTML}{62676D}

\theoremstyle{plain}
\newtheorem{theorem}{Theorem}[section]
\newtheorem{proposition}[theorem]{Proposition}
\newtheorem{lemma}[theorem]{Lemma}
\newtheorem{corollary}[theorem]{Corollary}

\theoremstyle{definition}
\newtheorem{definition}[theorem]{Definition}

\theoremstyle{remark}

\newcommand{\innerprod}[2]{\left\langle #1,#2 \right\rangle}

\newcommand{\angles}[2]{\left\langle #1,#2 \right\rangle}

\newcommand{\norm}[1]{\left \lVert #1 \right \rVert}
\newcommand{\abs}[1]{\left | #1 \right |}
\newcommand{\sign}[1]{\mathrm{sign}\left ( #1 \right )}

\newcommand{\ema}[2]{\mathsf{EMA}_{#2}\left[ #1 \right]}
\newcommand{\emahat}[2]{\hat{\mathsf{EMA}}_{#2}\left[ #1 \right]}

\newcommand{\adam}[3]{\mathsf{A}_{#2, #3} \left[ #1 \right]}

\newcommand{\qmin}{q_{\mathrm{min}}}
\newcommand{\bigO}[2][]{\mathcal{O}#1 \left ( #2 \right )}
\newcommand{\smallo}[1]{\mathrm{o} \left ( #1 \right )}

\newcommand{\toinfty}[1]{\overset{#1 \to \infty}{\longrightarrow}}

\newcommand{\bbR}{\mathbb{R}}
\newcommand{\bbN}{\mathbb{N}}

\newcommand{\Lcal}{\mathcal{L}}

\newcommand{\w}{\mathbf{w}}
\newcommand{\x}{\mathbf{x}}
\newcommand{\y}{\mathbf{y}}

\newcommand{\g}{\mathbf{g}}
\newcommand{\bfb}{\mathbf{b}}
\newcommand{\h}{\mathbf{h}}
\newcommand{\m}{\mathbf{m}}

\newcommand{\vb}{\mathbf{v}}
\newcommand{\ub}{\mathbf{u}}
\newcommand{\kb}{\mathbf{k}}

\newcommand{\zero}{{\mathbf{0}}}

\newcommand{\thetab}{\bm{\theta}}
\newcommand{\dtheta}[1]{\frac{d\thetab_{#1}}{d#1}}
\newcommand{\thetadir}[1]{\frac{\thetab_{#1}}{\norm{\thetab_{#1}}}}

\newcommand{\e}{\mathrm{e}}

\renewcommand{\tilde}{\widetilde}

\makeatletter
\newcommand{\litem}[2]{\item[#1]\protected@edef\@currentlabel{#1}\label{#2}}
\makeatother

\title{Watch Your Step: Adam's Margin Maximization Bias \\ 
Depends on the Learning Rate Schedule}

\author{Eitan Gronich, Gal Vardi\\
  Weizmann Institute of Science\\
  \texttt{\{eitan.gronich, gal.vardi\}@weizmann.ac.il}
}

\date{}
\begin{document}

\maketitle

\begin{abstract}
We study the implicit bias of Adam and its similarity to sign gradient descent 
on homogeneous models in binary classification tasks.
Perhaps surprisingly, we uncover that these depend on the learning rate schedule.
While previous works studied strongly decaying learning rate schedules $\eta(t)=\smallo{{t^{1/L-1}}}$, where $L$ is the homogeneity degree, and showed $\ell_\infty$-margin-maximization biases, we focus on the more common complementary regime, including constant and weakly decaying schedules. 
For a polynomial schedule $t^{\alpha}$ and homogeneity degree $L$, we show that when $\alpha > \frac 1L - \frac12$, Adam has no trajectory-independent 
margin-maximization bias for deep linear networks (in the KKT sense):
first, there is no non-trivial, locally Lipschitz, convex implicit regularizer in this setting, including any squared norm. Second, on width-$1$ networks, the result holds without requiring convexity, implying that any implicit regularizer for wide networks must be trivial when restricted to a width-$1$ network.
Nevertheless, we positively characterize the implicit bias in this regime as maximizing a \textit{trajectory-dependent} weighted-$\ell_2$ margin for any $L$-homogeneous model. The fundamental mechanism behind our results is  convergence of Adam's preconditioner to a trajectory-dependent vector, which also allows showing that the limiting parameter direction depends on initialization. In the intermediate regime $\frac 1L - 1 \leq \alpha \leq \frac1L - \frac12$ we prove an $\ell_\infty$ bias, conditional on gradient directional convergence to a full-support vector. Of independent interest, we establish a loss decay rate for full batch Adam on general homogeneous models, whose leading order depends only on the second-moment parameter.  
\end{abstract}

\section{Introduction}
The \emph{implicit bias} or \emph{implicit regularization} of optimizers refers to their tendency to favor solutions with certain properties, often associated with generalization, even when no \emph{explicit} regularization is used in training. On classification tasks and homogeneous models (e.g. deep ReLU networks), optimizers such as gradient descent and the steepest descent family have been shown to possess \emph{margin-maximization} implicit biases, selecting among all perfect-accuracy solutions those which maximize a certain confidence margin \citep{lyu_li_homogeneous,tsilivis2025}.

Adam \citep{kingma2014adam} and its derivatives are popular
optimizers for training large language models (LLMs) and other deep neural networks. An improved understanding of their implicit biases is therefore 
essential to strengthening the theoretical foundations of modern AI.
Earlier works \citep{zhang2024_adam_linear, fan2025implicit, gronich2026implicitbiasadammuon} on linear and smooth homogeneous models trained with exponentially-tailed losses showed that Adam has a margin-maximization bias towards the $\ell_\infty$ margin when the learning rate schedule $\eta(t)$ undergoes a strong decay $\eta(t)=\smallo{t^{1/L-1}}$. Here, $L$ is the homogeneity degree, equal to the network depth for linear or ReLU networks. 
Moreover, these works show similarity between late-phase Adam
and \emph{sign gradient descent (sign-GD)}, which also has a bias towards $\ell_\infty$-margin maximization \citep{tsilivis2025}. 
More broadly, the similarity between Adam and sign-GD (including variations with momentum) is assumed, derived or empirically tested in many prior works on optimization geometry,
generalization, and optimizer design
\citep{balles2020geometry,zhuang2020adabelief,liu2024sophia,kunstner2024heavy,kunstner2025scalinglawsgradientdescent,li2025optimization,bernstein2024old}.

The schedule $\eta(t)$ plays a key role, since it changes the effective memory horizon along the trajectory when the momentum hyperparameters $\beta_1,\beta_2$ remain fixed \citep{belotto2020general}, as is the case in practice. Motivated by the common constant and weakly decaying schedules, we extend the characterization of \citet{zhang2024_adam_linear, gronich2026implicitbiasadammuon} to $\eta(t) \asymp t^\alpha$ with $\alpha \geq \frac1L-1$, obtaining the following surprising results:
\begin{enumerate}
    \item When $\alpha > \frac 1L - \frac 12$, including a constant learning rate for $L > 2$,
    we prove nonexistence results for 
    implicit regularizers of Adam. We formalize this by studying 
    regularizers 
    inspired by squared norms: locally Lipschitz functions $R(\thetab)$, where $\thetab$ are the network's parameters, satisfying for all separable datasets $\{(\x_i,y_i)\}$ and Adam trajectories that directional limit points of $\thetab$ are proportional to KKT points of 
    \[
    \min_{\thetab} R(\thetab) \quad \text{s.t.} \quad \forall i : y_if(\x_i;\thetab) \geq 1 \ .
    \]
    \begin{enumerate}
        \item We first show that no non-constant $R(\thetab)$ satisfies this definition for width-$1$ linear networks. Thus, Adam does not maximize any margin in width-$1$ linear networks. 
        
        \item We then lift the construction to a deep linear network with general widths, 
        proving triviality of any \emph{convex} regularizer.
    \end{enumerate}
    Both results rule out squared norms as implicit regularizers of Adam, implying there is no margin-maximization bias in the usual sense. In particular, Adam's late-stage behavior differs significantly from that of sign-GD.
    \item Despite the above results implying that no single $R(\thetab)$ regularizes \emph{all} trajectories, we show on the flip side that Adam has a \emph{trajectory-dependent} margin-maximization implicit bias. Specifically, directional limit points of Adam are KKT points of a weighted-$\ell_2$ margin-maximization objective defined by a trajectory-dependent preconditioning diagonal matrix. 
    This positive result, as well the negative ones, rely on a key mechanism of convergence of Adam's second-moment preconditioner to a trajectory-dependent limit.
    \item In the intermediate range $\frac 1L - 1 \leq \alpha \leq \frac 1L - \frac 12$, we extend the 
    result of \citet{gronich2026implicitbiasadammuon}, and prove that  Adam on homogeneous models maximizes the $\ell_\infty$ margin. Our result is conditional on gradient directional convergence to a full-support vector.
    \item Of independent interest, we establish a loss decay rate for (full batch) Adam on general homogeneous models, whose leading order depends only on the second-moment parameter:
    \begin{align}\label{eq:introduction_loss_rate}
        - \log \Lcal &= \frac{c_2}2t \pm \smallo{t}, \qquad & (\alpha > \frac 1L - 1), \\
        - \log \Lcal &= \frac{c_2}2t \pm \bigO{\log t}, \qquad & (\alpha > \frac 1L - \frac12),
    \end{align}
    where $c_2$ is the continuous analogue of $- \log \beta_2$ (see Section~\ref{sec:problem_setting} for definitions). This both shows a direct connection between the value of Adam's $\beta_2$ parameter and the loss decay rate, and demonstrates that this asymptotic rate is 
    essentially 
    independent of the learning rate schedule.
    In contrast, earlier results for the regime $\alpha < \frac{1}{L}-1$ give the schedule-dependent rate $-\log \Lcal \asymp t^{L(\alpha+1)} = \smallo{t}$~\citep{zhang2024_adam_linear,gronich2026implicitbiasadammuon};
    our analysis also derives this rate, under weaker assumptions.
\end{enumerate}
\section{Related work}
An $\ell_2$-margin-maximization implicit bias of gradient descent was first proved on linear predictors by \citet{soudry2017_gd_linear}, and subsequent work investigated how architecture and initialization shape the implicit bias of certain deep networks \citep{ji2018gradient,gunasekar2018implicit,yun2020unifying,moroshko2020implicit}. General homogeneous models were studied by \citet{lyu_li_homogeneous}, who showed that any directional limit point of the trajectory is the direction of a KKT point of the $\ell_2$-max-margin problem, while \citet{ji2020directional} proved for homogeneous models definable in an o-minimal structure that the parameter direction converges under gradient descent. See \citet{vardi2023implicit} for a survey of the implicit bias of gradient descent.

A common family of non-Euclidean optimizers is \textit{steepest descent}, whose special cases include sign gradient descent and spectral descent (with momentum: Signum \citep{bernstein2018signsgd} and idealized Muon \citep{jordan2024muon}, respectively). \citet{gunasekar2018characterizing} proved a margin-maximization bias for steepest descent on linear models, where the margin geometry is not $\ell_2$ but the norm defining the optimizer. This was generalized by \citet{tsilivis2025} to homogeneous models, showing that any limit point of $\thetadir{t}$ is the direction of a KKT point of the appropriate max-margin problem. Specifically, this result implies that sign-GD has a bias towards $\ell_\infty$-margin maximization.

On Adam, \citet{wang2021_adam_eps_homoegeneous} and \citet{wang2022momentum} first established a bias toward $\ell_2$-margin maximization. Their analysis considers a regime in which Adam's stability constant $\varepsilon$ asymptotically dominates the denominator, leading to gradient-descent-like behavior. Analyses omitting the stability constant include the works of \citet{zhang2024_adam_linear} and \citet{fan2025implicit} establishing $\ell_\infty$-margin maximization on linear models (in the binary and multiclass settings, respectively), with a decaying learning rate, 
while \citet{baek2026persample} showed that deterministic batching can induce an implicit bias different from that of full-batch Adam. Recently, \citet{gronich2026implicitbiasadammuon} established an $\ell_\infty$ bias for Adam on smooth, homogeneous networks under a learning-rate schedule satisfying $\eta(t) = \smallo{t^{\frac{1}{L}-1}}$, where $L$ denotes the degree of homogeneity. 

For AdamW, which incorporates explicit weight decay, \citet{xie2024implicit} studied smooth models and losses and showed that trajectory limit points satisfy the KKT conditions for loss minimization subject to a bound on the $\ell_\infty$ norm of the parameters.

Impossibility results for implicit bias have been proved before by \citet{vardi2021implicit} for squared-loss regression with gradient flow on ReLU networks, and by \citet{dauber2020can,suggala2018connecting} for other learning problems. \citet{qian2019implicit} and \citet{wang2021_adam_eps_homoegeneous} observed a converging, trajectory-dependent preconditioner for AdaGrad in binary classification (on linear and homogeneous models, respectively), 
and \citet{bai2026understandingadamconvergencehighly} noticed the  mechanism of second-moment convergence (which they term \enquote{decoupling}) in Adam on highly degenerate polynomials.

\section{Problem setting}\label{sec:problem_setting}
Throughout this work, we consider a binary classification task defined by a dataset $\{(\x_i, y_i)\}_{i=1}^m \subseteq \bbR^d \times \{-1,1\}$. We study models $f(\x; \thetab)$, parameterized by $\thetab \in \bbR^p$, which satisfy the following assumptions: 
\begin{enumerate}[label=(M\arabic*), align=left]
    \item $f$ is locally Lipschitz in $\thetab$ and admits a chain rule (see Appendix~\ref{sec:clarke_loss}).\label{model_ass:1_locally_lipschitz}
    \item $f$ is $L$-homogeneous for $L \geq 1$: $
    \forall z>0, \x \in \bbR^d, \thetab \in \bbR^p:~~f(\x; z\thetab) = z^Lf(\x; \thetab)$.\label{model_ass:2_homogeneous}
\end{enumerate}
For example, deep linear networks, ReLU networks and convolutional networks (all without bias terms) of depth $L$ satisfy these conditions. The classification \textit{margin} of a homogeneous model, with respect to a norm $\norm{\cdot}$, is defined as
\begin{equation}\label{eq:margin}
    \gamma(\thetab) = \min_{i \in [m]}y_if\left(\x_i;\frac{\thetab}{\norm{\thetab}}\right)\ .
\end{equation}
We write $q_i(\thetab) = y_if(\x_i;\thetab)$ and $\qmin(\thetab) = \min_i q_i(\thetab)$ as shorthands.

We analyze Adam, where we follow \citet{zhang2024_adam_linear,fan2025implicit,gronich2026implicitbiasadammuon, baek2026persample}, among others, by setting the stability constant $\varepsilon$ to $0$. This choice is motivated by the fact that in asymptotic analysis, a positive $\varepsilon$ dominates the decaying denominator $\sqrt{\hat \vb_t}$, while in practice $\varepsilon$ typically remains negligible throughout optimization. 
We also follow \citet{barakat2021convergence, ma2022qualitative, belotto2020general, wang2021_adam_eps_homoegeneous, gronich2026implicitbiasadammuon} in working with a continuous-time version of Adam, defined as follows. Consider an empirical loss $\Lcal(\thetab)$ and a choice of subgradients $\g_t \in \partial \Lcal(\thetab_t)$, then:
\begin{gather}
    \frac{d\thetab_t}{dt}
    = -\eta(t)\frac{\hat \m_t}{\sqrt{\hat \vb_t}},
    \label{eq:adam} \\
    \m_t = \ema{\g_t}{c_1}, \quad
    \hat \m_t = \emahat{\g_t}{c_1}, \quad
    \vb_t = \ema{\g_t^2}{c_2}, \quad
    \hat \vb_t = \emahat{\g_t^2}{c_2}\ .
\end{gather}
Here we denote by $\ema{g}{c}, \emahat{g}{c}$ (respectively) the plain and bias-corrected exponential moving average (momentum) term for $g$ with a continuous smoothing parameter $c > 0$ analogous to the discrete parameter $\beta \in [0, 1)$ (or more precisely, analogous to $- \log \beta$; see Appendix~\ref{sec:ema} for definitions and references). Note that division, square root and the EMA operations are applied coordinate-wise.
Throughout this work, we assume $c_1 \geq c_2$, analogous to the common setting $\beta_1 \leq \beta_2$. For any $j\in [p]$ with $\hat \vb_t[j] = 0$ we set $\dtheta{t}[j] = 0$; this is consistent with the practical $\varepsilon > 0$ setting of Adam, since $\hat \vb_t[j] = 0$ implies $\hat \m_t[j] = 0$. 
We consider a polynomial learning rate schedule of the form
\begin{equation}\label{eq:lr_schedule}
    \eta(t)=\eta_0(1+t)^{\alpha}, \qquad \alpha > -1, \quad \eta_0 > 0 \ .
\end{equation}
Such learning rate schedules are ubiquitous in practice as well as theoretical analyses. 

The losses we consider, following \citet{gronich2026implicitbiasadammuon}, are \textit{log-concave, exponentially-tailed} losses. Namely, we observe losses $\Lcal(\thetab):\bbR^p \to \bbR$ of the form 
\begin{align}\label{eq:loss_main_text}
\Lcal(\thetab) &= \sum_{i=1}^m \ell(y_if(\x_i; \thetab)) ~, \qquad \ell \in \mathrm L_{\e},
\end{align}
where $\mathrm L_{\e}$ is the class of functions of the form $\ell(u) = e^{-\varphi(u)}$ for $\varphi \in C^2(\bbR)$ which is convex, strictly increasing and has bounded first and second derivatives (see Appendix~\ref{sec:clarke_loss}).
Notably, the exponential and logistic losses ($\ell(u)=e^{-u}$ and $\ell(u)=\log(1+e^{-u})$, respectively) 
are in $\mathrm L_{\e}$.

We introduce two trajectory assumptions: 
\begin{enumerate}[label=(T\arabic*), align=left]
    \item Positive-margin interpolation: $\exists t_0\geq0, \gamma_0 > 0,\forall t \geq t_0: \gamma_2(\thetab_t) := \frac{\qmin(\thetab_t)}{\norm{\thetab_t}_2^L}\geq \gamma_0$.\label{traj_ass:positive_margin_interpolation}
    \item Loss decay: $\Lcal(\thetab_t)\toinfty{t} 0$. \label{traj_ass:loss_decay}
\end{enumerate}
The lower bound $\gamma_0$ on the margin may be arbitrarily small, and the choice of norm here is inconsequential; we also note that~\ref{traj_ass:loss_decay} may be weakened to require a sufficiently small loss on a sufficiently long window (see Lemma~\ref{lemma:finite-low-loss-window}).

Our main question of interest relates to \emph{implicit margin regularizers}, which we define below:

\begin{definition}[Implicit Margin 
Regularizer for Homogeneous Models in the KKT-Stationarity Sense]\label{def:implicit_regularizer}
Let $p \geq 1$, let $f(\x; \thetab) : \bbR^{d}\times \bbR^p \to \bbR$ satisfy~\ref{model_ass:1_locally_lipschitz},~\ref{model_ass:2_homogeneous} and let $\ell(u) :\bbR\to \bbR$ be a 
loss function. We say that a locally Lipschitz $R(\thetab) : \bbR^p \to \bbR$ that admits a chain rule is an implicit margin regularizer of an optimizer $O$ on $f$ with the loss $\ell$, under a set of trajectory assumptions $\mathcal T$, if $R$ satisfies the following: 

For any binary classification dataset $\{(\x_i, y_i)\}_{i=1}^m \subseteq \bbR^d \times \{\pm 1\}$, if $\thetab_t$ is optimized with $O$ and the empirical loss $\Lcal(\thetab) = \sum_{i=1}^m \ell(y_if(\x_i;\thetab))$ on $f$, satisfies the trajectory assumptions $\mathcal T$ and \ref{traj_ass:positive_margin_interpolation},\ref{traj_ass:loss_decay}, and if $\frac{\thetab_t}{\norm{\thetab_t}_2} \toinfty{t} \bar \thetab$ 
then $\hat \thetab := \frac{\bar \thetab}{\qmin(\bar \thetab)^{1/L}}$ is a KKT point of 

\begin{equation}\label{eq:R_KKT}
\min_{\thetab} R(\thetab) \quad \text{s.t.}\quad \forall i\in[m]:~y_if(\x_i; \thetab) \geq 1 \ .   
\end{equation}    
\end{definition}
Throughout this work we often write \enquote{implicit regularizer} or just \enquote{regularizer} for short; no other types of regularizers are considered.

Requiring Assumptions~\ref{traj_ass:positive_margin_interpolation}
and~\ref{traj_ass:loss_decay}, together with directional convergence
$\frac{\thetab_t}{\norm{\thetab_t}_2} \toinfty{t} \bar\thetab$,
in Definition~\ref{def:implicit_regularizer} only strengthens our
nonexistence results: no implicit regularizer exists even when restricted to trajectories satisfying these conditions.

Nevertheless, all three conditions are natural in the study of the
implicit bias of homogeneous models. Positive-margin interpolation,
as required by Assumption~\ref{traj_ass:positive_margin_interpolation},
is a form of the realizability premise underlying late-phase
implicit-bias analysis. Some works assume momentary realizability \citep{lyu_li_homogeneous,tsilivis2025} while others impose a stronger requirement
of directional convergence to a positive-margin limit
\citep{gunasekar2018implicit,nascon2019_gradientdescent,gronich2026implicitbiasadammuon}.
Loss decay (Assumption~\ref{traj_ass:loss_decay}) is likewise standard
in max-margin analyses of separable classification, where convergence
of the training loss to zero is either assumed
\citep{gunasekar2018characterizing,nascon2019_gradientdescent,
nacson2019lexicographic} or proved
\citep{soudry2017_gd_linear,lyu_li_homogeneous, tsilivis2025, gronich2026implicitbiasadammuon}.
Directional convergence is also explicitly assumed in several works
\citep{gunasekar2018implicit,nascon2019_gradientdescent,
nacson2019lexicographic,gronich2026implicitbiasadammuon} and proved in some works on gradient descent and adaptive methods \citep{ji2020directional, wang2021_adam_eps_homoegeneous}, and provides a natural sufficient condition
for convergence of the predictor up to scale, since the parameter
direction determines a homogeneous predictor up to an overall scaling.
Indeed, since the goal in studying the implicit bias is to understand which interpolators gradient methods prefer, it is natural to assume convergence to zero loss and to a specific predictor (up to scaling), and to analyze the properties of this interpolator.

Our result for the range $\frac 1L - 1\leq \alpha \leq \frac1L - \frac12$ (\Cref{thm:intermediate}) assumes directional convergence of the
gradients to a vector with full support, which, in our setting, implies directional convergence of the parameters. Our $\alpha > \frac 1L - \frac 12$ positive result (\Cref{thm:fast_regime_kkt_main_text}) requires no directional convergence
and instead applies to every directional limit point. 

Using Definition~\ref{def:implicit_regularizer}, we ask: is there an implicit margin regularizer of Adam on homogeneous networks in the learning rate regime $\alpha > \frac 1L -1$?

\paragraph{Notation.} 
We
use standard asymptotic notations $\mathcal O, \Omega, \Theta$ as well as $\lesssim, \gtrsim, \asymp$ (with the same respective meanings), and $\mathrm{o}$. We focus on behavior as $t \to \infty$, and so asymptotic notations hide all multiplicative constants independent of $t$, unless stated otherwise. By $\log u$ we refer to the natural logarithm unless stated otherwise. Vectors are bold-faced; time-dependent vectors are written as $\w_t$, and their coordinates $\w_t[j]$ or $w_j(t)$ interchangeably. $\norm{\cdot}_p$ is the standard $\ell_p$ norm for $1 \leq p \leq \infty$. For $n \in \bbN$, we denote $[n] := \{1,\dots,n\}$. 
For a scalar $z$ we use $\sign{z}=1$ if $z>0$, and $-1$ if $z<0$. For convenience we define $\sign{0}=0$.  
For a vector $\g \in \bbR^p$ we denote by $\sign{\g}$ the vector $\sign{\g}[j] := \sign{\g [j]}$ for $j \in [p]$.
We denote the $p-1$ dimensional $\ell_2$ unit-sphere by $\mathbb S^{p-1} := \{\ub \in \bbR^p \mid \norm{\ub}_2 = 1\}$. The notation $C^p$ refers to the class of $p$-times continuously differentiable functions.

\section{Main results}
In this section we provide our main findings.
We identify three regimes, which we name \emph{slow, intermediate and fast}, referring to the potential for parameter movement as measured by $\int_0^t \eta$. Interestingly, while the slow and intermediate regimes share an $\ell_\infty$ bias under certain conditions, the fast regime differs in its implicit bias but shares a loss decay rate of $\Lcal \sim \exp(-\frac{c_2}2t \pm \smallo{t})$ with the intermediate regime (excluding its lower endpoint).

We denote for convenience the following quantity, noting its value ranges in the three regimes:
\begin{equation}
    \zeta = \alpha -\left(\frac 1L - 1\right), \qquad     \text{Slow:} \ \zeta < 0, \qquad \text{Intermediate:} \ 0 \leq \zeta \leq \frac 12, \qquad \text{Fast:} \ \zeta > \frac12 \ .
\end{equation}

\paragraph{Slow Regime.}
The first theorem is due with minor changes in assumptions 
to \citet{gronich2026implicitbiasadammuon}; we state it here for completeness.

\begin{theorem}[Adam in the slow regime admits an $\ell_\infty$-max-margin implicit regularizer; Paraphrased from \citet{gronich2026implicitbiasadammuon}]
    For any $\alpha < \frac 1L - 1$ (i.e. $\zeta < 0$), and $p \geq 1$, the 
    function
    $R(\thetab)=\frac12 \norm{\thetab}_\infty^2$ is an implicit margin regularizer for Adam on any $C^1$ model $f$ satisfying~\ref{model_ass:2_homogeneous}, and with any $\ell \in \mathrm L_{\e}$ (\Cref{eq:loss_main_text}). Moreover, when $\frac{\thetab_t}{\norm{\thetab_t}}$ converges and under \ref{traj_ass:positive_margin_interpolation},\ref{traj_ass:loss_decay}, it holds for any $j \in [p]$ with $\abs{\g_t[j]} \geq \Omega(\norm{\g_t}_1)$ that
    \begin{equation}
    \frac{\abs{\hat \m_t[j]}}{\sqrt{\hat \vb_t[j]}} \toinfty{t} 1  \ .
    \end{equation}
\end{theorem}
\paragraph{Fast Regime.}
In stark contrast, in the fast regime 
(which notably includes a constant learning rate for any $L > 2$), we prove that late-phase behavior is history-dependent in a way that precludes a universal implicit regularizer. An explicit construction shows that there is no implicit regularizer even for the simple width-$1$ linear network with input dimension $2$:
\begin{equation}\label{eq:construction_model_main_text}
    \forall \thetab = (\ub, \bfb) = (u_1,u_2,b_1,...,b_{L-1}) \in \bbR^{L+1}: \qquad f(\x; \thetab) := P(\bfb)\innerprod{\ub}{\x}, \qquad P(\bfb):= \prod_{j=1}^{L-1}b_j  \ .
\end{equation}

\begin{theorem}[Adam in the fast regime admits no regularizer for the width-$1$ model.]\label{thm:no_regularizer_narrow_main_text}
    Let $p \geq 2$ and let $f_p$ be the width-$1$ model of depth $L$ with input dimension $d=2$, as in~\eqref{eq:construction_model_main_text} (here $L=p-1$). When $\alpha > \frac 1L - \frac 12$ (i.e. $\zeta > \frac 12$), any implicit margin regularizer (\Cref{def:implicit_regularizer}) $R$ of Adam on $f_p$ with the exponential loss is constant. 
    
\end{theorem}
 The proof of Theorem~\ref{thm:no_regularizer_narrow_main_text} shows that $\bar \thetab = \lim_{t \to \infty}\frac{\thetab_t}{\norm{\thetab_t}_2}$ varies continuously on fixed datasets by varying initialization only. Indeed, there exist open domains $U_1,\dots,U_{2^{L+2}} \subseteq \mathbb S^{p-1}$, whose union covers $\mathbb S^{p-1}$ up to a null set, so that for each $U_i$ and any compact subset $A \subseteq U_i$, there exists a two-point dataset $S_A$ with $\frac{\thetab_t}{\norm{\thetab_t}_2}$ converging to any point in $A$ with choice of initialization (see Corollary~\ref{cor:compact_fixed_dataset_reachability} for details and Appendix~\ref{sec:experiment_details} for empirical validation).

While already a surprising result, Theorem~\ref{thm:no_regularizer_narrow_main_text} does not rule out an implicit regularizer for wide linear networks, if it is trivial when restricted to the width-$1$ model. Hence we extend the result of Theorem~\ref{thm:no_regularizer_narrow_main_text} to linear networks of arbitrary widths and input dimensions, showing that no \textit{convex} regularizer exists.

\begin{theorem}[Adam in the fast regime admits no convex regularizer for arbitrary-width deep linear networks]\label{thm:no_regularizer_wide_main_text}
    Let $p \geq 2$ and let $f_p$ be a scalar-output deep linear network with input dimension $d \geq 2$, $p$ parameters, arbitrary depth $L$ and arbitrary hidden widths. When $\alpha > \frac 1L - \frac 12$ (i.e. $\zeta > \frac 12)$, any convex implicit margin regularizer (\Cref{def:implicit_regularizer}) $R$ of Adam on $f_p$ with the exponential loss is constant.
\end{theorem}
In particular, Theorems~\ref{thm:no_regularizer_narrow_main_text},~\ref{thm:no_regularizer_wide_main_text} show that no squared norm is an implicit regularizer in the fast regime, hence there is no max-margin bias in the usual sense.

Theorems~\ref{thm:no_regularizer_narrow_main_text},~\ref{thm:no_regularizer_wide_main_text} follow from a mechanism of convergence of the normalized second moments $e^{c_2t}\vb_t[j]$ to trajectory-dependent limits $\omega_j$ in the fast regime. Using this same mechanism, one can positively characterize the implicit bias of Adam in the fast regime as maximizing a margin whose geometry is controlled by the limits $\omega_j$. The following theorem holds for a general homogeneous model, not necessarily a linear network.

\begin{theorem}[Adam in the fast regime maximizes a trajectory-dependent margin.] \label{thm:fast_regime_kkt_main_text} 
Let $f$ satisfy~\ref{model_ass:1_locally_lipschitz},~\ref{model_ass:2_homogeneous}, and let $\ell \in \mathrm L_{\e}$ (\Cref{eq:loss_main_text}). When $\alpha > \frac 1L - \frac 12$ (i.e. $\zeta > \frac 12$), it holds under~\ref{traj_ass:positive_margin_interpolation},~\ref{traj_ass:loss_decay} for any $j \in [p]$ that
    \begin{equation}
         \frac{\abs{\hat \m_t[j]}}{\sqrt{\hat \vb_t[j]}} \leq \bigO{t^{-\zeta}}, \qquad e^{c_2t}\vb_t[j]  \toinfty{t} \omega_j < \infty, \qquad \omega_j - e^{c_2t}\vb_t[j] \leq \bigO{t^{-2\left(\zeta - \frac12\right)}} \ .
    \end{equation}
    Consequently, Adam asymptotically follows a preconditioned momentum gradient descent trajectory. Namely, 
    it holds for a diagonal matrix $D_\omega$ that
    \begin{equation}\label{eq:main_text_asymyptotic_weighted_l2_direction}
      \dtheta{t} = -D_\omega \eta(t)e^{\frac{c_2}2t}\m_t\left(1+\smallo{1}\right), \qquad (D_\omega)_{j,j} = \begin{cases}
        \omega_j^{-1/2}  &:\omega_j > 0, \\
        1  &:\omega_j = 0
    \end{cases} \ .  
    \end{equation}
    Additionally, for any limit point $\bar \thetab$ of $\frac{\thetab_t}{\norm{\thetab_t}}$, we have that $\hat \thetab := \frac{\bar \thetab}{\qmin(\bar \thetab)^{1/L}}$ is a KKT point of
    \begin{equation}\label{eq:preconditioned_kkt_eq_main_text}
      \min_{\thetab} \frac12 \thetab^{\top}D_\omega^{-1} \thetab \qquad \text{s.t.} \qquad \forall i \in [m]: ~y_if(\x_i;\thetab) \geq 1 \ .  
    \end{equation}
    Here the choice of $(D_\omega)_{j,j} = 1$ for inactive coordinates is arbitrary.
\end{theorem}

\paragraph{Intermediate Regime.}
In the intermediate regime, our result extends the $\ell_\infty$ bias of the slow regime, conditional on gradient directional convergence to a full-support direction; proving or refuting an $\ell_\infty$ bias with no such assumption remains open. We separate this regime from the slow regime 
due to the qualitative difference in the Adam ratio $\frac{\abs{\hat \m_t[j]}}{\sqrt{\hat \vb_t[j]}}$, which is a decaying polynomial instead of asymptotically tending to $1$, and due to the difference in the loss decay rate which we discuss later.
It is also distinct from the fast regime since the assumption of directional convergence of the gradient to a full-support vector holds in the construction of Theorems~\ref{thm:no_regularizer_narrow_main_text},~\ref{thm:no_regularizer_wide_main_text}; thus, the fast regime does not have an $\ell_\infty$ bias even under this assumption. 
\begin{theorem}[Adam in the intermediate regime admits an $\ell_\infty$-max-margin implicit regularizer under gradient convergence to a full-support vector]
\label{thm:intermediate}
    For any $\frac 1L - 1 \leq \alpha \leq \frac 1L - \frac 12$ (i.e. $0 \leq \zeta \leq \frac 12$) and $p \geq 1$, the 
    function
    $R(\thetab)=\frac12 \norm{\thetab}_\infty^2$ is an implicit margin regularizer (\Cref{def:implicit_regularizer}) for Adam on any $f$ satisfying~\ref{model_ass:1_locally_lipschitz},~\ref{model_ass:2_homogeneous} and with any $\ell \in \mathrm L_{\e}$ (\Cref{eq:loss_main_text}), under the trajectory assumption $\forall j \in [p]:\frac{\g_t[j]}{\norm{\g_t}_1} \toinfty{t} \bar \g_\infty [j] \neq 0$. Moreover, under this assumption and~\ref{traj_ass:positive_margin_interpolation},~\ref{traj_ass:loss_decay} it holds for any $j \in [p]$ that
    \begin{equation}
        \frac{\abs{\hat \m_t[j]}}{\sqrt{\hat \vb_t[j]}} \leq \bigO{t^{-\zeta}} \ .
    \end{equation}
\end{theorem}

\paragraph{Loss Decay Rates.}
We provide loss decay rates, showing that $-\log \Lcal(\thetab_t) = \frac{c_2}{2}t \pm \smallo{t}$ in the intermediate and fast regimes (whenever $\zeta>0$), thus uncovering a direct connection between the rate and the value of $c_2$ (analogous to $- \log \beta_2$ in discrete Adam). In particular, this demonstrates that Adam adapts to its learning rate schedule: the leading order $\frac{c_2}2t$ is independent of the schedule within the intermediate and fast regimes. This result contrasts with the slow regime, 
in which we show that $-\log \Lcal(\thetab_t) \asymp t^{L(\alpha+1)} = \smallo{t}$ -- a rate which is both sublinear and schedule-dependent. Similar slow-regime results were previously shown in the proof of Lemma C.19 in~\citet{gronich2026implicitbiasadammuon}, assuming $f \in C^1$ and directional convergence of the parameters (assumptions which we omit), and in Theorem 4.5 in~\citet{zhang2024_adam_linear} for discrete Adam in the special case of linear classifiers. For more precise error terms, see Theorem~\ref{thm:loss_decay_rates_appendix}.

\begin{theorem}\label{thm:loss_decay_rates_main_text}
    Let $f$ satisfy~\ref{model_ass:1_locally_lipschitz}~\ref{model_ass:2_homogeneous}, let $\ell \in \mathrm L_{\e}$ (\Cref{eq:loss_main_text}), and let $\thetab_t$ be an Adam trajectory of $f$ assuming ~\ref{traj_ass:positive_margin_interpolation},~\ref{traj_ass:loss_decay}. Writing $U(t):=-\log\Lcal(\thetab_t)$, it holds that
    \begin{equation}
    \begin{alignedat}{2}
        U(t) &= \frac{c_2}{2}t \pm \bigO{\log t}, &\qquad& \zeta > \frac12, \\
        U(t) &= \frac{c_2}{2}t \pm \smallo{t}, && 0<\zeta \leq \frac12, \\
        U(t) &= \Theta\left(t^{L(\alpha+1)}\right), && \zeta \leq 0 \ .
    \end{alignedat}
    \end{equation}
\end{theorem}

\section{Proof ideas}

In this section, we discuss some main proof ideas. All formal proofs are deferred to the appendix.

\subsection{Second-moment convergence}
Recalling the Adam update \(
    \frac{d\thetab_t}{dt} = -\eta(t)\frac{\hat \m_t}{\sqrt{\hat \vb_t}} \ 
\),
studying the asymptotic properties of Adam boils down to the behavior of $\m_t, \vb_t$, which can be written explicitly as:
\[
\m_t = c_1e^{-c_1t}\int_0^t e^{c_1s}\g_s \, ds \ , \qquad \vb_t = c_2e^{-c_2t}\int_0^t e^{c_2s}\g_s^2 \, ds \ .
\]

We discover a critical transition at $\zeta = \frac12$, separating the fast regime from the intermediate and slow regimes:
\begin{proposition}\label{prop:second_moment_cases_main_text}
    Under assumptions \ref{traj_ass:positive_margin_interpolation},~\ref{traj_ass:loss_decay} for Adam, when $\zeta > \frac 12$, $e^{c_2t}\vb_t[j] \toinfty{t} \omega_j < \infty$ for all $j \in [p]$, and when $\zeta \leq \frac 12$, $e^{c_2t}\vb_t[j] \toinfty{t} \infty$ for all $j \in [p]$ with $\abs{\g_t[j]} \geq \Omega(\norm{\g_t}_1)$.    
\end{proposition}

Proposition~\ref{prop:second_moment_cases_main_text} is proved in Appendix~\ref{sec:dynamic_analysis} in Corollary~\ref{cor:second_moment_freezing} and Proposition~\ref{prop:no-scalar-freezing}. We now use Proposition~\ref{prop:second_moment_cases_main_text} to analyze the three regimes.

\subsection{Three regimes}

\vspace{1.0em}
\noindent
\textbf{Slow regime: Adam ratio approximates sign-GD.}
Suppose $-1<\alpha<\frac1L-1$ (i.e. $\zeta<0$).
Under the assumption that $f \in C^1$ and parameter direction converges, Lemma C.19 from \citet{gronich2026implicitbiasadammuon} gives, through slow variation of the model outputs $\abs{\frac{dy_if(\x_i;\thetab_t)}{dt}} = \smallo{1}$, that
\begin{equation}
    \m_t = \g_t + \smallo{\norm{\g_t}_2}, \qquad \vb_t = \g_t^2 + \smallo{\norm{\g_t}_2^2} \ .
    \label{eq:phase-slow-buffer-tracking}
\end{equation}
Now fix a coordinate $j$ satisfying 
the eventual lower bound $\abs{\g_t[j]}\geq a_j\norm{\g_t}_2, a_j>0$. Continuity of $\g_t$ and \eqref{eq:phase-slow-buffer-tracking} then imply
\begin{equation}
    \frac{\hat \m_t[j]}{\g_t[j]}\toinfty{t}1,
    \quad
    \frac{\hat \vb_t[j]}{\g_t^2[j]}\toinfty{t}1, \qquad \frac{\hat \m_t[j]}{\sqrt{\hat \vb_t[j]}} = \sign{\g_t[j]}(1+ \smallo{1}) \ ,
    \label{eq:phase-slow-tracking}
\end{equation}
inducing asymptotic sign-GD behavior, hence an $\ell_\infty$ bias.

\vspace{1.0em}
\noindent
\textbf{Intermediate regime: sign-GD with a scalar envelope.}
Now suppose $\frac1L-1\leq\alpha\leq\frac1L-\frac12$, (i.e. $0 \leq \zeta \leq \frac 12$), and assume
$$\forall j \in [p]: \qquad \frac{\g_t[j]}{\norm{\g_t}_1} \toinfty{t} \bar \g_\infty [j] \neq 0 \ .$$ 
Since $c_1 \geq c_2$ and $\norm{\g_t}_1 \to 0$, Proposition~\ref{prop:second_moment_cases_main_text} gives $\int_0^t e^{c_1s}\norm{\g_s}_1 \,ds  \gtrsim \int_0^t e^{c_2s}\norm{\g_s}_1^2 \, ds \to \infty$, implying that the directional limit transfers to the exponential moving average (Lemma~\ref{lemma:weighted-cesaro}):
\[
\frac{\hat \m_t[j]}{\ema{\norm{\g_t}_1}{c_1}} \toinfty{t} \bar \g_\infty [j], \quad  \frac{\hat \vb_t[j]}{\ema{\norm{\g_t}_1^2}{c_2}} \toinfty{t} \bar \g_\infty^2 [j] \ .
\]
\[
\frac{\hat \m_t}{\sqrt{\hat \vb_t}} = \left(\frac{\ema{\norm{\g_t}_1}{c_1}}{\sqrt{\ema{\norm{\g_t}_1^2}{c_2}}}\right)\sign{\bar \g_\infty}(1 + \smallo{1}) \ .
\]
In other words, $\thetab_t$ asymptotically follows a sign-GD trajectory with a scalar envelope.

\vspace{1.0em}
\noindent
\textbf{Fast regime: weighted momentum GD.}
Proposition~\ref{prop:second_moment_cases_main_text} gives $\vb_t[j]=e^{-c_2t}(\omega_j+\smallo{1})$, $\omega_j < \infty$, so Adam sees an asymptotically frozen
diagonal preconditioner for momentum gradient descent:
\[
\dtheta{t}[j] = -\eta(t)\frac{\hat \m_t[j]}{\sqrt{\hat \vb_t[j]}} = -\eta(t)e^{\frac{c_2}2t}\omega_j^{-1/2}\m_t[j]\left(1+\smallo{1}\right) \ .
\]
In the following subsection we show why this rules out a universal implicit bias.

\begin{figure}[t]
  \centering

  \begin{subfigure}[t]{0.49\textwidth}
    \centering
    \vspace{0pt}
    \includegraphics[
      height=0.62\linewidth,
      keepaspectratio
    ]{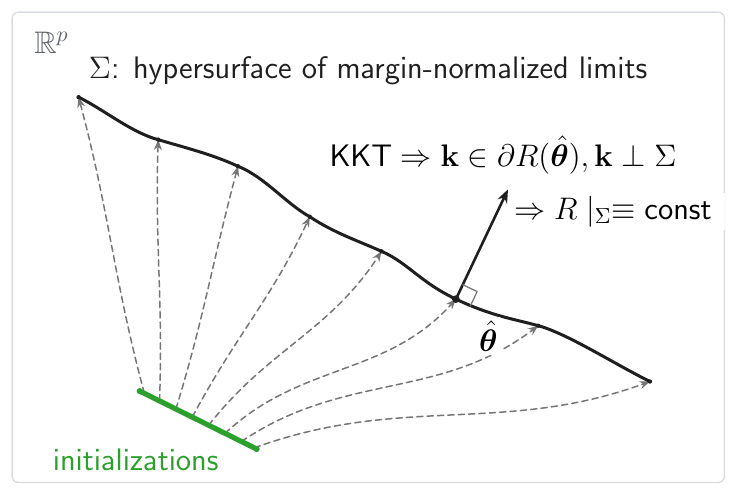}
    \caption{}
    \label{fig:adam-construction-kkt}
  \end{subfigure}
  \hfill
  \begin{subfigure}[t]{0.49\textwidth}
    \centering
    \vspace{0pt}
    \includegraphics[
      height=0.62\linewidth,
      keepaspectratio,
    ]{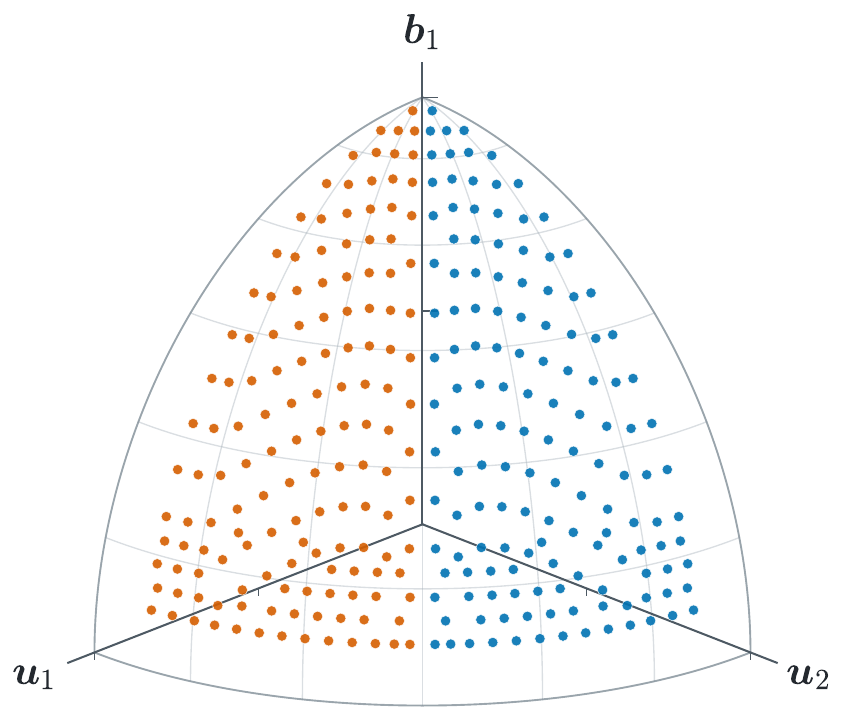}
    \caption{}
    \label{fig:adam-sphere}
  \end{subfigure}

  \caption{\textbf{Initialization-dependent limiting direction forces constancy of $R$ in the fast
  regime.}
  (a) Varying the initialization produces distinct margin-normalized limits
  on the hypersurface $\Sigma = \{\thetab \mid f(\x_1; \thetab) = 1\}$, due to large early-phase gradients. Since only $\x_1$ has an active margin there, KKT forces $\kb \perp \Sigma$ for the subgradient $\kb \in \partial R(\hat \thetab)$, implying $R$ is constant on $\Sigma$.
  (b) Initialization-dependent parameter directions on the unit sphere, reached in experiments running Adam on the width-$1$ model~\eqref{eq:construction_model_main_text} (with $L=2$), each marker representing a different initialization. Blue and orange markers denote the
  \(u_1<u_2\) and \(u_2<u_1\) sectors, respectively, obtained from two distinct yet symmetric datasets. 
  }
  \label{fig:adam-construction}
\end{figure}

\subsection{No universal implicit regularizer in the fast regime}\label{subsec:no_universal_main_text}
We introduce the explicit width-$1$ construction for the fast regime, which serves to prove Theorem~\ref{thm:no_regularizer_narrow_main_text}. For a full proof see Appendix~\ref{section:width_1_appendix}. The core idea is as follows: large initialization-dependent gradients early in the trajectory set the frozen direction of the second moment $\vb_t$, while the first moment $\m_t$ asymptotically tracks the late-phase gradients regardless of initialization. Thus, the late-phase update direction, and hence the limiting parameter direction, is made arbitrary. This implies the triviality of the regularizer $R$. Theorem~\ref{thm:no_regularizer_wide_main_text} then follows from a lifting argument to an arbitrarily wide network. 

Concretely, observe the model~\eqref{eq:construction_model_main_text} and the following dataset parameterized by $M > B >1, \Lambda > 0$, 
\begin{equation}\label{eq:construction_dataset_main_text}
 \x_1 = \frac1\Lambda(B,1), \quad \x_2 = \frac1\Lambda(M,0), \qquad y_1=y_2 = 1   \ .
\end{equation}
The main argument shows that, allowing a continuously parameterized initialization $\thetab_0 = \thetab_0(\rho, z, \pi_1,...,\pi_{L-1})$:
\begin{proposition}[Arbitrary margin-normalized convergence to a hypersurface]\label{prop:arbitrary_directional_convergence_main_text}
    For any $B >1, \Lambda > 0$, denote the hypersurface
    \begin{equation}\label{eq:sigma_BC_main_text}
  \Sigma_{B,\Lambda} = \left \{\thetab\in \bbR^{L+1} \mid  0 < u_1 < u_2 , \forall j \in [L-1] ~ b_j > 0, f(\x_1; \thetab)=1 \right \}  \ .
    \end{equation}
    Then for any $\hat \thetab \in \Sigma_{B,\Lambda}$, and for any large enough $M > 0$ there exists an initialization $\thetab_0$ for which
    $$\frac{\thetab_t}{\qmin^{1/L}(\thetab_t)} \toinfty{t} \hat{\thetab} \ .$$
    Moreover, $\x_1$ is the unique minimum-margin example for $\hat \thetab$, that is, $f(\x_2; \hat \thetab) >1$.
\end{proposition}

\begin{corollary}\label{cor:constant_on_sigma_main_text} 
Any implicit margin regularizer $R$ on $f_p$ (Equation~\eqref{eq:construction_model_main_text}) with the exponential loss is constant on $\Sigma_{B, \Lambda}$.    
\end{corollary}
\begin{proof}
 Take any $C^1$ path $\psi(s) \subseteq \Sigma_{B,\Lambda}$. Fixing $s$ and invoking Proposition~\ref{prop:arbitrary_directional_convergence_main_text} and the definition of $R$ implies the existence of $\kb_s \in \partial R(\psi(s)), \h_s \in \partial f(\x_1; \psi(s))$, for which
$$\kb_s = \lambda_s \h_s \ .$$
But $\innerprod{\h_s}{\dot \psi(s)}=0$ since $\Sigma_{B,\Lambda}$ is a level set of $f(\x_1; \cdot)$, so by the chain rule
\[
 (R\circ\psi)'(s)
 =\innerprod{\kb_s}{\dot\psi(s)}=0
 \quad\text{for almost every }s.
\]   
\end{proof}

Figure~\ref{fig:adam-construction} describes the main triviality argument and demonstrates arbitrary initialization-dependent directional convergence in experiments (see Appendix~\ref{sec:experiment_details} for full experimental details).

Continuously varying $B,\Lambda$, and utilizing common boundary points of the hypersurfaces $\Sigma_{B,\Lambda}$, gives constancy of $R$ on the whole open sector where $0 < u_1 < u_2$ and $b_j > 0$ for all $j$, and a symmetry argument proves constancy on the whole of $\bbR^{L+1}$, proving Theorem~\ref{thm:no_regularizer_narrow_main_text}. Then, a lifting argument is used to generalize the result to networks of arbitrary width and input dimension (see Appendix~\ref{sec:arbitrary_widths}); since the lifted hypersurfaces do not cover the whole space, but do linearly span it, convexity of $R$ is required to imply constancy (Theorem~\ref{thm:no_regularizer_wide_main_text}).

\section{Experiments}

\begin{figure*}[!htb]
  \centering

  \raisebox{-0.5\height}{\includegraphics[width=0.40\linewidth]
      {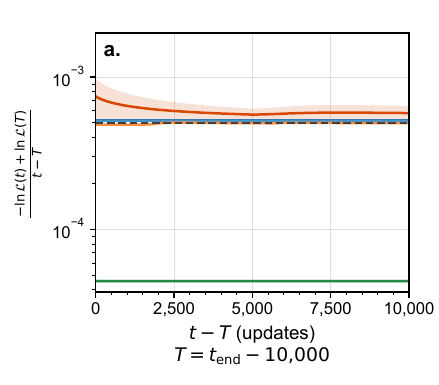}}\hfill
  \raisebox{-0.5\height}{\includegraphics[width=0.40\linewidth]
      {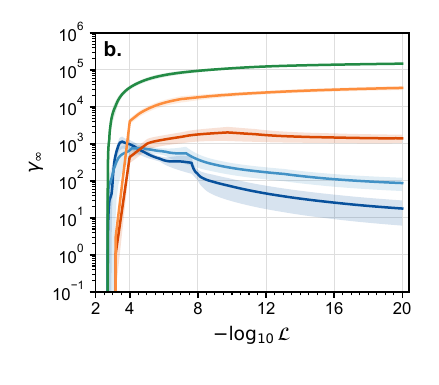}}\hfill
  \raisebox{-0.5\height}{\includegraphics[width=0.17\linewidth]
      {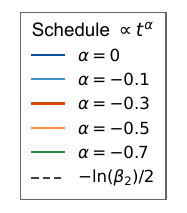}}

  \par\medskip

  \raisebox{-0.5\height}{\includegraphics[width=0.40\linewidth]
      {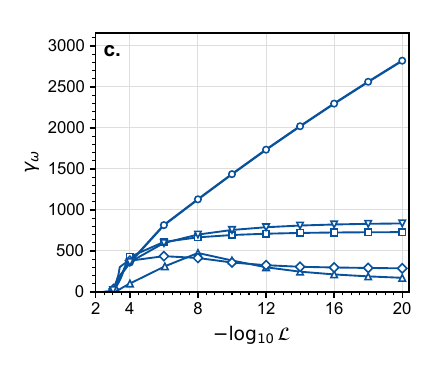}}\hfill
  \raisebox{-0.5\height}{\includegraphics[width=0.40\linewidth]
      {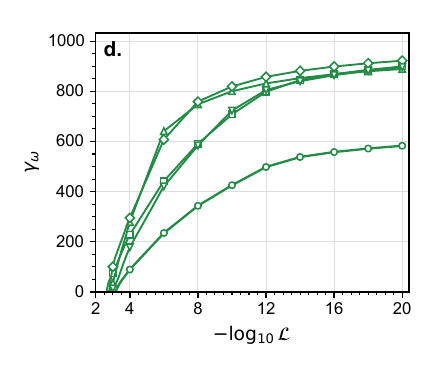}}\hfill
  \raisebox{-0.5\height}{\includegraphics[width=0.17\linewidth]
      {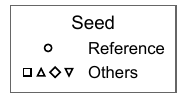}}

  \caption{
    \textbf{Adam loss decay rate and margins on a 3-layer network with $\varepsilon=0$ and initialization scale $10^{-4}$}.
    (a) Late-phase slope of $-\ln\mathcal{L}$ on the $10,000$ last
    steps of each trajectory, and a reference at $-\ln(\beta_2)/2$.
    (b) The $\ell_\infty$ margin.
    (c)--(d) For $\alpha=0$ and $\alpha=-0.7$, respectively,
    the weighted margin $\gamma_\omega$ using weights from one
    reference seed's endpoint.
    In (a)--(b), lines are means across 5 seeds and shading denotes
    one standard error.
  }
  \label{fig:experiment_results_main_text}
\end{figure*}

We empirically validate our results by training $3$-layer fully-connected ReLU networks to classify $m=1024$ MNIST \citep{lecun2002gradient} images (even vs. odd digits) with the logistic loss. Here the critical schedule rates are $\frac 1L - 1 \approx -0.67, \ \frac 1L  -\frac12 \approx -0.17$. We compare margin values for three regimes of Adam: slow ($t^{-0.7}$), intermediate ($t^{-0.5}, t^{-0.3}$) and fast ($t^{-0.1}, t^{0}$), with a common base learning rate $3 \times 10^{-4}$.
For experimental details see Appendix~\ref{sec:experiment_details}.\footnote{Code for reproducing the experiments is available at
\url{https://github.com/EitanGronich/adam-implicit-bias-lr-schedule}.}

For our first experiment we train the networks with $\varepsilon = 0$ until training loss reaches a small value of $10^{-20}$ (as explained in Section~\ref{sec:problem_setting}, we set the update to $0$ for coordinates with $\hat \vb_t = 0$). Initial parameters are multiplied by a factor of $10^{-4}$, since a small initialization is known to accelerate arrival to the asymptotic regime \citep{moroshko2020implicit,woodworth2020kernelrichregimesoverparametrized}. Results are shown in Figure~\ref{fig:experiment_results_main_text}. Late-phase slopes of $- \ln \Lcal$ closely match the predicted value $\frac{c_2}2$ (i.e. $-\frac{\ln (\beta_2)}2$) 
for $\alpha \geq -0.5$ , while the slow schedule $\alpha=-0.7$ displays a considerably slower rate. 
The $\ell_\infty$ margin, denoted $\gamma_\infty$, separates the three regimes according to our implicit bias characterization: the slow schedule achieves the largest margin, with fast schedules displaying small, decreasing margins. The margins of the intermediate schedules lie between them. We also plot the trajectory-dependent weighted-$\ell_2$ margin, 
denoted here $\gamma_\omega$, approximating the asymptotic weights $\omega$ by calculating them from the trajectory endpoint $t_{\text{end}}$. To make margin scales comparable, we normalize $\omega^{1/2}$ to unit $\ell_2$ norm; for each fixed $\omega$, this only rescales margin values by a uniform constant. Explicitly,\footnote{By this definition $\norm{\cdot}_\omega$ is a seminorm if $\vb_{t_{\text{end}}}[j] = 0$ for some $j$; one could choose an arbitrarily small constant for $\omega_j$ for these coordinates (in Theorem~\ref{thm:fast_regime_kkt_main_text} we choose $\omega_j=1$).} 
\begin{equation}\label{eq:gamma_omega}
\gamma_\omega(\thetab) = \frac{\qmin(\thetab)}{\norm{\thetab}_{\omega}^L}, \qquad \norm{\thetab}_{\omega} = \left(\sum_j \overline {\omega_j^{1/2}}\theta_j^2\right)^{1/2}, \qquad  \omega =\beta_2^{-t_{\text{end}}}\vb_{t_{\text{end}}}, \qquad \overline{\omega^{1/2}} = \frac{\omega^{1/2}}{\norm{\omega^{1/2}}_2} \ .
\end{equation}

For each of the schedules \(\alpha=0\) and \(\alpha=-0.7\), we select a reference run, fix \(\omega\) using its endpoint weights, and evaluate the resulting margin along its trajectory and those of four additional seeds. In the fast regime, the reference trajectory attains the largest margin under its own endpoint metric among the trajectories compared, consistent with our characterization of the trajectory-dependent implicit bias. In the slow regime, the weighted margin is not maximized by its own trajectory relative to others, consistent with the different implicit bias predicted for this regime. See Appendix~\ref{sec:experiment_details} for corresponding plots with the remaining schedules and reference seeds.

\begin{figure}[t]
  \centering

  \raisebox{-0.5\height}{\includegraphics[width=0.40\linewidth]
      {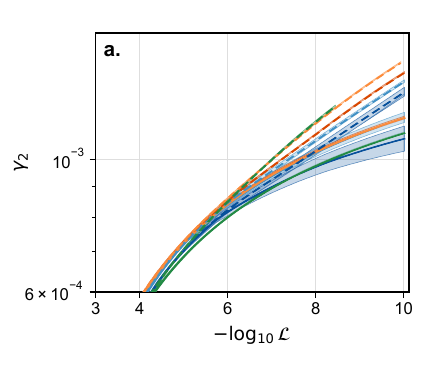}}\hfill\raisebox{-0.5\height}{\includegraphics[width=0.40\linewidth]
      {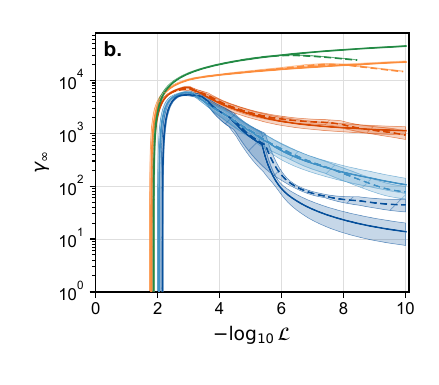}}\hfill\raisebox{-0.5\height}{\shortstack{\includegraphics[width=0.17\linewidth]
        {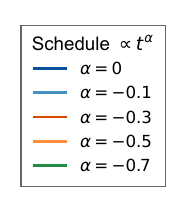}
      \\[\smallskipamount]
      \includegraphics[width=0.17\linewidth]
        {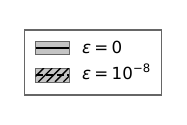}}}

  \par\medskip

  \raisebox{-0.5\height}{\includegraphics[width=0.40\linewidth]
      {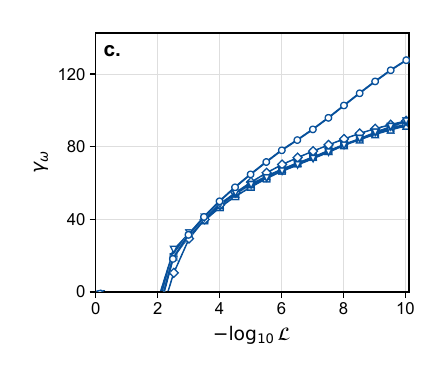}}\hfill\raisebox{-0.5\height}{\includegraphics[width=0.40\linewidth]
      {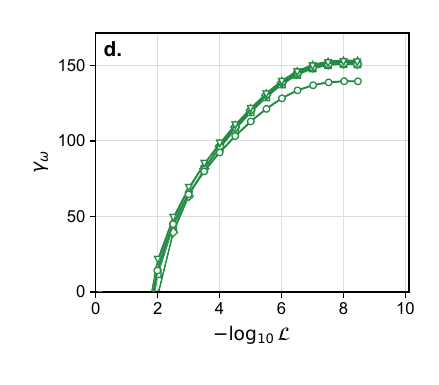}}\hfill\raisebox{-0.5\height}{\includegraphics[width=0.17\linewidth]
      {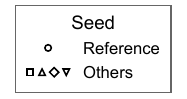}}

  \caption{
    \textbf{Adam margins for $\varepsilon=0,10^{-8}$ on a 3-layer network with
    initialization scale $1$.}
    (a)--(b) $\ell_2$ and $\ell_\infty$ margins, respectively.
    Solid curves and plain bands correspond to $\varepsilon=0$;
    dashed curves and hatched bands correspond to
    $\varepsilon=10^{-8}$.
    Curves show means across five seeds; shading denotes one
    standard error, with boundaries outlined.
    (c)--(d) For $\alpha=0$ and $\alpha=-0.7$, respectively,
    the weighted margin $\gamma_\omega$ using weights from one
    reference seed's endpoint.
  }
  \label{fig:epsilon-margins}
\end{figure}

Figure~\ref{fig:epsilon-margins} shows the results of our second experiment, comparing $\varepsilon = 0$ and $\varepsilon=10^{-8}$ (the default PyTorch~\citep{paszke2019pytorch} value) 
in a 
setting with standard initialization. Here we train until a loss of $10^{-10}$ or up to 1 million steps. For $\varepsilon = 10^{-8}$, \(\varepsilon\) eventually dominates \(\sqrt{\hat{\mathbf v}_t}\) in the denominator, so that
all trajectories are expected to have an implicit bias towards $\ell_2$ margin maximization (see \citet{wang2021_adam_eps_homoegeneous} for an analysis covering some cases). See Appendix~\ref{sec:experiment_details} for an analysis of the effect of $\varepsilon$ on the Adam update along the trajectory.

The \(\varepsilon=0\) and \(\varepsilon=10^{-8}\) runs exhibit similar margins through most of training, consistent with \(\varepsilon\) having negligible effect during this stage. In the late phase of the $\varepsilon = 10^{-8}$ trajectory, corresponding to a more significant effect of $\varepsilon$, we see a decrease in $\gamma_\infty$ for $\alpha \leq -0.5$, 
and an increase in $\gamma_2$. Nevertheless, $\gamma_\infty$ shows a clear ordering between the slow, intermediate and fast regimes until the end of training; this suggests 
a significant lingering $\ell_\infty$ maximization effect from the stage where $\varepsilon$ is negligible. 

The weighted-\(\ell_2\) margin comparison shows the same pattern with \(\varepsilon=10^{-8}\) and standard initialization: the reference trajectory attains the largest margin under its own endpoint metric among the trajectories compared for \(\alpha=0\), but not for \(\alpha=-0.7\). For additional seeds, schedules and $\varepsilon = 0$ see Appendix~\ref{sec:experiment_details}.

\section{Discussion and future directions}
Sign-based descent has served as a tractable surrogate for Adam, informing explanations of its optimization geometry and performance, as well as motivating optimizer design and training recommendations \citep{balles2020geometry,zhuang2020adabelief,liu2024sophia,kunstner2024heavy,li2025optimization}. On the other hand, theoretical and empirical studies have identified differences between Adam and sign-based methods, including Signum \citep{cattaneo2024implicit,orvieto2025secret,baek2026persample}. Our results show that similarity between implicit biases of Adam and sign gradient descent on homogeneous models depends on the learning rate schedule, further highlighting that transferring conclusions from sign descent to Adam requires careful justification for the regime under consideration.

For the different learning-rate schedule regimes, we analyzed the implicit bias of Adam and obtained both negative and positive results.
The trajectory-dependent weighted-$\ell_2$ bias of Adam in the fast regime naturally raises the question of transferability of implications of the $\ell_2$ KKT bias of gradient descent. These include, notably, a vulnerability to data leakage: previous works demonstrated that training data can be reconstructed from networks trained with gradient descent \citep{haim2022reconstructing, buzaglo2023deconstructing, oz2026reconstructing,golbari2025impmia}, while a recent work argues reconstruction is difficult for Adam because of its supposed $\ell_\infty$ bias \citep{habani2026optimizerleaksmostoptimizer}.
Since we show that in some common settings Adam does not maximize the $\ell_\infty$ margin, and instead maximizes a trajectory-dependent weighted $\ell_2$ margin, it
remains open whether 
its implicit bias
can be leveraged to build a data privacy attack. 
Additional implications of $\ell_2$ KKT conditions include 
positive implications in some settings for benign overfitting and generalization \citep{frei2023benign,frei2023double,kornowski2023tempered}. 

Finally, examining the implicit bias of Adam on non-homogeneous models, as in \citet{nacson2019lexicographic,kunin2022asymmetric,cai2025_nearlyhomogeneous}, is a promising and highly motivated line of future inquiry, due to the prevalence of these models in practice.

\vspace{-0.02cm}
\section*{AI Disclosure}
\vspace{-0.02cm}
We were assisted by AI tools for this work. AI was used to help ideate and simplify proofs, to write code and run experiments, and to suggest improvements in writing and literature review. We have reviewed and verified all AI-assisted work.

\vspace{-0.02cm}
\section*{Acknowledgments}
\vspace{-0.02cm}
This research is supported by the Israel Science Foundation (grant No. 2574/25), by a research grant from Mortimer Zuckerman (the Zuckerman STEM Leadership Program), and by research grants from the Center for New Scientists at the Weizmann Institute of Science, and the Shimon and Golde Picker -- Weizmann Annual Grant.  

\FloatBarrier

\bibliographystyle{plainnat}
\bibliography{refs}

\clearpage

\tableofcontents
\newpage

\appendix

\section{Clarke Subgradients and Loss Families}\label{sec:clarke_loss}
We use the standard Clarke subdifferential \citep{Cla75}, which is well defined for any locally Lipschitz function $R$ and denoted $\partial R$. For further discussion of Clarke subdifferentials for classes of regular functions, see~\citet{Dav+20, bolte2007clarke}. We also use the following standard definition (e.g.~\citet{Dav+20}):
\begin{definition}\label{def:admits_chain_rule}
We say a locally Lipschitz function $R : \bbR^{p} \to \bbR$ \emph{admits a chain rule} if for any locally absolutely continuous path $\psi : [0,1] \to \bbR^p$, for almost any $s \in [0,1]$ and for any $h \in \partial R(\psi(s))$ it holds that
\[
\frac{dR(\psi(s))}{ds} = \angles{\dot \psi(s)}{h} \ .
\]
\end{definition}

Recall Equation~\ref{eq:loss_main_text} for the definition of the family of log-concave, exponentially tailed losses $\mathrm L_{\e}$.
\begin{lemma}
     Let $\thetab_t :[0,\infty)\to \bbR^p$ be locally absolutely continuous, let $\{\x_i, y_i\}_{i=1}^m \subseteq \bbR^d \times \{\pm 1\}$ and let $\Lcal(\thetab) = \sum_{i=1}^m \ell(y_if(\x_i; \thetab))$ where $\ell =e^{-\varphi} \in \mathrm L_{\e}$ and $f$ is locally Lipschitz and admits a chain rule. For almost every $t$, any subgradient $\g_t\in\partial\Lcal(\thetab_t)$ can be written as
    \begin{equation}\label{eq:clarke_gradient_decomposition}
        \g_t
        =-\sum_{i=1}^m
          \varphi'(q_i(t))e^{-\varphi(q_i(t))}\mathbf h_i(t),
        \qquad q_i(t) := y_if(\x_i; \thetab_t), \qquad 
        \mathbf h_i(t)\in\partial q_i(\thetab_t),
    \end{equation}
    and it holds that
    \begin{equation}\label{eq:loss_chain_rule}
        \frac{d}{dt}\Lcal(\thetab_t)
        =\innerprod{\g_t}{\dot\thetab_t}.
    \end{equation}
\end{lemma}
\begin{proof}
    Equation~\eqref{eq:clarke_gradient_decomposition} follows, for instance, from Corollary A.2 of \citet{gronich2026implicitbiasadammuon}. \eqref{eq:loss_chain_rule} follows since $f(\x_i; \cdot)$, and therefore $\Lcal$, admit a chain rule (for closure under composition see \citet{lyu_li_homogeneous}, Theorem I.4).
\end{proof}

\section{EMA - Definition and Lemmas}\label{sec:ema}
We follow \citet{barakat2021convergence, ma2022qualitative, belotto2020general, wang2021_adam_eps_homoegeneous, gronich2026implicitbiasadammuon} in working with continuous-time versions of momentum and Adam. 

We define for a locally integrable $g \in L^1_{\mathrm{loc}}$,
\begin{align}
    \ema{g}{c}(t) &:= c\int_0^t e^{-c(t-s)}g(s)\,ds \ ,\label{eq:def_ema} \\
    \emahat{g}{c}(t) &:=
    \begin{cases}
    \displaystyle (1-e^{-ct})^{-1} \ema{g}{c}(t),&t>0,\\[1ex]
    g(0),&t=0,
    \end{cases} \label{eq:def_ema_hat} 
\end{align}
And for a locally square-integrable $g \in L^2_{\mathrm{loc}}$
\begin{align} 
    \adam{g}{c_1}{c_2}(t) &:= \begin{cases} 
    \frac{\emahat{g}{c_1}(t)}{\sqrt{\emahat{g^2}{c_2}(t)}}, & \emahat{g^2}{c_2}(t) > 0, \\
    0, &  \text{otherwise}
    \end{cases} \ .
\end{align}
We define continuous-flow Adam for a measurable choice of subgradients $\g_t \in \partial \Lcal(\thetab_t)$, smoothing constants $c_1,c_2>0$ and a learning rate schedule $\eta(t)$ with the following ODE:
\begin{equation}
    \dtheta{t} = -\eta(t)\adam{\g_t}{c_1}{c_2} \ ,
\end{equation}
for almost every $t$.

Discrete Adam~\citep{kingma2014adam} with time steps $n \geq 0$ and subgradients $\g_n \in \partial \Lcal(\thetab_{n-1})$ for $n \geq 1$ is defined as follows for $\beta_1,\beta_2 \in [0,1)$ and $\varepsilon \geq 0$:
\begin{alignat}{4}
    \m_n &= \beta_1\m_{n-1} + (1-\beta_1)\g_n
    &\quad &(n \geq 1),
    &\qquad \m_0 &= \zero,
    &\qquad \hat \m_n &= (1-\beta_1^n)^{-1}\m_n, \\
    \vb_n &= \beta_2\vb_{n-1} + (1-\beta_2)\g_n^2
    &\quad &(n \geq 1),
    &\qquad \vb_0 &= \zero,
    &\qquad \hat \vb_n &= (1-\beta_2^n)^{-1}\vb_n, \\
    \thetab_n &= \thetab_{n-1}
    - \eta_n \frac{\hat \m_n}{\sqrt{\hat \vb_n}+\varepsilon}
    &\quad &(n \geq 1).
\end{alignat}
Here the square, square root and division are taken coordinate-wise.

The definitions of continuous momentum follow previous works. See the discussion on the connection between $c_i$ and $\beta_i$ in \citet[Appendix B.2]{gronich2026implicitbiasadammuon}, which explains why $c_i \approx - \log (\beta_i)$ when matching the continuous time clock to the iteration count $t \leftrightarrow n$. Note that while the works of \citet{barakat2021convergence,belotto2020general} consider a training regime in which the hyperparameters $\beta_i$ are adjusted during training to account for learning rate decay, we adopt the practical setting in which $\beta_i$ remain fixed, therefore fixing $c_i$ and adding an external learning rate schedule $\eta(t)$ to the definition of Adam.

We proceed to prove useful lemmas about EMA and the Adam ratio.

\begin{lemma}[Continuous EMA ODE]\label{lemma:continuous_ema_ode}
    For any $g \in L_{\mathrm{loc}}^1$, and any $c > 0$, it holds almost everywhere that
    \[
        \frac{d}{dt}\ema{g}{c} = c\left(g(t) - \ema{g}{c}(t)\right)
    \]
\end{lemma}
\begin{proof}
    Follows from direct differentiation.
\end{proof}

\begin{lemma}[Adam Ratio is Universally Bounded]\label{lemma:adam_universal_ratio}
    Let $c_1>c_2/2>0$, and let $g \in L_{\mathrm{loc}}^2$. Then,
    for every $t\geq0$,
    \[
        \left|\mathsf A_{c_1,c_2}[g](t)\right|
        \leq K_A(c_1,c_2)
        :=\frac{c_1}{\sqrt{c_2(2c_1-c_2)}}.
    \]
\end{lemma}
\begin{proof}
At \(t=0\), the bound follows from \(\lvert\mathsf A[g](0)\rvert\le1\le K_A\). Hence assume \(t>0\).

    If $\emahat{g^2}{c_2}(t) = 0$ then $\adam{g}{c_1}{c_2}(t) = 0$ by definition. Otherwise, Cauchy--Schwarz gives
    \[
        \left|\int_0^t e^{-c_1(t-s)}g(s)\,ds\right|^2
        \leq
        \left(\int_0^t e^{-c_2(t-s)}g(s)^2\,ds\right)
        \left(\int_0^t e^{-(2c_1-c_2)(t-s)}\,ds\right).
    \]
    Substituting the normalization factors in the definition of
    $\mathsf A_{c_1,c_2}$ yields
    \[
        \left|\mathsf A_{c_1,c_2}[g](t)\right|
        \leq
        \frac{c_1}{\sqrt{c_2(2c_1-c_2)}}
        \frac{\sqrt{(1-e^{-c_2t})(1-e^{-(2c_1-c_2)t})}}
             {1-e^{-c_1t}}.
    \]
    Now,
    \[
        (1-e^{-c_1t})^2
        -(1-e^{-c_2t})(1-e^{-(2c_1-c_2)t})
        =
        \left(e^{-c_2t/2}
              -e^{-(2c_1-c_2)t/2}\right)^2
        \geq0.
    \]
    The last fraction is therefore at most one, as required.
\end{proof}

\begin{lemma}[Time Derivative of Uncorrected Adam Ratio]\label{lemma:time_derivative_uncorrected_adam_ratio} Let $g \in L^2_{\mathrm{loc}}$ and denote $m = \ema{g}{c_1}, v=\ema{g^2}{c_2}$. The quotient \(m/\sqrt v\) is locally absolutely continuous on \(\{v>0\}\), and the following identity holds for almost every \(t>0\) with \(v(t)>0\):
    \[
        \frac{d}{dt} \left(\frac{m}{\sqrt{v}} \right) = c_1\frac{g}{\sqrt{v}} - \left(c_1-\frac{c_2}2 + \frac{c_2}2\frac{g^2}{v}\right)\frac{m}{\sqrt v}
    \]
\end{lemma}
\begin{proof}
    Direct differentiation, using definitions of $m,v$:
    \begin{align*}
         \frac{d}{dt} \left(\frac{m}{\sqrt{v}} \right) &= \frac{c_1(g - m)\sqrt v  - m\frac{c_2(g^2-v)}{2\sqrt v}}{v}  \ .
    \end{align*}
\end{proof}

\begin{lemma}[Weighted Ces\`aro principle]
\label{lemma:weighted-cesaro}
Let $W:[0,\infty)\to[0,\infty) \in L^1_{\mathrm{loc}}$ with
$\int_0^tW(s)\,ds\to\infty$, and let $h$ be so that $Wh \in L^1_{\mathrm{loc}}$. If
$h(t)\to h_\infty \in (-\infty, \infty)$, then
\begin{equation}
    \frac{\int_0^tW(s)h(s)\,ds}
         {\int_0^tW(s)\,ds}
    \longrightarrow h_\infty.
    \label{eq:weighted_cesaro}
\end{equation}
\end{lemma}

\begin{proof}
Fix $\varepsilon>0$ and choose $T>0$ such that
$\abs{h(s)-h_\infty}\leq\varepsilon/2$ for all $s\geq T$.
Since $W$ and $Wh$ are locally integrable,
\[
    C_T:=\int_0^T W(s)\abs{h(s)-h_\infty}\,ds<\infty.
\]
For all sufficiently large $t\geq T$, the denominator
$\int_0^t W(s)\,ds$ is positive. Splitting the integral at $T$
and using $W\geq0$, we obtain
\[
    \begin{aligned}
        \abs{
            \frac{\int_0^t W(s)h(s)\,ds}
                 {\int_0^t W(s)\,ds}
            -h_\infty
        }
        &=
        \frac{\abs{\int_0^t W(s)(h(s)-h_\infty)\,ds}}
             {\int_0^t W(s)\,ds}\\
        &\leq
        \frac{\int_0^T W(s)\abs{h(s)-h_\infty}\,ds}
             {\int_0^t W(s)\,ds}
        +
        \frac{\int_T^t W(s)\abs{h(s)-h_\infty}\,ds}
             {\int_0^t W(s)\,ds}\\
        &\leq
        \frac{C_T}{\int_0^t W(s)\,ds}
        +\frac{\varepsilon}{2}
         \frac{\int_T^t W(s)\,ds}
              {\int_0^t W(s)\,ds}\\
        &\leq
        \frac{C_T}{\int_0^t W(s)\,ds}
        +\frac{\varepsilon}{2}.
    \end{aligned}
\]
Since $C_T$ is finite and $\int_0^t W(s)\,ds\toinfty{t}\infty$,
there exists $T_1\geq T$ such that
\[
    \frac{C_T}{\int_0^t W(s)\,ds}\leq\frac{\varepsilon}{2}
    \qquad\text{for all }t\geq T_1.
\]
Thus the absolute difference in~\eqref{eq:weighted_cesaro}
is at most $\varepsilon$ for every $t\geq T_1$.
Since $\varepsilon>0$ was arbitrary, the result follows.
\end{proof}

\begin{corollary}[Weighted EMA]\label{lemma:weighted_ema}
Let $0 \leq g \in L^1_{\mathrm{loc}}$, $gh \in L^1_{\mathrm{loc}}$, and assume $h(t) \toinfty{t} h_\infty \in (-\infty, \infty)$ and $e^{ct}\ema{g}{c}(t) \toinfty{t} \infty$. Then
\[
\frac{\ema{gh}{c}(t)}{\ema{g}{c}(t)} \toinfty{t} h_\infty, \qquad \frac{\emahat{gh}{c}(t)}{\emahat{g}{c}(t)} \toinfty{t} h_\infty \ .
\]
\end{corollary}
\begin{proof}
    Follows by definition from~\ref{lemma:weighted-cesaro}, after cancelling $e^{-ct}$ and bias correction terms.
\end{proof}

\begin{lemma}[EMA Weighted by Slowly Varying Multiplier]\label{lemma:ema_with_multiplier}
    Let $0 < g \in L^1_{\mathrm{loc}}$ and let $A:[0,\infty)\to(0,\infty)$ be locally absolutely continuous. Let
    $\varepsilon:[0,\infty)\to\mathbb R$ be bounded and measurable with
    $\varepsilon(t)\to0$. Let $c>0$. Suppose that for some $\mu > 0$
    \[
    \log\left(e^{ct}g(t) \right) = \mu t \pm \smallo{t}, \qquad \abs{\frac{d}{dt} \log A(t)} \leq \bigO{\frac1t} \ .
    \]
    Then,
    \[
        \frac{\mathsf{EMA}_{c}[A(1+\varepsilon)g](t)}
             {\mathsf{EMA}_{c}[g](t)}
        =A(t)(1+\smallo{1}).
    \]
\end{lemma}
\begin{proof}
    Intuitively, the exponential factor in $g$ makes both EMAs depend asymptotically only
    on times close to $t$, while the logarithmic derivative bound makes
    $A$ nearly constant on that window. To make this precise, denote 
    $$\tilde g(t) = e^{ct}g(t), \quad J(t) = \int_0^t \tilde g(s)\,ds, \quad E_t(s) = \frac{A(s)}{A(t)}(1+\varepsilon(s))-1 \ .$$
    Note that the required statement is equivalent, after cancelling shared exponentials and normalizing constants, to:
    \[
        \int_0^t \tilde g(s)\frac{A(s)}{A(t)}(1+\varepsilon(s))\,ds= (1+\smallo{1})J(t) \ ,
    \]
    and hence also to:
    \[
        \int_0^t \tilde g(s)E_t(s)\,ds= \smallo{J(t)} \ .
    \]
    By the assumption, $\log \tilde g(t) = \mu t + \smallo{t}$; this implies $\log J(t) = \mu t + \smallo{t}$ as well. Indeed, for any $0 < \delta < \mu$, eventually $e^{(\mu - \delta)t}\leq \tilde g(t) \leq e^{(\mu + \delta)t}$ for $t \ge t_{\delta}$, so
    \[
    e^{(\mu-\delta)t} \lesssim \int_{t_\delta}^t e^{(\mu-\delta)s}\, ds\leq J(t) \leq \int_{0}^{t_\delta}\tilde g(s)ds \, +\int_{t_\delta}^t e^{(\mu + \delta)s}  \, ds \lesssim e^{(\mu + \delta)t} \ .
    \]
    Taking logarithm and $\delta \to 0$ proves the claim. Next, by the logarithmic derivative assumption, there exist constants $K>0$ and $T_0 \geq 1$ such that $\abs{\frac{d}{dt}\log A} \leq \frac{K}{t}$ for $t \geq T_0$, so
    \begin{equation}\label{eq:A_ratio_bounded}
    \left(\frac{s}{t}\right)^K \leq \frac{A(s)}{A(t)} \leq \left(\frac ts \right)^K, \qquad T_0 \leq s\leq t.    
    \end{equation}
    For $0\leq s<T_0\leq t$, we first apply
\eqref{eq:A_ratio_bounded} at $s=T_0$ to obtain
\[
    \frac{A(T_0)}{A(t)}
    \leq \left(\frac{t}{T_0}\right)^K.
\]
Since $A$ is continuous and positive, the constants
\[
    M:=\max_{0\leq s\leq T_0}A(s),
    \qquad
    B:=\sup_{s\geq0}\abs{\varepsilon(s)}
\]
are finite, and $A(T_0)>0$. Consequently,
\begin{equation}\label{eq:A_ratio_bounded_small_s}
    \frac{A(s)}{A(t)}
    =\frac{A(s)}{A(T_0)}\frac{A(T_0)}{A(t)}
    \leq \frac{M}{A(T_0)}
          \left(\frac{t}{T_0}\right)^K, \qquad s < T_0 \leq t.    
\end{equation}
Combining~\eqref{eq:A_ratio_bounded}--~\eqref{eq:A_ratio_bounded_small_s} gives for some $C > 0$ and sufficiently large $T_1$,
\[
    \begin{aligned}
        \abs{E_t(s)}\leq \frac{A(s)}{A(t)}(1+\abs{\varepsilon(s)})+1\leq Ct^K, \qquad 0 \leq s \leq t, \quad T_1 \leq t \ .
    \end{aligned}
\]
    Fix $\delta \in (0,1)$. On $[0, (1-\delta)t]$ for $t \geq T_1$, we have
    \[
    \abs{\frac{1}{J(t)} \int_0^{(1-\delta)t} \tilde g(s) E_t(s) \, ds }\leq Ct^K \frac{J((1-\delta)t)}{J(t)} = e^{-\mu\delta t + \smallo{t}} \toinfty{t} 0 \ .
    \]
    On $[(1-\delta)t, t]$, \eqref{eq:A_ratio_bounded} gives via the triangle inequality,
    \[
    \sup_{(1-\delta)t \leq s \leq t} \abs{E_t(s)} \leq (1-\delta)^{-K}-1 + (1-\delta)^{-K}\sup_{s \geq (1-\delta)t}\abs{\varepsilon(s)} \toinfty{t} (1-\delta)^{-K} - 1 \ .
    \]
    Which yields
    \[
    \limsup_{t \to \infty}\abs{\frac{1}{J(t)}\int_{(1-\delta)t}^t \tilde g(s) E_t(s) \, ds} \leq (1-\delta)^{-K} - 1 \ . 
    \]
    Taking $\delta \to 0$ and combining with the estimate on $[0, (1-\delta)t]$ gives the desired result.
\end{proof}

\begin{lemma}[$L^p$ bounds]\label{lemma:Lp_bound}
    Let $1 \leq p < \infty$, $g \in L_{\mathrm{loc}}^p$, and $0<\lambda < c$. Then
    \[
        \left(\int_0^t \abs{e^{\lambda s}\ema{g}{c}(s)}^p\,ds\right)^{1/p} \leq \frac{c}{c-\lambda}\left(\int_0^t \abs{e^{\lambda s}g(s)}^p\,ds\right)^{1/p} \ .
    \]
\end{lemma}
\begin{proof}
    Fix $t>0$. Define $h(u)=e^{\lambda u}g(u)$ for $0\leq u\leq t$, and set
$h(u)=0$ otherwise. For $0\leq s\leq t$,
\[
e^{\lambda s}\mathsf{EMA}_{c}[g](s)
=c\int_0^s e^{-(c-\lambda)(s-u)}h(u)\,du
=c\int_0^\infty e^{-(c-\lambda)v}h(s-v)\,dv.
\]
Minkowski's integral inequality states that, for $p\geq1$,
\[
\left(\int_0^t
    \left|\int_0^\infty F(s,v)\,dv\right|^p ds
\right)^{1/p}
\leq
\int_0^\infty
    \left(\int_0^t |F(s,v)|^p ds\right)^{1/p}dv.
\]
We apply it with $F(s,v)=ce^{-(c-\lambda)v}h(s-v)$. Since $h$ vanishes
outside $[0,t]$, a change of variables gives, for every $v\geq0$,
\[
\int_0^t |h(s-v)|^p\,ds
=\int_{-v}^{t-v}|h(u)|^p\,du
\leq\int_0^t e^{p\lambda u}|g(u)|^p\,du.
\]
Consequently,
\[
\begin{aligned}
\left(\int_0^t
    \left|e^{\lambda s}\mathsf{EMA}_{c}[g](s)\right|^p\,ds
\right)^{1/p}
&\leq c\int_0^\infty e^{-(c-\lambda)v}
    \left(\int_0^t |h(s-v)|^p\,ds\right)^{1/p}dv\\
&\leq \frac{c}{c-\lambda}
    \left(\int_0^t e^{p\lambda u}|g(u)|^p\,du\right)^{1/p}.
\end{aligned}
\]
\end{proof}

\section{Adam Dynamics Analysis}\label{sec:dynamic_analysis}
Define the momentum-corrected loss $\Lcal_A$, introduced by \citet{barakat2021convergence}:
\begin{equation}\label{eq:corrected_loss_definition}
    H(t):=\sum_{j=1}^p\frac{m_j(t)^2}{\sqrt{v_j(t)}} = \angles{\m_t}{\frac{\m_t}{\sqrt{\vb_t}}}, \qquad \hat \eta (t) := \frac{\sqrt{1 - e^{-c_2t}}}{1-e^{-c_1t}}\eta(t),
    \qquad
    \Lcal_A(t):=\Lcal(\thetab_t)
       +\frac{\widehat\eta(t)}{2c_1}H(t).
\end{equation}
Summands with $\vb_t[j]=0$ are interpreted as zero.

Set
\begin{equation}\label{eq:mu_definition}
    \mu(t):=2c_1-\frac{c_2}{2}
       -\frac{d}{dt}\log\widehat\eta(t).
\end{equation}
For the polynomial schedule $\eta(t)=\eta_0(1+t)^{\alpha}$ it holds that
$\mu(t)\to\mu_\infty:=2c_1-c_2/2 \geq \frac32 c_1$. Therefore, let $t_A > 0$ with
\[
    \mu(t)\geq\mu_0, \qquad c_1<\mu_0<\mu_\infty,
    \qquad(t\geq t_A).
\]
Also, with the exception of Lemmas~\ref{lemma:surrogate-loss-dissipation},~\ref{lemma:normal_corrected_loss_decay_equivalent} which require no assumptions, and Lemma~\ref{lemma:finite-low-loss-window} which requires only~\ref{traj_ass:positive_margin_interpolation}, we work under the assumptions $\forall t \geq t_0: \gamma_2(\thetab_t) \geq \gamma_0 > 0$ and $\Lcal(\thetab_t) \toinfty{t} 0$ (Assumptions~\ref{traj_ass:positive_margin_interpolation},~\ref{traj_ass:loss_decay}). Lemma~\ref{lemma:normal_corrected_loss_decay_equivalent} shows that Assumption~\ref{traj_ass:loss_decay} implies $\Lcal_A(t) \toinfty{t} 0$. Therefore, wherever we assume~\ref{traj_ass:loss_decay} we assume w.l.o.g that $t_0$ is large enough so that $t_0 \geq t_A$ and
\begin{equation}\label{eq:corrected_loss_tail_bounds}
 \Lcal(\thetab_t) \leq \Lcal_A(t)< 1,
    \qquad (t \geq t_0).
\end{equation}
We also observe the following useful quantities on $t > t_0$:
\begin{align}
    U(t) &= - \log \Lcal(\thetab_t), \qquad &U_A(t) &= - \log \Lcal_A(t), \\
    G(t) &= U^{1- \frac 1L}\Lcal, \qquad &G_A(t) &= U_A^{1 - \frac 1L}\Lcal_A, \\
    Q(t) &= - \log G(t), \qquad &Q_A(t) &= - \log G_A(t) \ , \\
    V(t) &= \int_{t_0}^t e^{c_2s}G^2(s)\, ds, \qquad &V_A(t) &= \int_{t_0}^t e^{c_2s}G_A^2(s)\, ds \ , \\
    &&r_A(t) &= \left(\frac{d}{dt}\log V_A\right)^{1/2} \ .
\end{align}

\begin{lemma}[Corrected-loss and radial identities]
\label{lemma:surrogate-loss-dissipation}
For almost any $t$,
\begin{align}
    \dot H(t)
    ={}&2c_1\innerprod{\g_t}{\frac{\m_t}{\sqrt{\vb_t}}}
       -\left(2c_1-\frac{c_2}{2}\right)H(t) -\frac{c_2}{2}\sum_{j=1}^p
        \frac{m_j(t)^2g_j(t)^2}{v_j(t)^{3/2}},
    \label{eq:H_derivative}
\end{align}
and the following dissipation identity holds:
\begin{equation}\label{eq:corrected_loss_dissipation}
    \dot\Lcal_A(t)
    =-\mu(t)\bigl(\Lcal_A(t)-\Lcal(\thetab_t)\bigr)
     -\frac{c_2\widehat\eta(t)}{4c_1}
       \sum_{j=1}^p
       \frac{m_j(t)^2g_j(t)^2}{v_j(t)^{3/2}} \leq -\mu(t)\bigl(\Lcal_A(t)-\Lcal(\thetab_t)\bigr).
\end{equation}
In particular, $\Lcal_A$ is nonincreasing on every interval on which
$\mu\geq0$.
In addition,
\begin{equation}\label{eq:common_radial_momentum_equation}
    \left(\frac{d}{dt}+c_1\right)
    \left(-\innerprod{\thetab_t}{\m_t}\right)
    =-c_1\innerprod{\thetab_t}{\g_t}+\widehat\eta(t)H(t).
\end{equation}
\end{lemma}

\begin{proof}
Differentiate $H=\sum_jm_j^2v_j^{-1/2}$ and substitute
$\dot m_j=c_1(g_j-m_j)$ and $\dot v_j=c_2(g_j^2-v_j)$ to obtain
\eqref{eq:H_derivative}.  By the chain rule,
\[
    \dot\Lcal(\thetab_t)
    =\angles{\g_t}{\dtheta{t}}=-\widehat\eta(t)
      \innerprod{\g_t}{\frac{\m_t}{\sqrt{\vb_t}}}.
\]
When $\Lcal_A$ is differentiated, this term cancels the first term
in \eqref{eq:H_derivative}.  Using
$\widehat\eta H/(2c_1)=\Lcal_A-\Lcal$ gives
\eqref{eq:corrected_loss_dissipation}.
Finally,
\[
    \frac{d}{dt}\left(-\innerprod{\thetab_t}{\m_t}\right)
    =-\innerprod{\dot\thetab_t}{\m_t}
     -\innerprod{\thetab_t}{\dot\m_t}
    =\widehat\eta H-c_1\innerprod{\thetab_t}{\g_t}
       +c_1\innerprod{\thetab_t}{\m_t},
\]
which is \eqref{eq:common_radial_momentum_equation}.
\end{proof}

\begin{lemma}[Equivalence of ordinary and corrected loss decay]
\label{lemma:normal_corrected_loss_decay_equivalent}
It holds that
\[
    \Lcal(\thetab_t)\toinfty{t} 0
    \quad\Longleftrightarrow\quad
    \Lcal_A(t)\toinfty{t}0.
\]
\end{lemma}

\begin{proof}
The reverse implication follows immediately from
$0<\Lcal\leq\Lcal_A$.  Conversely, suppose that $\Lcal(t)\to0$.
Lemma~\ref{lemma:surrogate-loss-dissipation} makes $\Lcal_A$ nonincreasing
on a tail, so it has a limit $A_\infty\geq0$.  If $A_\infty>0$, then
\[
    w(t):=\frac{\Lcal_A(t)-\Lcal(t)}{\Lcal_A(t)}\longrightarrow1.
\]
For $U_A(t):=-\log\Lcal_A(t)$, \eqref{eq:corrected_loss_dissipation} gives for $t \geq t_A$,
\[
    \dot U_A(t)\geq\mu_0w(t).
\]
Thus $\dot U_A(t)\geq\mu_0/2$ eventually, contradicting
$U_A(t)\to-\log A_\infty<\infty$.  Hence $A_\infty=0$.
\end{proof}

The next lemma translates loss decay into radial and gradient scales.

\begin{lemma}[Asymptotic Rates]\label{lemma:freezing-tail-scales}
Under Assumptions~\ref{traj_ass:positive_margin_interpolation},~\ref{traj_ass:loss_decay}, the following hold:
\begin{align}
    U(t)&\asymp\qmin(t)\asymp\norm{\thetab_t}_2^L,
    \label{eq:log_loss_margin_parameter_scales}\\
    \norm{\g_t}_2
    &\asymp
    G(t) \asymp \norm{\thetab_t}^{L-1}\Lcal(\thetab_t),
    \label{eq:gradient_loss_scale}\\
    L\bigl(U(t)-\varphi(0)\bigr)\Lcal(\thetab_t)
    &\leq-\innerprod{\thetab_t}{\g_t}
    \lesssim U(t)\Lcal(\thetab_t).
    \label{eq:radial_gradient_loss_scale}
\end{align}
\end{lemma}

\begin{proof}
Since $\ell$ is decreasing,
\[
    e^{-\varphi(\qmin(t))}
    \leq\Lcal(\thetab_t)
    \leq m e^{-\varphi(\qmin(t))}.
\]
Loss decay therefore gives $\qmin(t)\to\infty$.  Convexity, positivity,
and boundedness of $\varphi'$ imply
$\varphi(\qmin)\asymp \qmin$ on $[t_0, \infty)$, since
\[
   0< \varphi'(\qmin(t_0))\leq \frac{\varphi(\qmin(t)) - \varphi(\qmin(t_0))}{\qmin(t)- \qmin(t_0)} \leq \sup_{\bbR} \varphi '\ .
\]
Thus $- \log \Lcal \asymp \qmin$. Denote $i_{*}  = i_*(t)\in [m]$ for which $\qmin(t) = q_{i_*}(t)$. The assumed
lower bound on $\gamma_2(\thetab_t)$, homogeneity and continuity of $f$ on the unit ball give, for all $i \in [m]$,
\begin{equation}\label{eq:qmin_homogeneous}
        \gamma_0 \norm{\thetab_t}_2^L\leq \qmin(t) = y_{i_*}f(\x_{i_*}; \thetab) \leq y_{i}f(\x_{i}; \thetab) \leq \norm{\thetab_t}_2^L \sup_{\norm{\ub} = 1} y_i f(\x_i; \ub) \ ,
\end{equation}
so $q_i \asymp \qmin \asymp \norm{\thetab_t}_2^L$, proving
\eqref{eq:log_loss_margin_parameter_scales}.

$L-1$-homogeneity of $\h_i(t) \in \partial q_i$, local Lipschitzness of $f$ (implying $\h_i$ are bounded on the unit sphere) and the Euler identity give for some $C > 0$
\[
    \norm{\mathbf h_i(t)}_2
    \leq C\norm{\thetab_t}_2^{L-1},
    \qquad
    \innerprod{\mathbf h_i(t)}{\thetab_t}=Lq_i(t).
\]
For some $C>0$, the first relation and \eqref{eq:clarke_gradient_decomposition} give
\[
    \norm{\g_t}_2
    \leq C\norm{\thetab_t}_2^{L-1}\Lcal(\thetab_t).
\]
Convexity of \(\varphi\) gives
\[
    \varphi(0)
    \geq\varphi(q_i(t))-q_i(t)\varphi'(q_i(t)).
\]
Moreover,
\(e^{-\varphi(q_i(t))}\leq\Lcal(\thetab_t)=e^{-U(t)}\),
so \(\varphi(q_i(t))\geq U(t)\).
The Euler identity and Cauchy--Schwarz therefore give
\begin{align*}
    L\bigl(U(t)-\varphi(0)\bigr)\Lcal(\thetab_t)
       &\leq L\sum_{i=1}^m
          \bigl(\varphi(q_i(t))-\varphi(0)\bigr)
          e^{-\varphi(q_i(t))}\\
       &\leq L\sum_{i=1}^m
          q_i(t)\varphi'(q_i(t))e^{-\varphi(q_i(t))}\\
       &=-\innerprod{\thetab_t}{\g_t}
        \leq\norm{\thetab_t}_2\norm{\g_t}_2.
\end{align*}
Combining this with
\(\norm{\g_t}_2\leq C\norm{\thetab_t}_2^{L-1}\Lcal(\thetab_t)\),
\(U(t)-\varphi(0)\asymp U(t)\), and
\eqref{eq:log_loss_margin_parameter_scales} proves both
\eqref{eq:gradient_loss_scale} and
\eqref{eq:radial_gradient_loss_scale}.
\end{proof}

We show that $G_A$ is an upper bound on $G$, but that their respective accumulated second moments $V_A$ and $V$ are asymptotically comparable.
\begin{lemma}[Accumulated gradient scales]
\label{lemma:accumulated-gradient-scales}
Suppose Assumptions~\ref{traj_ass:positive_margin_interpolation},~\ref{traj_ass:loss_decay} hold. Then, for all sufficiently large $t$,
\[
    G(t)\leq G_A(t).
\]
Moreover,
\begin{equation}\label{eq:accumulated_gradient_scale_comparison}
    V_A(t)\asymp V(t).
\end{equation}
\end{lemma}

\begin{proof}
Lemma~\ref{lemma:normal_corrected_loss_decay_equivalent} gives
$U_A(t)\to\infty$.
Since $U\geq U_A$ and the function $u\mapsto u^{1-1/L}e^{-u}$
is decreasing for sufficiently large $u$, we have $G\leq G_A$
eventually.

Choose $T$ sufficiently large, with $V(T),V_A(T)>0$.
The bound $G\leq G_A$ gives
\[
    V(t)
    \leq V(T)+V_A(t)-V_A(T)
    \lesssim V_A(t).
\]
For the reverse bound, since
$U_A^{1-1/L}\Lcal\leq G\leq G_A$,
$\mu_0>c_2$, and $U_A\to\infty$, $\dot \Lcal_A \leq -\mu_0(\Lcal_A - \Lcal)$ by
\eqref{eq:corrected_loss_dissipation} gives, after increasing $T$
if necessary,
\begin{align*}
    \dot G_A
    &=\left(1-\frac{1-1/L}{U_A}\right)
      U_A^{1-1/L}\dot\Lcal_A\\
    &\leq-\mu_0\left(1-\frac{1-1/L}{U_A}\right)
       \left(G_A-U_A^{1-1/L}\Lcal\right)\\
    &\leq-c_2(G_A-G).
\end{align*}

Set $g_T(u):=G(T+u) \geq 0$ for $u\geq0$. Integrating the inequality from
$T$ to $T+u$ gives
\[
G_A(T+u)
\leq e^{-c_2u}G_A(T)+\mathsf{EMA}_{c_2}[g_T](u).
\]
Lemma~\ref{lemma:Lp_bound} with $p=2$, $\lambda=c_2/2$, and $c=c_2$, together
with the triangle inequality in $L^2$, now yields
\begin{align*}
\sqrt{V_A(t)-V_A(T)}
&= \left(\int_0^{t-T} e^{c_2(T+u)}G_A^2(T+u)\,du\right)^{1/2}
\\
&\leq e^{c_2T/2}G_A(T)
       \left(\int_0^{t-T}e^{-c_2u}\,du\right)^{1/2}+e^{c_2T/2}
       \left(\int_0^{t-T}
       \left|e^{c_2u/2}\mathsf{EMA}_{c_2}[g_T](u)\right|^2\,du
       \right)^{1/2}\\
       &\leq \frac{e^{c_2T/2}G_A(T)}{\sqrt{c_2}}
       +2e^{c_2T/2}\left(\int_0^{t-T}e^{c_2u} g_T(u)^2\,du\right)^{1/2} \\
&= \frac{e^{c_2T/2}G_A(T)}{\sqrt{c_2}}
       +2\sqrt{V(t)-V(T)}.
\end{align*}
Consequently, $(a+b)^2\leq2a^2+2b^2$ gives
\[
V_A(t)\leq V_A(T)
    +\frac{2e^{c_2T}G_A(T)^2}{c_2}
    +8\bigl(V(t)-V(T)\bigr)
\lesssim V(t).
\]
\end{proof}

\begin{lemma}[Corrected Loss Asymptotics]\label{lemma:surrogate_loss_quantities}Assume~\ref{traj_ass:positive_margin_interpolation},~\ref{traj_ass:loss_decay}. There exists $C>0$ so that for all large enough $t$,
\begin{gather}
     \norm{\g_t}_2\leq Ce^{-Q_A(t)},
    \qquad
    \vb_t[j]\leq Ce^{-c_2t}V_A(t),
    \qquad
    r_A(t)^2=\frac{\dot V_A(t)}{V_A(t)},
    \label{eq:corrected_scale_relations}\\
    \dot U_A\geq
    \mu_0\frac{\Lcal_A-\Lcal}{\Lcal_A}
    =\frac{\mu_0\widehat\eta H}{2c_1\Lcal_A} \geq 0, \label{eq:U_A_derivative_lower_bound} \\
    \dot Q_A
    =\left(1-\frac{1-1/L}{U_A}\right)\dot U_A
    \geq0, \qquad  Q_A
    =U_A-\left(1-\frac1L\right)\log U_A
    \sim U_A
    \toinfty{t}\infty.\label{eq:Q_A_properties}
\end{gather}
\end{lemma}
\begin{proof}
Lemma~\ref{lemma:normal_corrected_loss_decay_equivalent} gives
$U_A(t)\to\infty$. Rearranging equation~\eqref{eq:corrected_loss_dissipation} gives~\eqref{eq:U_A_derivative_lower_bound}, and using the definition of $Q_A$ gives~\eqref{eq:Q_A_properties}. Thus $Q_A$ is eventually nondecreasing and tends to infinity.
By
Lemmas~\ref{lemma:freezing-tail-scales},~\ref{lemma:accumulated-gradient-scales}, for some $C > 0$
\[
    \norm{\g_t}_2
    \leq C G(t)
    \leq Ce^{-Q_A(t)}.
\]
The second-moment formula then gives the asserted bound on $\vb_t[j]$.
The identity for $r_A$ follows directly from the definition.
\end{proof}

Next, we show an additional desirable property of $\Lcal_A$, namely that a similar relation to~\eqref{eq:radial_gradient_loss_scale} holds between $-\angles{\thetab_t}{\m_t}$ and $U_A\Lcal_A$.
\begin{lemma}[Radial Momentum]\label{lemma:radial_momentum}
Assume~\ref{traj_ass:positive_margin_interpolation},~\ref{traj_ass:loss_decay}. There exists $a > 0, T_m > 0$ with
\begin{equation}\label{eq:radial_momentum_lower_bound}
    -\innerprod{\thetab_t}{\m_t}
    \geq aU_A(t)\Lcal_A(t)
    \qquad (t > T_m).
\end{equation}
\end{lemma}
\begin{proof}
    By Lemma~\ref{lemma:freezing-tail-scales} and using $U \geq U_A$, choose $a>0$ and $T_m$ so that
$-c_1\innerprod{\thetab_t}{\g_t}\geq a\mu_0U_A\Lcal$. Since $\mu_0>c_1$ and $U_A\to\infty$, we may increase
the tail time so that $U_A\geq\frac{2\mu_0}{\mu_0-c_1}>1$. 
Using $(U_A\Lcal_A)'=-(U_A-1)\Lcal_A\dot U_A$ and the radial
identity~\eqref{eq:common_radial_momentum_equation}, we obtain,
almost everywhere on this tail,
\begin{align*}
    &\left(\frac{d}{dt}+c_1\right)
    \left(-\innerprod{\thetab_t}{\m_t}-aU_A\Lcal_A\right)\\
    &\qquad=
    -c_1\innerprod{\thetab_t}{\g_t}
    +\widehat\eta H
    +a(U_A-1)\Lcal_A\dot U_A
    -ac_1U_A\Lcal_A\\
    &\qquad\geq
    a\mu_0U_A\Lcal
    +a\mu_0(U_A-1)(\Lcal_A-\Lcal)
    -ac_1U_A\Lcal_A\\
    &\qquad=
    a\left(
        [\mu_0(U_A-1)-c_1U_A]\Lcal_A
        +\mu_0[U_A-(U_A-1)]\Lcal
    \right)\\
    &\qquad=
    a\left(
        [(\mu_0-c_1)U_A-\mu_0]\Lcal_A
        +\mu_0\Lcal
    \right)\\
    &\qquad\geq
    \frac a2(\mu_0-c_1)U_A\Lcal_A.
\end{align*}
Where the first inequality uses the choice of $a$, discards the
nonnegative term $\widehat\eta H$, and applies
\eqref{eq:U_A_derivative_lower_bound} in the form
\[
    \Lcal_A\dot U_A\geq\mu_0(\Lcal_A-\Lcal).
\]
The RHS has an infinite $e^{c_1t}$-weighted integral. Indeed, if
$\int_T^\infty e^{c_1t}U_A\Lcal_A\,dt<\infty$, then~\eqref{eq:corrected_scale_relations}, together with $e^{-Q_A}\leq U_A\Lcal_A$, gives
$\int_T^\infty e^{c_1t}\norm{\g_t}_2\,dt<\infty$.
The first-moment formula then implies
$\norm{\m_t}_2=\bigO{e^{-c_1t}}$ . For each coordinate whose second moment
is positive at some time, there exists $a_j>0$ such that
$\vb_t[j]\geq a_j e^{-c_2t}$ on a common tail; all other coordinates
have zero velocity. Hence there exists $C>0$ such that
\[
    \norm{\dot\thetab_t}_2
    \leq C(1+t)^\alpha e^{-(c_1-c_2/2)t},
\]
which is integrable, contradicting $\norm{\thetab_t}_2\to\infty$.
Multiplying the preceding differential inequality by $e^{c_1t}$
and integrating from $T$ to $t$ gives
\begin{align*}
    &e^{c_1t}
    \left(-\innerprod{\thetab_t}{\m_t}
          -aU_A(t)\Lcal_A(t)\right)\\
    &\quad\geq
    e^{c_1T}
    \left(-\innerprod{\thetab_T}{\m_T}
          -aU_A(T)\Lcal_A(T)\right)
    +\frac a2(\mu_0-c_1)
      \int_T^t e^{c_1s}U_A(s)\Lcal_A(s)\,ds.
\end{align*}
The initial term is a fixed finite number, whereas the integral
tends to infinity. Since $\mu_0>c_1$, the right-hand side is
therefore positive for all sufficiently large $t$.
Dividing by $e^{c_1t}>0$ yields~\eqref{eq:radial_momentum_lower_bound}.
\end{proof}

The following lemma uses the estimates produced so far to reveal a negative feedback mechanism forcing $r_A$ to be small and $Q_A$ to be large: large $r_A$, corresponding to a momentary large \enquote{corrected gradient} $G_A$, would force growth of $Q_A$ and hence suppress the growth of $r_A$.

\begin{lemma}[Lower-bound feedback]
\label{lemma:adam_lower_bound_feedback}
Suppose assumptions~\ref{traj_ass:positive_margin_interpolation},~\ref{traj_ass:loss_decay} hold. 
Then, for all sufficiently large $t$, $Q_A$ is nondecreasing, $Q_A(t)\to\infty$, and there are constants
$k_{Q_A}>0$ and $\delta>c_2/2$ such that
\begin{gather}
    \frac{d}{dt}\log r_A(t)
    =\frac{c_2}{2}-\dot Q_A(t)-\frac12r_A(t)^2,
    \label{eq:r_A_log_derivative}\\
    \dot Q_A(t)\geq
    \min\left\{\delta,\,
      k_{Q_A}(1+t)^\alpha r_A(t)Q_A(t)^{1-\frac1L}\right\}.
    \label{eq:Q_A_growth_lower_bound}
\end{gather}
The differential relations hold almost everywhere.
\end{lemma}

\begin{proof}
~\eqref{eq:r_A_log_derivative} follows from differentiating the definition of $r_A$. We proceed to prove~\eqref{eq:Q_A_growth_lower_bound}. First consider times when $\Lcal < \frac{\Lcal_A}{2}$. For such times, \eqref{eq:U_A_derivative_lower_bound} gives
$\dot U_A\geq\mu_0/2$. Since $\mu_0>c_1\geq c_2$ and
using~\eqref{eq:Q_A_properties}, increasing the tail time gives
$\dot Q_A\geq\delta$ for some $\delta>c_2/2$.

At times when $\Lcal\geq\Lcal_A/2$, we have $U=U_A+\bigO{1}$ and
$\norm{\thetab_t}_2^L\asymp U_A$. Lemma~\ref{lemma:radial_momentum} and Cauchy--Schwarz imply that there exists $C>0$ such that
\[
    a^2U_A^2\Lcal_A^2
    \leq\innerprod{\thetab_t}{\m_t}^2 = \innerprod{\thetab_t\vb_t^{1/4}}{\frac{\m_t}{\vb_t^{1/4}}}^2
    \leq\left(\sum_{j=1}^p\thetab_t[j]^2\sqrt{\vb_t[j]}\right)H
    \leq C U_A^{2/L}e^{-c_2t/2}\sqrt {V_A}\,H.
\]
By the definitions of $r_A$, $V_A$, and $G_A$,
\[
    r_A
    =\sqrt{\frac{\dot V_A}{V_A}}
    =\frac{e^{c_2t/2}G_A}{\sqrt{V_A}}
    =U_A^{1-\frac1L}
      \frac{e^{c_2t/2}\Lcal_A}{\sqrt{V_A}}.
\]
Substituting this identity into the preceding bound yields
\[
    \frac{H}{\Lcal_A}
    \geq\frac{a^2}{C}U_A^{1-\frac1L}r_A.
\]
Since $U_A\to\infty$, we may increase the tail time so that
\(
    1-\frac{1-1/L}{U_A}\geq\frac12.
\)
Together with $\dot U_A\geq0$ and~\eqref{eq:U_A_derivative_lower_bound},
this shows that there exists $k_{Q_A}>0$ such that, almost everywhere at the times under consideration,
\begin{align*}
    \dot Q_A
    &=\left(1-\frac{1-1/L}{U_A}\right)\dot U_A\geq\frac12\dot U_A\geq\frac{\mu_0}{4c_1}
          \widehat\eta\,\frac{H}{\Lcal_A}\geq\frac{a^2\mu_0}{4c_1C}
          \widehat\eta\,U_A^{1-\frac1L}r_A\\
    &\geq k_{Q_A}(1+t)^\alpha r_AQ_A^{1-\frac1L},
\end{align*}
where the last inequality uses
$\widehat\eta\asymp(1+t)^\alpha$ and $Q_A\asymp U_A$,
with the fixed positive constants absorbed into $k_{Q_A}>0$.\end{proof}

\begin{lemma}[Upper-bound feedback]\label{lemma:adam-speed-estimates}
Suppose assumptions~\ref{traj_ass:positive_margin_interpolation},~\ref{traj_ass:loss_decay} hold, and set
\[
    s_A(t):=\norm{\frac{\m_t}{\sqrt{\vb_t}}}_2,
    \qquad \nu:=1-\frac1L,\qquad
    \zeta:=\alpha+\nu,\qquad \lambda:=c_1-\frac{c_2}{2}>0,
\]
where coordinates with zero second moment contribute zero.
On a sufficiently late tail, $s_A$ is locally absolutely continuous,
$s_A=\bigO{1}$, and there exists $C > 0$ with
\begin{align}
    \dot s_A
    &\leq-\lambda s_A+C e^{c_2t/2}\norm{\g_t}_2
     \leq-\lambda s_A+C e^{c_2t/2-Q},
    \label{eq:adam_ratio_differential_bound}\\
    \abs{\frac{d}{dt}Q^{1/L}}&\leq C\eta s_A.
    \label{eq:Q_root_derivative_bound}
\end{align}
The differential inequalities hold almost everywhere. 
\end{lemma}

\begin{proof}
Coordinates with identically zero second moment have zero velocity
and can be omitted. For each remaining coordinate, the second-moment
formula gives $v_j(t)\geq a_j e^{-c_2t}$ eventually, with $a_j>0$.
Since there are finitely many coordinates, these bounds hold on a common tail.
The coordinate ratios, and hence $s_A$, are locally absolutely continuous
on that tail. Lemma~\ref{lemma:adam_universal_ratio} and the bounded
bias-correction factors imply that there exists $C>0$ such that $s_A\leq C$.
The moment equations and Cauchy--Schwarz give
\begin{align*}
    \frac12\frac{d}{dt}s_A^2
    &=c_1\sum_j\frac{m_jg_j}{v_j}-\lambda s_A^2
        -\frac{c_2}{2}\sum_j\frac{m_j^2g_j^2}{v_j^2}\\
    &\leq c_1s_A\norm{\frac{\g_t}{\sqrt{\vb_t}}}_2-\lambda s_A^2.
\end{align*}
Dividing by $s_A$ where it is positive, and using $\dot s_A=0$
almost everywhere on $\{s_A=0\}$, gives the first inequality
in~\eqref{eq:adam_ratio_differential_bound}.
The second follows from Lemma~\ref{lemma:freezing-tail-scales}
and the identity $G=U^\nu\Lcal=e^{-Q}$.
For some $C>0$, the loss chain rule and the same gradient-scale estimate give
\begin{equation}\label{eq:dot_U_upper_bound}
        \abs{\dot U}
    =\widehat\eta\,
      \frac{\abs{\innerprod{\g_t}{\m_t/\sqrt{\vb_t}}}}{\Lcal(\thetab_t)}
    \leq C\eta U^\nu s_A.
\end{equation}
Since $Q=U-\nu\log U\sim U>0$ and $U\to\infty$, the chain rule gives,
on a sufficiently late tail,
\[
    \abs{\frac{d}{dt}Q^{1/L}}
    =\frac{Q^{-\nu}}{L}
      \left(1-\frac{\nu}{U}\right)\abs{\dot U}
    \leq C\eta\left(\frac{U}{Q}\right)^\nu s_A
    \leq C\eta s_A.
\]
This proves~\eqref{eq:Q_root_derivative_bound}.
\end{proof}

\begin{lemma}[Coarse Decay Bounds]\label{lemma:coarse_lower_bounds}Assume~\ref{traj_ass:positive_margin_interpolation},~\ref{traj_ass:loss_decay}. Then $r_A$ is bounded on a tail. If $\zeta \geq 0$, then
\begin{equation}\label{eq:coarse_Q_A_r_A_bounds}
    Q(t) \geq Q_A \gtrsim t, \qquad r_A(t) \lesssim t^{-\zeta}.
\end{equation}
\end{lemma}
\begin{proof}
Write $\nu=1-1/L$, and observe a tail where preceding lemmas hold. Since $\dot Q_A\geq0$,
\eqref{eq:r_A_log_derivative} gives
\[
    \frac{d\log r_A}{dt}\leq \frac{c_2}{2}-\frac12r_A^2,
\]
so $r_A$ is bounded by a simple barrier argument.

Suppose $\zeta\geq0$. Boundedness of $r_A$ and
\eqref{eq:Q_A_growth_lower_bound}, considering the two entries in
the minimum, imply that there exists $C>0$ such that
\[
    r_A^2
    \leq C\left(1+(1+t)^{-\alpha}Q_A^{-\nu}\right)\dot Q_A.
\]
Indeed, let $M$ be an upper bound for $r_A$. If $k_{Q_A}(1+t)^\alpha r_AQ_A^\nu\geq\delta$, then $\dot Q_A\geq\delta$, so $r_A^2\leq M^2\leq (M^2/\delta)\dot Q_A$. Otherwise, $\dot Q_A\geq k_{Q_A}(1+t)^\alpha r_AQ_A^\nu$, giving $r_A^2\leq Mr_A\leq (M/k_{Q_A})(1+t)^{-\alpha}Q_A^{-\nu}\dot Q_A$.

For $T\leq s\leq t$, the inequality $\alpha\geq-\nu$ implies
$(1+s)^{-\alpha}\leq(1+t)^\nu$. Integrating the preceding bound
and using~\eqref{eq:r_A_log_derivative} therefore yields for some $C > 0$
\[
    \frac{c_2}{2}t-C
    \leq Q_A(t)+\frac12\int_T^t r_A(s)^2\,ds
    \leq C Q_A(t)+C(1+t)^\nu Q_A(t)^{1-\nu}.
\]
Dividing by $1+t$ shows that $Q_A(t)/(1+t)$ is bounded away
from zero. Together with $G\leq G_A$, this proves $Q\geq Q_A\gtrsim t$.

Finally, set $\tilde r_A(t)=(1+t)^\zeta r_A(t)$. The lower bound on $Q_A$
and Lemma~\ref{lemma:adam_lower_bound_feedback} give for some $c > 0$
\[
    \frac{d\log \tilde r_A}{dt}
    \leq \frac{c_2}{2}+\frac{\zeta}{1+t}-\min\{\delta,c\tilde r_A\}.
\]
Since $\delta>c_2/2$, this is negative on a sufficiently late tail
whenever $\tilde r_A$ exceeds a fixed constant. Thus $\tilde r_A$ is bounded,
proving $r_A\lesssim t^{-\zeta}$.
\end{proof}

\begin{corollary}[Second-moment freezing]\label{cor:second_moment_freezing}
    Assume~\ref{traj_ass:positive_margin_interpolation},~\ref{traj_ass:loss_decay}. If $\zeta > \frac 12$ then $V,V_A$ are bounded, and 
    \[
    \forall j \in [p]: e^{c_2t}\vb_t[j] \toinfty{t} \omega_j < \infty, \qquad \omega_j-c_2\int_0^t e^{c_2s}g_j(s)^2\,ds \leq \bigO{t^{-2\zeta+1}} \ .
    \]
\end{corollary}
\begin{proof}
Lemma~\ref{lemma:coarse_lower_bounds} gives
\[
    \log\frac{V_A(t)}{V_A(T)}
    =\int_T^t r_A(s)^2\,ds
    \lesssim\int_T^\infty s^{-2\zeta}\,ds<\infty.
\]
Thus $V_A$ is bounded, and so is $V$ by
Lemma~\ref{lemma:accumulated-gradient-scales}. Since $e^{c_2t}\vb_t[j]$ is nondecreasing and $e^{c_2t}\vb_t[j]\lesssim  V(t)$, it has a finite limit. Moreover, again due to Lemma~\ref{lemma:accumulated-gradient-scales}, $e^{c_2t}\g_t[j]^2 \lesssim e^{c_2t}G_A(t)^2 \lesssim V_A(t)r_A(t)^2 \lesssim t^{-2\zeta}$, so there exists $C>0$ such that
\begin{align*}
    0
    &\leq
    \omega_j-c_2\int_0^t e^{c_2s}g_j(s)^2\,ds = c_2\int_t^\infty e^{c_2s}\g_s[j]^2\,ds \leq C\int_t^\infty(1+s)^{-2\zeta}\,ds =\bigO{(1+t)^{-2\zeta+1}}\\
    &=\bigO{
       (1+t)^{-2\left(\alpha-\left(\frac1L-\frac12\right)\right)}
       }.
\end{align*}
\end{proof}

On the flip side, we can show that in the intermediate and slow regimes the second (and first) moments diverge:
\begin{proposition}[No moment freezing in the intermediate and slow regimes]
\label{prop:no-scalar-freezing}
Suppose Assumptions~\ref{traj_ass:positive_margin_interpolation},~\ref{traj_ass:loss_decay} hold and $-1 < \alpha\leq1/L-1/2$.  Then
\begin{equation}
    \int_0^\infty e^{c_2s}\norm{\g_s}_1^2\,ds=\infty, \qquad \int_0^\infty e^{c_1s}\norm{\g_s}_1\,ds=\infty~.
    \label{eq:weighted_gradient_integrals_diverge}
\end{equation}
\end{proposition}

\begin{proof}
Assume, toward a contradiction, that
\begin{equation}\label{eq:finite_weighted_gradient_square_integral}
    \int_0^\infty e^{c_2s}\norm{\g_s}_1^2\,ds<\infty.
\end{equation}
There exist $T > 0, \{a_j\}_{j \in [p]}$ so that for any $j$ with $\omega_j > 0$, it holds that $\forall t \geq T:\sqrt{\vb_t[j]} \geq a_je^{-\frac{c_2}{2}t}$, thus
\begin{equation}\label{eq:adam_ratio_square_integrable}
\int_T^\infty \frac{{\m_s[j]}^2}{\vb_s[j]}\,ds
 \leq  a_j^{-2}\int_T^\infty e^{c_2s}\m_s[j]^2\,ds \leq a_j^{-2}\left(\frac{c_1}{c_1-\frac{c_2}2}\right)^2 \int_0^\infty e^{c_2s}\g_s[j]^2 \, ds < \infty \ .   
\end{equation}
Where the last inequality is~\eqref{eq:finite_weighted_gradient_square_integral}, and the second-to-last is Lemma~\ref{lemma:Lp_bound} with $p=2,c=c_1,\lambda=\frac{c_2}2$.
 Therefore, integrating the Adam equation, applying Cauchy--Schwarz and \eqref{eq:adam_ratio_square_integrable},
\begin{align}
    \norm{\thetab_t}_2
    &\leq\norm{\thetab_T}_2
      +\left(\int_T^t\widehat\eta(s)^2\,ds\right)^{1/2}
       \left(\int_T^\infty
       \norm{\frac{\m_s}{\sqrt{\vb_s}}}_2^2\,ds\right)^{1/2} \leq \bigO{\left(\int_T^t\widehat\eta(s)^2\,ds\right)^{1/2}}.
    \label{eq:theta_CS}
\end{align}
Where,
\begin{equation}
    \left(\int_T^t\widehat\eta(s)^2\,ds\right)^{1/2}
    =
    \begin{cases}
      \bigO{t^{\alpha+1/2}},&\alpha>-1/2,\\
      \bigO{\sqrt{\log t}},&\alpha=-1/2,\\
      \bigO{1},&\alpha<-1/2.
    \end{cases}
    \label{eq:eta_L2_growth}
\end{equation}
If $\alpha < \frac1L - \frac12$, the above bounds imply
\begin{equation}
    \norm{\thetab_t}_2= \smallo{t^{1/L}}.
    \label{eq:theta_little_o_growth}
\end{equation}

For the endpoint $\alpha = \frac1L - \frac12$, fix $S\geq T$ sufficiently large. Applying Cauchy--Schwarz as in~\eqref{eq:theta_CS} shows that there exists $C>0$, independent of $S$ and $t\geq S$, such that
\begin{align*}
    \norm{\thetab_t}_2
    &\leq\norm{\thetab_S}_2
      +\left(\int_S^t\widehat\eta(s)^2\,ds\right)^{1/2}
       \left(\int_S^\infty
       \norm{\frac{\m_s}{\sqrt{\vb_s}}}_2^2\,ds\right)^{1/2} \\
    &\leq\norm{\thetab_S}_2
      +Ct^{1/L}\left(\int_S^\infty
       \norm{\frac{\m_s}{\sqrt{\vb_s}}}_2^2\,ds\right)^{1/2},
\end{align*}
where $C>0$ is independent of $S$ and $t\geq S$, since $\widehat\eta(s)^2=\bigO{s^{2/L-1}}$. Dividing by $t^{1/L}$ and taking the limit superior as $t\to\infty$ gives
\[
    \limsup_{t\to\infty}\frac{\norm{\thetab_t}_2}{t^{1/L}}
    \leq C\left(\int_S^\infty
       \norm{\frac{\m_s}{\sqrt{\vb_s}}}_2^2\,ds\right)^{1/2}.
\]
Letting $S\to\infty$ and using~\eqref{eq:adam_ratio_square_integrable}, we obtain~\eqref{eq:theta_little_o_growth} also at the endpoint.

This further implies 
$$q_i(\thetab_t) = \smallo{t}, \qquad U(t) = - \log \Lcal(\thetab_t) = \smallo{t} \ .$$ 
But this is a contradiction to \eqref{eq:finite_weighted_gradient_square_integral}, since this implies divergence of $e^{c_2t}\norm{\g_t}_1^2$:
\begin{equation}
    e^{c_2t}\norm{\g_t}_1^2
    \gtrsim\exp\left(c_2t-2U(t)\right)
    \longrightarrow\infty \ .
    \label{eq:weighted_gradient_explodes}
\end{equation}
As for the first moment, recall that $\Lcal \to 0$ implies $\norm{\g_t}_1 \to 0$. Also, $c_1 \geq c_2$. Together this implies for large enough $t$,
\[
e^{c_1t}\norm{\g_t}_1 \geq e^{c_2t}\norm{\g_t}_1^2 \ ,
\]
hence $\int_0^\infty e^{c_1s}\norm{\g_s}_1\,ds = \infty$ by comparison.
\end{proof}

\begin{corollary}[Tracking and Speed Estimates]\label{cor:tracking_and_speed_estimates}
Assume~\ref{traj_ass:positive_margin_interpolation},~\ref{traj_ass:loss_decay} and $\zeta \geq 0$ and fix $j \in [p]$. If $\zeta > \frac12$ or if $\frac{\abs{\g_t[j]}}{\norm{\g_t}_1}$ is eventually bounded away from 0, then
\begin{equation}\label{eq:adam_speed_tracking_estimates}
    {\frac{\abs{\m_t[j]}}{\sqrt{\vb_t[j]}}} \leq \bigO{t^{-\zeta}}, \qquad \frac{\g_t^2[j]}{\vb_t[j]}\leq \bigO{t^{-2\zeta}} \ .
\end{equation}
\end{corollary}
\begin{proof}
Coordinates with identically zero second moment are immediate under the convention.
Otherwise, set $V_j(t):=e^{c_2t}\vb_t[j]$ and choose $T$
sufficiently large that $V_j(T)>0$.

We first show that $V_A/V_j$ is bounded. If $\zeta>1/2$, this
follows from Corollary~\ref{cor:second_moment_freezing} and
$V_j(t)\geq V_j(T)>0$.
Under the second condition, Lemma~\ref{lemma:freezing-tail-scales}
gives $\abs{\g_t[j]}\gtrsim G(t)$, so for some $c > 0$
\[
    V_j(t)
    =V_j(T)+c_2\int_T^t e^{c_2s}\g_s[j]^2\,ds
    \geq V_j(T)+c\bigl(V(t)-V(T)\bigr)
    \gtrsim V(t)\asymp V_A(t).
\]

In either case, the gradient-scale estimate~\eqref{eq:corrected_scale_relations} gives
\[
    \frac{\g_t[j]^2}{\vb_t[j]}
    \lesssim
    \frac{e^{c_2t}G_A(t)^2}{V_j(t)}
    =\frac{V_A(t)}{V_j(t)}r_A(t)^2
    \lesssim t^{-2\zeta}.
\]
The last inequality follows from
Lemma~\ref{lemma:coarse_lower_bounds}, using boundedness of $r_A$
when $\zeta=0$.

Next, $z_j:=\abs{\m_t[j]}/\sqrt{\vb_t[j]}$ is locally
absolutely continuous on this tail. Lemma~\ref{lemma:time_derivative_uncorrected_adam_ratio} implies that there exists $C>0$ such that,
almost everywhere,
\[
    \dot z_j
    \leq-\lambda z_j
       +c_1\frac{\abs{\g_t[j]}}{\sqrt{\vb_t[j]}}
    \leq-\lambda z_j+Ct^{-\zeta},
    \qquad \lambda=c_1-\frac{c_2}{2}>0.
\]
Integrating yields
\[
    z_j(t)
    \leq e^{-\lambda(t-T)}z_j(T)
       +C\int_T^t e^{-\lambda(t-s)}s^{-\zeta}\,ds
    \lesssim t^{-\zeta},
\]
as required. 
\end{proof}

\begin{theorem}[Loss decay rate]\label{thm:loss_decay_rates_appendix}
    Let $f$ satisfy~\ref{model_ass:1_locally_lipschitz},~\ref{model_ass:2_homogeneous}, let $\ell \in \mathrm L_{\e}$ and let $\thetab_t$ be an Adam trajectory of $f$ assuming ~\ref{traj_ass:positive_margin_interpolation},~\ref{traj_ass:loss_decay}. Writing $U(t):=-\log\Lcal(\thetab_t)$ and $\nu:=1-1/L$, we have
    \begin{equation}
    \begin{alignedat}{2}
        U(t)
        &=\frac{c_2}{2}t+(\zeta+\nu)\log t\pm\bigO{1},
        &\qquad& \zeta>\frac12,\\
        \frac{c_2}{2}t-\bigO{\log t}\leq U(t)
        &\leq\frac{c_2}{2}t+\bigO{\sqrt t},
        && \zeta=\frac12,\\
        \frac{c_2}{2}t-\bigO{t^{1-2\zeta}}\leq U(t)
        &\leq\frac{c_2}{2}t+\bigO{t^\zeta},
        && 0<\zeta<\frac12,\\
        \Omega(t)\leq U(t)
        &\leq\frac{c_2}{2}t+\nu\log t+\bigO{1},
        && \zeta=0, \\
        U(t) &= \Theta\left(t^{L(\alpha+1)}\right), && \zeta < 0 \ .
    \end{alignedat}
    \end{equation}
Here asymptotic notation hides constants which may depend on hyperparameters, data or initialization (but not on $t$).
\end{theorem}
\begin{proof}
First suppose $\zeta<0$. Lemma~\ref{lemma:adam-speed-estimates}
and $Q\sim U$ give $U(t)=\bigO{t^{L(\alpha+1)}}$.
Equation~\eqref{eq:dot_U_upper_bound} gives
\[
    \abs{\dot U}\lesssim\eta U^\nu
    =\bigO{t^{L(\alpha+1)-1}}=\smallo{1}.
\]
Consequently,
\[
    \frac{d}{dt}\log G
    =\left(\frac{\nu}{U}-1\right)\dot U=\smallo{1},
    \qquad
    \frac{d}{dt}\left(e^{c_2t}G^2\right)
    =\left(c_2+\smallo{1}\right)\dot V.
\]
Since $U=\smallo{t}$, we have $e^{c_2t}G^2\to\infty$;
integration gives $e^{c_2t}G^2\sim c_2V$.

Lemma~\ref{lemma:accumulated-gradient-scales} now yields
\[
    r_A^2=\frac{e^{c_2t}G_A^2}{V_A}
    \gtrsim\frac{e^{c_2t}G^2}{V}\gtrsim1.
\]
Since $\eta Q_A^\nu\leq\eta U^\nu=\smallo{1}$,
\eqref{eq:Q_A_growth_lower_bound} implies
\[
    \dot Q_A\gtrsim\eta Q_A^\nu,
    \qquad \frac{d}{dt}Q_A^{1/L}\gtrsim\eta.
\]
Integrating gives $U(t)\geq Q_A(t)\gtrsim t^{L(\alpha+1)}$,
which proves the slow-regime estimate.

For the remaining cases, suppose $\zeta\geq0$.

We use the feedback mechanisms to define barrier variables on both sides. We first phrase all bounds using $Q$, and conclude with the relation between $Q$ and $U= -\log \Lcal$.
Recall $\nu = 1 - \frac 1L, \zeta = \alpha + \nu$, and set
\[
    B(t):=\frac{c_2}{2}t+\zeta\log(1+t).
\]
We first obtain the lower bound from the coarse estimates,
then construct an upper barrier.

Define
\[
    b(t):=\frac12\log V_A(t),
    \qquad Q_-(t):=Q_A(t)+b(t).
\]
The identity $r_A=e^{c_2t/2-Q_A}/\sqrt{V_A}$ and
Lemma~\ref{lemma:coarse_lower_bounds} imply that there exists $C>0$ such that
\[
    Q_-(t)=\frac{c_2}{2}t-\log r_A(t)\geq B(t)-C,
\]
and
\[
    Q(t)\geq Q_A(t)=Q_-(t)-b(t)\geq B(t)-b(t)-C.
\]
Moreover, $\dot b=r_A^2/2$, and by Lemma~\ref{lemma:coarse_lower_bounds}, $r_A =\bigO{t^{-\zeta}}$ whenever $\zeta > 0$, so $b(t)=\bigO{t^{1-2\zeta}}$ whenever $\frac12 >\zeta > 0$ and $b(t) = \bigO{\log t}$ when $\zeta = \frac12$.
If $\zeta>1/2$, then $b(t)=\bigO{1}$ by
Corollary~\ref{cor:second_moment_freezing}. When $\zeta=0$, $Q(t)\geq Q_A(t)\gtrsim t$ by Lemma~\ref{lemma:coarse_lower_bounds}.
This finishes the lower bound.

For the upper bound, there exists $C_s>0$ such that, on a common tail,
\[
    s_A(t)\leq S(t):=
    \begin{cases}
        C_s,&0\leq\zeta\leq1/2,\\
        C_s(1+t)^{-\zeta},&\zeta>1/2.
    \end{cases}
\]
The first case follows from Lemma~\ref{lemma:adam-speed-estimates}.
The second follows by applying
Corollary~\ref{cor:tracking_and_speed_estimates} to every coordinate.

Let $C_0$ be the constant in~\eqref{eq:Q_root_derivative_bound},
write $\lambda=c_1-c_2/2$, and fix $A>C_0/\lambda$.
Define the corrected upper variable
\[
    Q_+(t):=
    \left(Q(t)^{1/L}+A\eta(t)s_A(t)\right)^L.
\]
By the upper-feedback inequalities, there exists $C_A>0$ such that, almost everywhere on a
sufficiently late tail,
\begin{align*}
    \frac{d}{dt}Q_+^{1/L}
    &\leq
    \left(C_0-A\lambda+\frac{A\alpha}{1+t}\right)\eta s_A
       +C_A\eta e^{c_2t/2-Q}\\
    &\leq C_A\eta e^{c_2t/2-Q}.
\end{align*}

For a constant $D>0$ to be chosen, define
\[
    B_+(t):=
    \left((B(t)+D)^{1/L}+A\eta(t)S(t)\right)^L.
\]
The explicit choice of $S$ gives, in both regimes,
\[
    \eta S=\smallo{t^{1/L}},
    \qquad
    \frac{d}{dt}(\eta S)=\smallo{t^{-\nu}}.
\]
It follows that
\[
    \frac{d}{dt}B_+^{1/L}
    =
    \left(\frac{(c_2/2)^{1/L}}{L}+\smallo{1}\right)(1+t)^{-\nu}.
\]
Whenever $Q_+\geq B_+$, the inequality $s_A\leq S$ implies
$Q\geq B+D$. Hence, at such times,
\[
    \frac{d}{dt}Q_+^{1/L}
    \leq C_A\eta e^{c_2t/2-B-D}
    =C_A\eta_0e^{-D}(1+t)^{-\nu}.
\]
Choose $D$ sufficiently large. There exists $c>0$ such that, on a sufficiently late tail,
\[
    \frac{d}{dt}\left(Q_+^{1/L}-B_+^{1/L}\right)
    \leq-c(1+t)^{-\nu}
    \qquad\text{whenever }Q_+\geq B_+
\]
Since $\nu<1$, the integral of $(1+t)^{-\nu}$ diverges.
Thus $Q_+$ eventually enters and remains below $B_+$, so $Q\leq Q_+\leq B_+$ on a sufficiently
late tail. Since $B(t)+D\asymp t$ and $\eta S=\smallo{t^{1/L}}$,
we have
\[
    B_+(t)^{1/L}=(B(t)+D)^{1/L}+A\eta(t)S(t)\asymp t^{1/L}.
\]
For any fixed $t$, the mean value theorem on the interval $\left[(B(t)+D)^{1/L},(B(t)+D)^{1/L}+A\eta(t)S(t)\right]$, using $L\geq1$, gives, for some $C>0$,
\begin{align*}
    0\leq B_+(t)-(B(t)+D)
    &\leq LA\eta(t)S(t)
      \left((B(t)+D)^{1/L}+A\eta(t)S(t)\right)^{L-1}\\
    &\leq Ct^{1-1/L}\eta(t)S(t)
     =Ct^\nu\eta(t)S(t).
\end{align*}
If $0\leq\zeta\leq1/2$, then $S=C_s$, so
\[
    t^\nu\eta(t)S(t)
    =\eta_0 C_s t^\nu(1+t)^\alpha
    =\bigO{t^{\nu+\alpha}}
    =\bigO{t^\zeta}.
\]
If $\zeta>1/2$, then $S=C_s(1+t)^{-\zeta}$, and
$\alpha-\zeta=-\nu$ gives
\[
    t^\nu\eta(t)S(t)
    =\eta_0 C_s t^\nu(1+t)^{\alpha-\zeta}
    =\bigO{1}.
\]
Consequently, absorbing the fixed constant $D$ into the error terms,
\[
    Q(t)\leq B_+(t)
    =
    \begin{cases}
        B(t)+\bigO{t^\zeta},&0\leq\zeta\leq1/2,\\
        B(t)+\bigO{1},&\zeta>1/2.
    \end{cases}
\]

Finally, $Q=U-\nu\log U$ and $U\to\infty$ imply $Q/U\to1$.
Therefore
\[
U=Q+\nu\log\left(Q(1+\smallo{1})\right)
 =Q+\nu\log Q+\smallo{1}.
\]
The lower and upper bounds give $Q(t)\asymp t$, so
$\log Q(t)=\log t+\bigO{1}$.
If $\zeta>1/2$, this yields
\[
    U(t)=\frac{c_2}{2}t+(\zeta+\nu)\log t+\bigO{1}.
\]
For $0<\zeta\leq1/2$, the logarithmic terms are absorbed into
$\bigO{t^\zeta}$ in the upper bound. When $\zeta=0$, the bounds $Q(t)\gtrsim t$ and
$Q(t)\leq(c_2/2)t+\bigO{1}$ give
\[
    U(t)\gtrsim t,
    \qquad
    U(t)\leq\frac{c_2}{2}t+\nu\log t+\bigO{1}.
\]
This completes the estimates.
\end{proof}

The following lemma finds use specifically for the width-$1$ construction.
\begin{lemma}[Uniform early-phase dominance of the second moment]
\label{lemma:second_moment_early_dominance}
Fix $L\geq1$, $\alpha\in\mathbb R$ with
$\alpha>\frac1L-\frac12$, and $c_2,k,K>0$.
There exist constants $\varepsilon_0,C>0$, depending only on
$L,\alpha,c_2,k,K,\eta_0$, with the following property.
Fix $j\in[p]$ and write
\[
    \nu:=1-\frac1L,
    \qquad
    \zeta:=\alpha+\nu,
    \qquad
    I_j(t):=e^{c_2t}v_j(t).
\]
Assume that:
\begin{enumerate}
    \item Every gradient coordinate has a fixed sign, and
    $t\mapsto |g_i(t)|$ is nonincreasing on $[0,\infty)$
    for every $i\in[p]$.
    \item The following bounds hold:
    \[
        U(1/2)\geq\max\{1,4\nu\},
        \qquad
        kG(t)\leq|g_j(t)|\leq KG(t)
        \quad\text{for almost every }t\geq1/2.
    \]
\end{enumerate}

Then, if $L=1$, or if $L>1$ and $I_j(1)\leq\varepsilon_0$, 
\[
    \omega_j:=\lim_{t\to\infty}I_j(t)
    \leq CI_j(1)<\infty,
\]
and
\begin{equation}
\label{eq:moment_log_derivative_decay}
    \frac{\dot I_j(t)}{I_j(t)}
    \leq C(1+t)^{-2\zeta}
    \qquad\text{for almost every }t\geq1.
\end{equation}
Moreover, when $L>1$,
\[
    \omega_j=I_j(1)(1+\smallo{1})
    \qquad\text{as }I_j(1)\to0,
\]
uniformly over all trajectories satisfying these assumptions.

One also has
\[
    0\leq\omega_j-I_j(t)
    \leq CI_j(1)(1+t)^{1-2\zeta},
    \qquad t\geq1.
\]
\end{lemma}

\begin{proof}
All constants below depend only on $L,\alpha,c_2,k,K,\eta_0$.
The coordinatewise gradient assumptions and zero initial momentum
give $\forall i\in[p]:\abs{\hat m_i(t)}\geq\abs{g_i(t)}$,
therefore $g_i(t)\hat m_i(t)\geq g_i(t)^2$. Consequently,
\[
\begin{aligned}
    \dot U(t)
    &=\frac{\angles{\g_t}
        {\eta(t)\frac{\hat\m_t}{\sqrt{\hat\vb_t}}}}
        {\Lcal(\thetab_t)}\\
    &\geq
    \frac{\eta(t)g_j(t)\hat m_j(t)}
         {\Lcal(\thetab_t)\sqrt{\hat v_j(t)}}\\
    &\geq
    \eta(t)\sqrt{1-e^{-c_2t}}\,e^{c_2t/2}
    \frac{g_j(t)^2}{\Lcal(\thetab_t)\sqrt{I_j(t)}}.
\end{aligned}
\]
In particular, $U$ is nondecreasing, so
$U(t)\geq\max\{1,4\nu\}$ for $t\geq1/2$.
The gradient lower bound
$g_j(t)^2\geq k^2G^2(t)=k^2\Lcal^2U^{2\nu}$
and $e^U=\Lcal^{-1}$ therefore give, for some $c>0$,
\[
    \frac{d}{dt}\left(\frac{e^{U(t)}}{U(t)^{2\nu}}\right)
    =
    \frac{e^{U(t)}}{U(t)^{2\nu}}
    \left(1-\frac{2\nu}{U(t)}\right)\dot U(t)
    \geq
    c\eta(t)\frac{e^{c_2t/2}}{\sqrt{I_j(t)}}.
\]
For $t\geq1$, integrate over $[t-\frac12,t]$. Since $I_j$
is nondecreasing and $\eta(s)\asymp\eta(t)$ on this interval,
for some $c>0$, we obtain
\begin{equation}
\label{eq:loss_moment_window_bound}
    \frac{e^{U(t)}}{U(t)^{2\nu}}
    \geq
    c\eta(t)\frac{e^{c_2t/2}}{\sqrt{I_j(t)}}.
\end{equation}
The second-moment equation and the gradient upper bound give
\[
    \dot I_j(t)
    =c_2e^{c_2t}g_j(t)^2
    \leq c_2K^2e^{c_2t}U(t)^{2\nu}e^{-2U(t)}.
\]
Combining this with \eqref{eq:loss_moment_window_bound} shows that there exists $C>0$ such that
\begin{equation}
\label{eq:moment_log_derivative_bound}
    \frac{\dot I_j(t)}{I_j(t)}
    \leq C(1+t)^{-2\alpha}U(t)^{-2\nu},
    \qquad t\geq1.
\end{equation}
If $L=1$, then $\nu=0$ and $\alpha>1/2$, so this estimate
and its integral prove the logarithmic growth and multiplicative
bounds.

Suppose $L>1$, and take logarithms in
\eqref{eq:loss_moment_window_bound}. For some $C>0$, the bounds $U\geq1$ and
$\alpha\log(1+t)\geq-c_2t/4-C$ give, for all $t\geq1$,
\[
    U(t)
    \geq\frac{c_2}{4}t-\frac12\log I_j(t)-C.
\]
Consequently, as long as $I_j(t)\leq2I_j(1)$,
a sufficiently large $B:=\log(1/I_j(1))$ ensures for some $c > 0$
\begin{equation}\label{eq:U_lower_bound_t_plus_B}
    U(t)\geq c(t+B).
\end{equation}
On any such interval, \eqref{eq:moment_log_derivative_bound} implies for some $C > 0$
\begin{equation}\label{eq:log_I_j_I_j1}
      \log\frac{I_j(t)}{I_j(1)}
    \leq C\int_1^\infty
        (1+s)^{-2\alpha}(s+B)^{-2\nu}\,ds
    =:CD(B).  
\end{equation}
Since $\nu>0$, the integrand tends pointwise to zero as
$B\to\infty$. For $B\geq1$, it is bounded by the integrable
function $(1+s)^{-2\zeta}$. Thus $D(B)\toinfty{B}0$.

Choose $B_0\geq1$ large enough that the preceding lower bound
on $U$ holds and $CD(B_0)<\log2$, and set
$\varepsilon_0:=e^{-B_0}$.
A first time with $I_j(t)=2I_j(1)$ would contradict~\ref{eq:log_I_j_I_j1}. Hence~\eqref{eq:U_lower_bound_t_plus_B},~\eqref{eq:log_I_j_I_j1} hold for all $t\geq1$,
and
\[
    1\leq\frac{\omega_j}{I_j(1)}
    \leq e^{CD(B)}=1+\smallo{1}.
\]
Substituting $U(t)\geq c(t+B)$ into
\eqref{eq:moment_log_derivative_bound} also gives for some $C > 0$
\[
    \frac{\dot I_j(t)}{I_j(t)}
    \leq C(1+t)^{-2\zeta},
    \qquad t\geq1.
\]

Finally, $I_j$ is nondecreasing. Thus, for some $C>0$, its finite limit $\omega_j$
satisfies
\[
\begin{aligned}
    0\leq\omega_j-I_j(t)
    &=\int_t^\infty\dot I_j(s)\,ds\\
    &\leq C\omega_j
        \int_t^\infty(1+s)^{-2\zeta}\,ds\\
    &\leq CI_j(1)(1+t)^{1-2\zeta},
    \qquad t\geq1,
\end{aligned}
\]
which proves the remaining assertion.
\end{proof}

The following lemma softens the loss decay assumption to a finite-time low-loss condition.

\begin{lemma}[A finite low-loss window implies loss decay]
\label{lemma:finite-low-loss-window}
Assume~\ref{traj_ass:positive_margin_interpolation}. Define
\[
    K_A:=\frac{c_1}{\sqrt{c_2(2c_1-c_2)}},
    \qquad
    R_{\mathrm{pre}}
    :=\norm{\thetab_0}_2
      +\sqrt p\,K_A\int_0^{t_0}\eta(s)\,ds,
\]
and
\[
M_h = \sup_{i \in [m]} \sup_{\norm{\ub}_2=1}\sup_{\h \in \partial q_i(\ub)} \norm{\h}_1 < \infty,
\]
\begin{equation}\label{eq:positive_margin_global_gradient_bound}
        G_{\gamma_0}
    :=
    m\gamma_0^{\frac1L-1}M_h
    \sup_{q>0}q^{1-\frac1L}\varphi'(q)e^{-\varphi(q)}
    <\infty \ ,
\end{equation}
\[
    G_{\mathrm{pre}}
    :=\sup_{\norm{\thetab}_2\leq R_{\mathrm{pre}}}\;
      \sup_{\boldsymbol\xi\in\partial\Lcal(\thetab)}
      \norm{\boldsymbol\xi}_1 < \infty \ .
\]
Choose
$t_1>\max\{t_0,t_A\}$ sufficiently large that
\begin{equation}\label{eq:finite_window_momentum_washout}
    e^{-c_1(t_1-t_0)}(1-e^{-c_1t_0})\leq\frac12.
\end{equation}
Suppose that $t_2>t_1$ satisfies
\begin{equation}\label{eq:low_loss_window_length}
    t_2-t_1
    >
    \frac1{\mu_0}
    \log\max\left\{
      1,
      \frac{\widehat\eta(t_1)
             \left(\frac12G_{\mathrm{pre}}+G_{\gamma_0}\right)}
           {\ell(0)\sqrt{c_2(2c_1-c_2)}}
    \right\}
\end{equation}
and
\begin{equation}\label{eq:finite_low_loss_window}
    \Lcal(\thetab_t)\leq\frac12\ell(0)
    \qquad(t\in[t_1,t_2]),
\end{equation}
then
\begin{equation}\label{eq:corrected_loss_t2_threshold}
    \Lcal_A(t_2)<\ell(0),
\end{equation}
and this implies
\begin{equation}
    \Lcal(\thetab_t) \toinfty{t} 0 \ .
\end{equation}
\end{lemma}

\begin{proof}
Lemma~\ref{lemma:adam_universal_ratio} gives
$\norm{\dot\thetab_t}_2\leq\sqrt p\,K_A\eta(t)$.
Hence $\norm{\thetab_t}_2\leq R_{\mathrm{pre}}$ for $0\leq t\leq t_0$,
and local Lipschitzness of $\Lcal$ gives
$\norm{\g_t}_1\leq G_{\mathrm{pre}}<\infty$ there.
Splitting the momentum integral at $t_0$ and using
\eqref{eq:positive_margin_global_gradient_bound} and
\eqref{eq:finite_window_momentum_washout}, we obtain
\[
    \norm{\m_{t_1}}_1
    \leq e^{-c_1(t_1-t_0)}(1-e^{-c_1t_0})G_{\mathrm{pre}}
       +G_{\gamma_0}
    \leq \frac12G_{\mathrm{pre}}+G_{\gamma_0}.
\]
Lemma~\ref{lemma:adam_universal_ratio} gives
$\abs{m_j(t)}\leq K_A\sqrt{v_j(t)}$, and therefore
$H(t_1)\leq K_A\norm{\m_{t_1}}_1$.

Since $\Lcal_A\geq\Lcal$ and $\mu(t)\geq\mu_0$ on $[t_1,t_2]$,
\eqref{eq:corrected_loss_dissipation} and \eqref{eq:finite_low_loss_window}
give, for almost every $t\in[t_1,t_2]$,
\[
    \dot\Lcal_A(t)
    \leq-\mu_0\bigl(\Lcal_A(t)-\Lcal(\thetab_t)\bigr)
    \leq-\mu_0\bigl(\Lcal_A(t)-\tfrac12\ell(0)\bigr).
\]
For almost every $t\in[t_1,t_2]$,
\[
    \begin{aligned}
        \frac{d}{dt}\left[
            e^{\mu_0(t-t_1)}
            \left(\Lcal_A(t)-\frac12\ell(0)\right)
        \right]
        =
        e^{\mu_0(t-t_1)}
        \left[
            \dot\Lcal_A(t)
            +\mu_0\left(\Lcal_A(t)-\frac12\ell(0)\right)
        \right]\leq0.
    \end{aligned}
\]
By absolute continuity, integrating over $[t_1,t_2]$ gives
\[
    e^{\mu_0(t_2-t_1)}
    \left(\Lcal_A(t_2)-\frac12\ell(0)\right)
    \leq \Lcal_A(t_1)-\frac12\ell(0),
\]
and hence
\[
    \Lcal_A(t_2)
    \leq \frac12\ell(0)
       +e^{-\mu_0(t_2-t_1)}
        \left(\Lcal_A(t_1)-\frac12\ell(0)\right).
\]
By the definition of $\Lcal_A$ and the assumption
$\Lcal(\thetab_{t_1})\leq\ell(0)/2$,
\[
    \begin{aligned}
        \Lcal_A(t_1)-\frac12\ell(0)
        =
        \Lcal(\thetab_{t_1})-\frac12\ell(0)
        +\frac{\widehat\eta(t_1)}{2c_1}H(t_1)\leq \frac{\widehat\eta(t_1)}{2c_1}H(t_1).
    \end{aligned}
\]
Furthermore, the preceding momentum bound gives
\[
    H(t_1)
    \leq K_A\norm{\m_{t_1}}_1
    \leq
    \frac{c_1}{\sqrt{c_2(2c_1-c_2)}}
    \left(\frac12G_{\mathrm{pre}}+G_{\gamma_0}\right).
\]
Substituting these bounds and using~\eqref{eq:low_loss_window_length} yields
\[
    \Lcal_A(t_2)
    \leq \frac12\ell(0)
       +e^{-\mu_0(t_2-t_1)}
        \frac{\widehat\eta(t_1)
          \left(\frac12G_{\mathrm{pre}}+G_{\gamma_0}\right)}
             {2\sqrt{c_2(2c_1-c_2)}} < \ell(0).
\]
By absolute continuity and \eqref{eq:corrected_loss_dissipation},
$\Lcal_A$ is nonincreasing on $[t_2,\infty)$, with a limit
$A_\infty\geq0$. Suppose that $A_\infty>0$.
Since $\ell(q_i(t))\leq\Lcal_A(t_2)<\ell(0)$, convexity gives
\[
    q_i(t)\varphi'(q_i(t))
    \geq\varphi(q_i(t))-\varphi(0)
    \geq\log\frac{\ell(0)}{\Lcal_A(t_2)}.
\]
Set $k:=\min\{2,L\log(\ell(0)/\Lcal_A(t_2))\}>0$.
The Euler identity yields
\[
    -\innerprod{\thetab_t}{\g_t}
    =L\sum_{i=1}^m q_i(t)\varphi'(q_i(t))e^{-\varphi(q_i(t))}
    \geq k\Lcal(\thetab_t).
\]
Thus \eqref{eq:common_radial_momentum_equation} yields
\[
    \left(\frac{d}{dt}+c_1\right)
    \left(-\innerprod{\thetab_t}{\m_t}\right)
    \geq c_1k\Lcal(\thetab_t)+\widehat\eta(t)H(t)
    \geq c_1k\Lcal_A(t)
    \geq c_1kA_\infty.
\]
Consequently, multiplying by $e^{c_1(t-t_2)}$ and integrating implies
$-\innerprod{\thetab_t}{\m_t}\geq b:=kA_\infty/2>0$ eventually.

The positive-margin condition implies that there exists $c>0$ such that
\[
    \Lcal(\thetab_t)\lesssim e^{-c\norm{\thetab_t}_2^L},
    \qquad
    \norm{\thetab_t}_2^2\Lcal(\thetab_t)\lesssim1
\]
Also, \eqref{eq:positive_margin_global_gradient_bound} and the
second-moment formula, with finite $v_j(t_2)$, give
$\sqrt{v_j(t)}\lesssim1$, and hence $\norm{\m_t}_2^2\lesssim H(t)$.
Cauchy--Schwarz therefore implies, for all sufficiently large $t$,
\[
    b^2\Lcal(\thetab_t)
    \leq\norm{\thetab_t}_2^2\norm{\m_t}_2^2\Lcal(\thetab_t)
    \lesssim H(t).
\]
Hence $\Lcal(\thetab_t)\leq CH(t)$ for some $C>0$ and
$\Lcal_A(t)\leq(C+\widehat\eta(t)/(2c_1))H(t)$.
By \eqref{eq:corrected_loss_dissipation} and recalling \(\widehat\eta H/(2c_1)=\Lcal_A-\Lcal\), almost everywhere on a tail,
\[
    -\dot\Lcal_A(t)
    \geq\frac{\mu_0}{2c_1}\widehat\eta(t)H(t)
    \geq\frac{\mu_0 A_\infty}{2c_1}
          \frac{\widehat\eta(t)}{C+\widehat\eta(t)/(2c_1)}
    \gtrsim\frac{\widehat\eta(t)}{1+\widehat\eta(t)}.
\]
The right-hand side is comparable to $(1+t)^{\min\{\alpha,0\}}$,
whose integral diverges because $\alpha>-1$, contradicting
$\Lcal_A\geq0$. Thus $A_\infty=0$, and
$0<\Lcal(\thetab_t)\leq\Lcal_A(t)$ proves the remaining claim.
\end{proof}
\section{Adam - Fast Regime}

\subsection{Width-$1$ Construction}\label{section:width_1_appendix}
Recall the width-$1$ model used for the construction:
\begin{equation}\label{eq:construction_model_appendix}
    \forall \thetab = (\ub, \bfb) = (u_1,u_2,b_1,...,b_{L-1}) \in \bbR^{L+1}: \qquad f(\x; \thetab) := P(\bfb)\innerprod{\ub}{\x}, \qquad P(\bfb):= \prod_{j=1}^{L-1}b_j \ .
\end{equation}

And the following dataset, for $M > B >1, \Lambda > 0$,
\begin{equation}\label{eq:construction_dataset_appendix}
 \x_1 = \frac1\Lambda(B,1), \quad \x_2 = \frac1\Lambda(M,0), \qquad y_1=y_2 = 1   
\end{equation}

Where $\x_1$ is the unique minimum margin point for all trajectories, and $\x_2$ is the point supplying large early-time gradients.

We use the exponential loss $\Lcal(\thetab) = \sum_{i \in [2]}e^{-y_if(\x_i;\thetab)}$.

\begin{theorem}[Regularizer is constant]\label{thm:no_adam_reg}
    Let $L \geq 1, \alpha > \frac1L - \frac12$. If $R(\thetab) : \bbR^{L+1} \to \bbR$ is an implicit margin regularizer for Adam on the model~\eqref{eq:construction_model_appendix} with $\eta(t)=\eta_0(1+t)^\alpha$ and the exponential loss, then $R(\thetab)$ is constant on $\bbR^{L+1}$.
\end{theorem}

Theorem~\ref{thm:no_adam_reg} follows from proving constancy of $R$ on the following hypersurface (and its symmetries), defined for $B > 1 ,\Lambda > 0$:

\begin{equation}\label{eq:sigma_B_Lambda}
  \Sigma_{B,\Lambda} = \left \{\thetab = (u_1,u_2, b_1, ..., b_{L-1})\in \bbR^{L+1} \mid  0 < u_1 < u_2 , \forall j \in [L-1]:b_j > 0, f(\x_1; \thetab)=1 \right \}   \ .
\end{equation}

\begin{lemma}[Constancy on Hypersurfaces]\label{lemma:constancy_on_sigma_BC}
    For any $B > 1$, $\Lambda > 0$, $R$ is constant on $\Sigma_{B,\Lambda}$.
\end{lemma}

First we prove Theorem~\ref{thm:no_adam_reg} assuming Lemma~\ref{lemma:constancy_on_sigma_BC}.
\begin{proof}[Proof of Theorem~\ref{thm:no_adam_reg}]
    We prove $R$ is constant on $\Theta_+ = \left \{ 0 < u_1 < u_2, \forall j \in [L-1]:b_j > 0\right \}$. Other sign patterns and order relations between $u_1,u_2$ follow from data and initialization symmetries; see Corollary~\ref{cor:adam_symmetries_for_our_proof}. This allows showing that $R$ is constant on each such open domain; continuity of $R$ then forces constancy on the whole of $\bbR^{L+1}$.
    
    Let $\thetab^{(1)}=(\ub^{(1)}, \bfb^{(1)}),\thetab^{(2)}=(\ub^{(2)}, \bfb^{(2)}) \in \Theta_+$. For any $\Lambda > 0$ there exist unique $B^{(1)}, B^{(2)}$ with
    $$f\left(\Lambda^{-1}(B^{(1)},1); \thetab^{(1)}\right)=P(\bfb^{(1)})\innerprod{\ub^{(1)}}{\Lambda^{-1}(B^{(1)},1)} = 1 = P(\bfb^{(2)})\innerprod{\ub^{(2)}}{\Lambda^{-1}(B^{(2)},1)} = f\left(\Lambda^{-1}(B^{(2)},1); \thetab^{(2)}\right) \ .$$
    In fact,
    $$B^{(i)} = \frac{1}{u^{(i)}_1}\left(\frac{\Lambda}{P(\bfb^{(i)})} - u^{(i)}_2\right) \ .$$
    Fix large enough $\Lambda > 0$ with $B^{(1)}, B^{(2)} > 1$. By definition $\thetab^{(i)} \in \Sigma_{B^{(i)},\Lambda}$, and by Lemma~\ref{lemma:constancy_on_sigma_BC}, $R$ is constant on $\Sigma_{B^{(1)},\Lambda}$ and on $\Sigma_{B^{(2)},\Lambda}$. The point $((0,\Lambda),\bf 1)$ is a boundary point of both $\Sigma_{B^{(1)},\Lambda}, \Sigma_{B^{(2)},\Lambda}$, so by continuity of $R$,
    $$R\left(\thetab^{(1)}\right) = R\left((0,\Lambda),\bf 1\right) = R\left(\thetab^{(2)}\right)$$
\end{proof}

In turn, Lemma~\ref{lemma:constancy_on_sigma_BC} follows from the following major object of the proof. Compared to Proposition~\ref{prop:arbitrary_directional_convergence_main_text}, we strengthen the result by showing that a single dataset suffices for any fixed compact target set.

\begin{proposition}
[Arbitrary Directional Convergence]\label{prop:directional_convergence}

For any $B>1$, $\Lambda>0$, and nonempty compact
$K\subseteq\Sigma_{B,\Lambda}$, there exists $M_*=M_*(B,\Lambda,K)$
such that, for every $M\geq M_*$, every $\hat\thetab\in K$ is reached
on the fixed dataset~\eqref{eq:construction_dataset_appendix} by a suitable initialization:
\[
    \frac{\thetab_t}{\qmin^{1/L}(\thetab_t)}
    \toinfty{t}\hat\thetab.
\]
These initializations can be chosen from a compact family on which the
margin-normalized limit exists and depends continuously on initialization.
\end{proposition}

We show how Lemma~\ref{lemma:constancy_on_sigma_BC} follows from Proposition~\ref{prop:directional_convergence}.

\begin{proof}[Proof of Lemma~\ref{lemma:constancy_on_sigma_BC}]
   
    The map
\[
 (\delta,\bfb)\longmapsto
 \left(
   \frac{\Lambda\delta}{P(\bfb)(1+B\delta)},
   \frac{\Lambda}{P(\bfb)(1+B\delta)},
   \bfb
 \right),
 \qquad \delta\in(0,1),\quad\bfb\in(0,\infty)^{L-1},
\]
is a smooth parameterization of $\Sigma_{B,\Lambda}$.  Hence any two points
can be joined by a piecewise $C^1$ (and therefore absolutely continuous)
path $\psi:[0,1]\to\Sigma_{B,\Lambda}$.

Fix a point $\psi(s)=(\ub,\bfb) \neq \zero$ on such a path and choose $M$ so large
that
\[
 q_2(\psi(s))=\frac{MP(\bfb)u_1}{\Lambda}>1
 =\frac{P(\bfb)(Bu_1+u_2)}{\Lambda}=q_1(\psi(s)).
\]
Proposition~\ref{prop:directional_convergence} gives an initialization $\thetab_0$ for which 
\[
\hat \thetab = \frac{\thetab_t}{\qmin(\thetab_t)^{1/L}} \toinfty{t} \psi(s) \ ,
\]
and hence also directional convergence
$\frac{\thetab_t}{\norm{\thetab_t}_2}\to \frac{\psi(s)}{\norm{\psi(s)}_2}$.
Thus, since $R$ is an implicit margin regularizer,
there are $\kb_s\in\partial R(\psi(s))$ and
$\lambda_1,\lambda_2\geq0$ such that
\[
 \kb_s=\lambda_1\nabla q_1(\psi(s))
       +\lambda_2\nabla q_2(\psi(s)),
 \qquad
 \lambda_i(q_i(\psi(s))-1)=0.
\]
Since $q_2(\psi(s)) > q_1(\psi(s))$, it holds that $\lambda_2=0$.  At almost every
$s$ where $\dot\psi(s)$ exists, differentiation of
$q_1(\psi(s))\equiv1$ gives
\(
 \innerprod{\nabla q_1(\psi(s))}{\dot\psi(s)}=0
\).
The chain rule now
gives
\[
 (R\circ\psi)'(s)
 =\innerprod{\kb_s}{\dot\psi(s)}=0
 \quad\text{for almost every }s.
\]
The locally Lipschitz composition $R\circ\psi$ is absolutely continuous,
so it is constant.  Path connectedness proves the lemma.
\end{proof}

The remainder of the subsection is dedicated to proving Proposition~\ref{prop:directional_convergence}.

\subsubsection{Notations and Basic Properties}
We denote
$$\g_t = \nabla \Lcal(\thetab_t), \quad \mathbf p_t= -\g_t \ .$$
And we denote $g_{u_1}, g_{u_2}, g_{b_1},...,g_{b_{L-1}}$ for the scalar gradients (and similarly for $\mathbf p, \m, \vb$). We also use a vector notation $\thetab_t[i], i \in [L+1]$, where $\thetab_t[1]=u_1, \thetab_t[2] = u_2$ and $\thetab_t[i] = b_{i-2}$ for $i \in \{3,...,L+1\}$. We use the same notation for $\g_t[i], \mathbf{p}_t[i], \omega_i, I_i$ (defined below).

We denote 
$$I_{u_1}(t) = c_2\int_0^t e^{c_2s}p_{u_1}^2(s)ds, \quad I_{u_2}(t) = c_2\int_0^t e^{c_2s}p_{u_2}^2(s)ds, \quad I_{b_j} = c_2\int_0^t e^{c_2s}p_{b_j}^2(s)ds, \quad (j \in [L-1]) \ ,$$
and
$$\omega_{u_1} = \lim_{t \to \infty}I_{u_1}(t), \quad \omega_{u_2} = \lim_{t \to \infty}I_{u_2}(t), \quad \omega_{b_j} = \lim_{t \to \infty}I_{b_j}(t) \quad (j \in [L-1]) \ .$$

Define 
$$\xi_1 = \innerprod{\x_1}{\ub_t}, \quad \xi_2 = \innerprod{\x_2}{\ub_t}, \quad q_1 = P(\bfb)\xi_1 =f(\x_1; \thetab_t), \quad q_2 = P(\bfb)\xi_2 = f(\x_2; \thetab_t) \ .$$
The initialization family we consider is parameterized (in addition to $B,\Lambda,M$) by $\rho \geq 1, z \in [0, 3\log M]$, and $\pi=(\pi_1, ..., \pi_{L-1}) \in (0, \infty)^{L-1}$:

\begin{equation}\label{eq:initialization}
\forall j \in [L-1]:b_j(0) = \rho \pi_j, \qquad \ub(0) = \rho \pi_0(a, 1-Ba), \qquad P_\pi:=\prod_{j=1}^{L-1}\pi_j, \quad \pi_0 := \frac{\Lambda}{P_\pi}, \quad a:= \frac{1+\frac{z}{\rho^L}}{M} \ .    
\end{equation}

In the $L=1$ case we set as a convention $P(\bfb) \equiv 1, P_\pi=1,\pi_0 = \Lambda$.

For $M>B+1$, taking
\[
 \rho> \max \left \{1, \left(\frac{3(B+1)\log M}{M-B-1}\right)^{1/L} \right\} =: \rho_0(M,B)
\]
gives $0 < u_1(0) < u_2(0)$ for any $z, \pi$. The parameter $z$ controls the initial difference in confidence:
\begin{align*}
    q_1(0) &= P(\bfb(0))\frac1\Lambda\cdot \frac{\Lambda\rho}{P_\pi}\innerprod{(B,1)}{(a,1-Ba)} = \rho^L \ , \\
    q_2(0) &= P(\bfb(0))\frac1\Lambda\frac{\Lambda\rho}{P_\pi} \cdot Ma = \rho^L\left(1+\frac{z}{\rho^L}\right) = \rho^L + z \ .
\end{align*}
It holds with these notations that
\begin{align*}
\Lcal(\thetab_t) &= e^{-q_1} + e^{-q_2} \ ,\\
p_{u_2} &= \frac1\Lambda P(\bfb)e^{-q_1} \ ,\\
p_{u_1} &= \frac{B}{\Lambda}P(\bfb)e^{-q_1} + \frac M\Lambda P(\bfb)e^{-q_2} = Bp_{u_2} + \frac M\Lambda P(\bfb)e^{-q_2} \ , \\
\forall j \in [L-1]:p_{b_j} &= \frac{q_1}{b_j}e^{-q_1} + \frac{q_2}{b_j}e^{-q_2} \ .   
\end{align*}

Since all coordinates of $\bf p_t$ are strictly positive, all coordinates of $\thetab_t$ are strictly increasing, hence so are $\xi_1,\xi_2, q_1, q_2$. It is also simple to show that in fact $p_i$ are nonincreasing, since (recalling $\frac {d}{dt} \xi_1 > 0, \frac{d}{dt}b_j > 0$ and $q_1 \geq 1$)
\begin{align}\label{eq:d_dt_log_gradient_components}
    -\frac{d}{dt}\log\left(P(\bfb)e^{-q_1}\right) &= \dot q_1- \frac{d}{dt}\log P(\bfb) = \dot q_1 - \frac{d}{dt}\left(\log q_1 - \log \xi_1\right) \geq \dot q_1 - \frac{\dot q_1}{q_1} \geq 0, \\
    -\frac{d}{dt}\log \frac{q_1}{b_j}e^{-q_1} &= - \frac{d}{dt}\log q_1e^{-q_1} + \frac{d}{dt} \log b_j \geq \dot q_1 -\frac{\dot q_1}{q_1} \geq 0 \ .
\end{align}

An identical calculation holds for terms with $q_2$.

\begin{lemma}[Ordering of margins]\label{lemma:ordering_of_margins}
    For every $B > 1$ there exists some $M_0(B)$ so that for every $M \geq M_0(B)$,
    $$\forall t \geq0:\quad\xi_2(t)  \geq \xi_1(t), \quad q_2(t) \geq q_1 (t)\geq 1 \ ,$$
    And the differences $\xi_2 - \xi_1, q_2 - q_1$ are strictly increasing. It therefore holds that $\qmin(t)=q_1(t)$. Also, 
    $$\forall t \geq 0: q_2e^{-q_2} \leq q_1e^{-q_1} \ .$$
\end{lemma}
\begin{proof}
    Denote $\xi_\Delta(t) = \xi_2(t) - \xi_1(t)$. At initialization,
    $$\xi_\Delta(0) = \frac{\rho}{P_\pi}Ma - \frac{\rho}{P_\pi} = \frac{\rho}{P_\pi}\frac{z}{\rho^L} \geq 0 \ ,$$
    and since $\dot u_1(0) = \dot u_2(0) = \eta_0$,
    
    $$\dot \xi_\Delta(0) = \frac M\Lambda \dot u_1(0) - \frac 1\Lambda\left(B \dot u_1(0) + \dot u_2(0)\right) = \eta_0\frac{M-B-1}{\Lambda} > 0  \ .$$

    So there exists an initial interval on which $\xi_\Delta(t) \geq 0$. We show that on any such interval, we have $\dot \xi_\Delta > 0$. Indeed, we have $\xi_2 \geq \xi_1$ and $q_2 \geq q_1$, so
    $$Bp_{u_2} \leq p_{u_1} \leq Bp_{u_2} + \frac M\Lambda P(\bfb) e^{-q_1} = (B+M)p_{u_2}$$
    Therefore, 
    \[
      B(-m_{u_2})\leq -m_{u_1}
      \leq(B+M)(-m_{u_2}),
      \qquad v_{u_1}\leq(B+M)^2v_{u_2}.
    \]
    And since the bias correction terms cancel,
    $$\frac{\dot u_1}{\dot u_2} \geq \frac{B}{B+M}$$
    So
    $$\dot \xi_\Delta = \frac M\Lambda \dot u_1 - \frac 1\Lambda\left(B \dot u_1 + \dot u_2\right) = \frac 1\Lambda\left((M-B)\dot u_1 - \dot u_2\right) \geq \frac 1\Lambda \dot u_2\left(\frac{B(M-B)}{B+M}-1\right) > 0$$
    for large enough $M$ when $B > 1$. Therefore, a first-crossing argument proves $\xi_\Delta \geq 0$ everywhere (otherwise there would be a first crossing point $t_0$ where $\xi_\Delta(t_0) < 0$, but we have proved that on $[0,t_0]$ we have $\xi_\Delta \geq 0, \dot \xi_\Delta > 0$, a contradiction). Hence also $\dot \xi_\Delta > 0$ everywhere. Because
    $q_2-q_1=P(\bfb)\xi_\Delta$, with $P>0$ increasing, this difference is strictly
    increasing as well. Now, since $q_1, q_2$ are increasing, $q_2 \geq q_1$ and $q_1(0) = \rho^L \geq 1$, and since $x \to xe^{-x}$ is strictly decreasing on $[1, \infty)$ we get the final claim.
\end{proof}

\begin{lemma}[Coordinate Comparability]\label{lemma:coordinate_comparability} For any compact  $\Pi \subseteq (0,\infty)^{L-1}$, there exist constants $0 < k_j \leq K_j  < \infty,\  j \in [L+1]$, so that for any $\rho\geq \rho_0 \geq 1, z \in [0, 3 \log M]$ and $\pi \in \Pi$, it holds that
\[
    k_j \norm{\thetab_t}_2\leq \theta_j(t) \leq K_j\norm{\thetab_t}_2, \qquad k_j \norm{\mathbf p_t}_2\leq p_j(t) \leq K_j\norm{\mathbf p_t}_2, \qquad (t \geq 0).
\]
As a result, such constants exist for $-\m_t, \vb_t$.
\end{lemma}
\begin{proof}
    First note that $u_1 \asymp u_2$ since $Bp_{u_2} \leq p_{u_1} \leq (B+M)p_{u_2}$ implies $\frac{B}{B+M} \leq \frac{\dot u_1}{\dot u_2} \leq \frac{B+M}{B}$, and $\frac{1}{M-B}\leq \frac{u_1(0)}{u_2(0)} \leq 1$. We claim that this fact and Lemma~\ref{lemma:ordering_of_margins} imply that $\theta_jp_j \asymp q_1e^{-q_1}$ for all $j$, and in particular that there exist $k_p, K_p > 0$ with 
    \begin{equation}\label{eq:p_theta_ratios_bound}
     k_p\frac{\theta_j}{\theta_i}\leq\frac{p_i}{p_j} \leq K_p\frac{\theta_j}{\theta_i} \ , \qquad t \geq 0, \quad i,j \in [L+1]\ .   
    \end{equation}
    Indeed,
    \[
    u_1p_{u_1} = \frac{Bu_1}{Bu_1+u_2}q_1e^{-q_1}+q_2e^{-q_2}, \quad u_2p_{u_2} = \frac{u_2}{Bu_1+u_2}q_1e^{-q_1}, \quad b_jp_{b_j} = q_1e^{-q_1} + q_2e^{-q_2} \ ,
    \]
    and recall that $q_2 e^{-q_2} \leq q_1e^{-q_1}$. Note the constants $k_p,K_p$ in~\eqref{eq:p_theta_ratios_bound} do not depend on $\rho,z,\pi$, so proving comparability of parameters suffices for comparability of gradients.

    All constants below may depend on $M,\Pi$ and the structural
parameters, but not on $\rho,z,\pi$.
Set $\delta=(c_1-c_2/2)/2>0$, and choose $T>0$ such that
$(\log\hat\eta)'\leq\delta$ for $t\geq T$.
The initialization and the universal velocity bound imply that there exist $c_T,C_T>0$ such that
\[
    c_T\rho\leq\theta_i(t)\leq C_T\rho,
    \qquad 0\leq t\leq T,
    \qquad
    \dot\theta_i(T)\leq C_T.
\]
For $t\geq T$, write
\[
    \theta_i^{\mathrm{RMS}}(t)
    :=\theta_i(T)
      +\int_T^t\hat\eta(s)\frac{p_i(s)}{\sqrt{v_i(s)}}\,ds.
\]
Lemma~\ref{lemma:time_derivative_uncorrected_adam_ratio}
and the choice of $T$ give
\[
    \ddot\theta_i+\delta\dot\theta_i
    \leq c_1\dot\theta_i^{\mathrm{RMS}}.
\]
Also, monotonicity of $p_i$ gives
$-m_i(t)\geq(1-e^{-c_1t})p_i(t)$.
Integrating these two inequalities yields
\[
\begin{aligned}
    \theta_i(t)
    &\leq\theta_i(T)+\frac{\dot\theta_i(T)}{\delta}
       +\frac{c_1}{\delta}
        \bigl(\theta_i^{\mathrm{RMS}}(t)
              -\theta_i^{\mathrm{RMS}}(T)\bigr),\\
    \theta_i(t)
    &\geq\theta_i(T)
       +(1-e^{-c_1T})
        \bigl(\theta_i^{\mathrm{RMS}}(t)
              -\theta_i^{\mathrm{RMS}}(T)\bigr).
\end{aligned}
\]
There exists $c>0$ such that $\theta_i(T)\geq c\rho\geq c$ for every $i$, and hence there exist uniform constants $c_*,C_*>0$ such that
\[
    c_*\theta_i^{\mathrm{RMS}}(t)
    \leq\theta_i(t)
    \leq C_*\theta_i^{\mathrm{RMS}}(t),
    \qquad t\geq T.
\]

Define for $i,j$,
\[
    \mathcal K_{ij}(t)
    :=\max\left\{
        K_0,\,
        \max_{T\leq s\leq t}
        \frac{\theta_i^{\mathrm{RMS}}(s)}
             {\theta_j^{\mathrm{RMS}}(s)}
    \right\}, \qquad K_0 := \max_{i,j}\sup_{0 \leq s \leq T}\frac{\theta_i(s)}{\theta_j(s)}.
\]
For every $0\leq s\leq t$, the preceding bounds imply
$\theta_i(s)/\theta_j(s)\leq C\mathcal K_{ij}(t)$ for some $C > 0$.
Thus, for some $C>0$,~\eqref{eq:p_theta_ratios_bound} and the second-moment
formula give
\[
    \sqrt{\frac{v_j(s)}{v_i(s)}}
    \leq C\mathcal K_{ij}(t),
    \qquad T\leq s\leq t.
\]
Using~\eqref{eq:p_theta_ratios_bound} once more, there exists $C_1>0$ such that
\[
    \frac{
        \theta_i^{\mathrm{RMS}}(s)
        \dot\theta_i^{\mathrm{RMS}}(s)
    }{
        \theta_j^{\mathrm{RMS}}(s)
        \dot\theta_j^{\mathrm{RMS}}(s)
    }
    =
    \frac{\theta_i^{\mathrm{RMS}}(s)}
         {\theta_j^{\mathrm{RMS}}(s)}
    \frac{p_i(s)}{p_j(s)}
    \sqrt{\frac{v_j(s)}{v_i(s)}}
    \leq C_1\mathcal K_{ij}(t).
\]
So,
\[
\frac{d}{ds}\bigl(\theta_i^{\mathrm{RMS}}(s)\bigr)^2
\leq C_1\mathcal K_{ij}(t)
\frac{d}{ds}\bigl(\theta_j^{\mathrm{RMS}}(s)\bigr)^2.
\]
Fix $t > T$. If $\mathcal K_{ij}(t)>K_0$, choose $s_*\in[T,t]$ attaining
its maximum. Integrating on $[T,s_*]$ gives
\[
\begin{aligned}
\bigl(\theta_i^{\mathrm{RMS}}(s_*)\bigr)^2
-\bigl(\theta_i^{\mathrm{RMS}}(T)\bigr)^2
\leq C_1\mathcal K_{ij}(t)
\left[
\bigl(\theta_j^{\mathrm{RMS}}(s_*)\bigr)^2
-\bigl(\theta_j^{\mathrm{RMS}}(T)\bigr)^2
\right].
\end{aligned}
\]
And the choice of \(s_*\) gives
\[
\begin{aligned}
\theta_i^{\mathrm{RMS}}(s_*)
&=\mathcal K_{ij}(t)\theta_j^{\mathrm{RMS}}(s_*), \\
    \mathcal K_{ij}(t)^2
    \bigl(\theta_j^{\mathrm{RMS}}(s_*)\bigr)^2
    &\leq
    \bigl(\theta_i^{\mathrm{RMS}}(T)\bigr)^2+C_1\mathcal K_{ij}(t)
    \left(
        \bigl(\theta_j^{\mathrm{RMS}}(s_*)\bigr)^2
        -\bigl(\theta_j^{\mathrm{RMS}}(T)\bigr)^2
    \right).
\end{aligned}
\]
Since
$\theta_i^{\mathrm{RMS}}(T)
 \leq K_0\theta_j^{\mathrm{RMS}}(T)$
and $\theta_j^{\mathrm{RMS}}$ is increasing,
\[
    \mathcal K_{ij}(t)^2
    \leq K_0^2+C_1\mathcal K_{ij}(t),
\]
and hence $\mathcal K_{ij}(t)\leq K_0+C_1$.
This bound also holds when $\mathcal K_{ij}(t)=K_0$.

Applying this to every ordered pair proves uniform
comparability of the RMSprop coordinates, and hence of the
parameters.
Equation~\eqref{eq:p_theta_ratios_bound} then proves
comparability of the gradients. Positivity of the EMA kernels
gives the corresponding comparisons for $-\m_t$ and $\vb_t$.

\end{proof}

\begin{lemma}[Loss Decay and Positive Margin]\label{lemma:construction_loss_decay_positive_margin} It holds that $\Lcal(\thetab_t) \toinfty{t} 0$ and there exist $\gamma_0, t_0 > 0$ with $\forall t \geq t_0: \gamma(\thetab_t) \geq \gamma_0 > 0$.
\end{lemma}
\begin{proof}
    We begin with loss decay. Since all parameters are positive and increasing, so are $P(\bfb), q_1$. If towards a contradiction $q_1 \toinfty{t} Z < \infty$, this implies a positive lower bound $p_{u_2}(t) \geq p_* > 0$, and since additionally $p_{u_2}$ is nonincreasing,
    \[
    -\hat m_{u_2}(t) \geq p_* \ , \qquad \sqrt{\hat v_{u_2}(t)} \leq p_{u_2}(0), \qquad \dot u_2(t) \geq \frac{p_*}{p_{u_2}(0)}\eta(t) \ . 
    \]
    Since $\eta(t) = \eta_0(1+t)^{\alpha}, \alpha > -1$, this implies $u_2(t) \toinfty{t} \infty$, but this in turn implies $q_1(t) \toinfty{t} \infty$, a contradiction. Thus $\Lcal(\thetab_t) \leq 2e^{-q_1(t)} \toinfty{t} 0$.

    The margin lower bound follows from parameter coordinate comparability: $\qmin(\thetab_t) = q_1(t) = P(\bfb)\frac{Bu_1+u_2}{\Lambda} \gtrsim \norm{\thetab_t}_2^L$ since each parameter is comparable to the norm.       
\end{proof}

\begin{corollary}[Second Moment Freezing and Asymptotics]
\label{corr:construction_freezing_and_asymp}
Suppose $\alpha>\frac1L-\frac12$.
For all $j\in[L+1]$,
\begin{equation}\label{eq:frozen_weights_positive_finite}
    0<\omega_j<\infty \ ,
\end{equation}
\begin{equation}\label{eq:parameter_velocity_bound}
    \dot\theta_j(t)=\bigO{t^{1/L-1}} \ .
\end{equation}
Moreover,
\begin{equation}\label{eq:construction_margin_asymptotics}
    q_1(t)=\frac{c_2}{2}t+\smallo{t},
    \qquad
    q_2(t)-q_1(t)\longrightarrow\infty \ .
\end{equation}
\end{corollary}

\begin{proof}
Lemma~\ref{lemma:construction_loss_decay_positive_margin}
provides the conditions of
Theorem~\ref{thm:loss_decay_rates_appendix}, which gives
\[
    U(t)=\frac{c_2}{2}t
         +\left(\alpha+2-\frac2L\right)\log t+\bigO{1}.
\]
Since $q_1-\log2\leq U\leq q_1$, this proves the asserted
leading rate of $q_1$.

Lemma~\ref{lemma:freezing-tail-scales} and coordinate
comparability give
\[
    \theta_j(t)\asymp t^{1/L},
    \qquad
    p_j(t)\leq\norm{\g_t}_2
    \lesssim U(t)^{1-1/L}e^{-U(t)}
    =\bigO{e^{-c_2t/2}t^{-\zeta}},
\]
where $\zeta=\alpha+1-1/L>1/2$.
Consequently,
\[
    \omega_j=c_2\int_0^\infty e^{c_2t}p_j(t)^2\,dt<\infty,
\]
and positivity follows from $p_j>0$. Since $c_1>c_2/2$, the first-moment formula gives
\[
    |\hat m_j(t)|=\bigO{e^{-c_2t/2}t^{-\zeta}}.
\]
Also, $I_j(t)\to\omega_j>0$ implies
$\sqrt{\hat v_j(t)}\asymp e^{-c_2t/2}$. Thus
\[
    \dot\theta_j(t)
    =\eta(t)\frac{|\hat m_j(t)|}{\sqrt{\hat v_j(t)}}
    =\bigO{t^{\alpha-\zeta}}
    =\bigO{t^{1/L-1}}.
\]

Finally, Lemma~\ref{lemma:ordering_of_margins} gives
$\dot\xi_\Delta\geq k\dot u_2$ for some $k>0$.
Since $u_2(t)\asymp t^{1/L}\to\infty$, it follows that
$\xi_\Delta(t)\to\infty$. Therefore,
\[
    q_2(t)-q_1(t)
    =P(\bfb(t))\xi_\Delta(t)
    \geq P(\bfb(0))\xi_\Delta(t)
    \longrightarrow\infty.
\]
\end{proof}

\begin{lemma}[Bounds on the Second-Moment Weight Ratios]
\label{lemma:second_moment_weights_bounds}
For every compact set
$\Pi\subseteq(0,\infty)^{L-1}$, there exists
$k_{\Pi}>0$, independent of $M$, such that for every
$M\geq M_0(B)$ there exists $\rho_1=\rho_1(M,\Pi)$
with the following property. For every $\rho\geq\rho_1$,
$z\in[0,3\log M]$, and $\pi\in\Pi$,
\begin{equation}\label{eq:omega_layer_ratio_bounds}
    \frac18
    \leq
    \frac{\pi_0^2\omega_{u_2}}
         {\pi_j^2\omega_{b_j}}
    \leq8,
    \qquad j\in[L-1],
\end{equation}
and
\begin{equation}\label{eq:omega_input_ratio_bounds}
    B^2
    \leq\frac{\omega_{u_1}}{\omega_{u_2}}
    \leq(B+Me^{-z})^2.
\end{equation}
Moreover, when $z=0$,
\begin{equation}\label{eq:omega_input_ratio_lower_z_zero}
    \frac{\omega_{u_1}}{\omega_{u_2}}
    \geq k_{\Pi}M.
\end{equation}
For $L=1$, the bounds in~\eqref{eq:omega_layer_ratio_bounds}
are void, and the dependence on $\Pi$ is omitted.
\end{lemma}

\begin{proof}
Recall that
\[
    I_j(t)=c_2\int_0^t e^{c_2s}p_j(s)^2\,ds,
    \qquad
    \omega_j=I_j(\infty).
\]
By Corollary~\ref{corr:construction_freezing_and_asymp},
all these weights are positive and finite.
All limits as $\rho\to\infty$ below are uniform for
$z\in[0,3\log M]$ and $\pi\in\Pi$, with
$M,\Pi$ fixed. Constants indexed only by $\Pi$
are independent of $M$.

First, Lemma~\ref{lemma:ordering_of_margins} gives
$q_2(t)-q_1(t)\geq z$, and hence
\[
    B
    \leq\frac{p_{u_1}(t)}{p_{u_2}(t)}
    =B+Me^{-(q_2(t)-q_1(t))}
    \leq B+Me^{-z}.
\]
Squaring and integrating proves
\eqref{eq:omega_input_ratio_bounds}.

We next compare the hidden-layer weights with $\omega_{u_2}$.
The universal Adam-ratio bound (Lemma~\ref{lemma:adam_universal_ratio}) implies that there exists $K>0$ such that
\[
    0\leq\theta_j(t)-\theta_j(0)\leq Kt,
\]
where $K$ is independent of the initialization.
Since $Bu_1(0)+u_2(0)=\rho\pi_0$ and
$b_j(0)=\rho\pi_j$, it follows for $0\leq t\leq1$ that \[
\begin{aligned}
    0
    &\leq\frac{Bu_1(t)+u_2(t)}{\rho\pi_0}-1
    \leq\frac{(B+1)K}{\rho\pi_0},\\
    0
    &\leq\frac{b_j(t)}{\rho\pi_j}-1
    \leq\frac{K}{\rho\pi_j},
    \qquad j\in[L-1].
\end{aligned}
\]
Compactness of $\Pi$ bounds
$\pi_0,\pi_1,\ldots,\pi_{L-1}$ above and away from zero,
so these relative errors are $\bigO{\rho^{-1}}$, uniformly
on $[0,1]$, with constants independent of $M$.
Taking the product of the normalized $b_j$ coordinates, and
using $\prod_{j=1}^{L-1}\pi_j=\Lambda/\pi_0$, gives the first
estimate below. Taking the quotient of the normalized
$Bu_1+u_2$ and $b_j$ gives the second:
\begin{equation}\label{eq:weight_fixed_time_prefactors}
\begin{aligned}
    \frac{P(\bfb(t))}{\Lambda}
    &=\pi_0^{-1}\rho^{L-1}
      \left(1+\bigO{\rho^{-1}}\right),\\
    \frac{Bu_1(t)+u_2(t)}{b_j(t)}
    &=\frac{\pi_0}{\pi_j}
      \left(1+\bigO{\rho^{-1}}\right),
      \qquad j\in[L-1],
\end{aligned}
\end{equation}
uniformly on $[0,1]$. The constants here may depend on
$\Pi$, but not on $M$.

For $L>1$, the exact gradient formulas give
\[
    \frac{p_{b_j}}{p_{u_2}}
    =
    \frac{Bu_1+u_2}{b_j}
    \left(
      1+\frac{q_2}{q_1}e^{-(q_2-q_1)}
    \right).
\]
The parenthesis lies in $[1,2]$ by
Lemma~\ref{lemma:ordering_of_margins}. Thus the second line of
\eqref{eq:weight_fixed_time_prefactors}, followed by
squaring and integration, gives
\begin{equation}\label{eq:weight_fixed_time_ratios}
    \frac14-\smallo{1}
    \leq
    \frac{\pi_0^2I_{u_2}(1)}
         {\pi_j^2I_{b_j}(1)}
    \leq1+\smallo{1}.
\end{equation}
It remains to control the contribution after time $1$. In the following tail estimates, constants may depend on
$M,\Pi$, but not on $\rho,z,\pi$.

Lemma~\ref{lemma:coordinate_comparability} shows that parameter,
gradient and moment coordinates are comparable with constants
independent of $\rho,z,\pi$.
Consequently, homogeneity and Euler's identity give
\begin{equation}\label{eq:weight_uniform_gradient_scale}
    q_1\asymp\norm{\thetab_t}_2^L,
    \qquad
    \norm{\mathbf p_t}_2
    \asymp\norm{\thetab_t}_2^{L-1}\Lcal
    \asymp U^\nu e^{-U}.
\end{equation}
We verify the conditions of
Lemma~\ref{lemma:second_moment_early_dominance}.
Increasing the threshold for $\rho$, we may assume throughout
that $U(t)\geq\rho^L-\log2\geq\max\{1,4\nu\}$.
All gradients have fixed signs and are monotone decreasing
in absolute value.
Lemma~\ref{lemma:coordinate_comparability} and
\eqref{eq:weight_uniform_gradient_scale} imply that there exist $k,K>0$ such that
\[
    kU(t)^\nu e^{-U(t)}
    \leq p_j(t)
    \leq KU(t)^\nu e^{-U(t)},
    \qquad t\geq1/2,\quad j\in[L+1],
\]
with $k,K>0$ independent of $\rho,z,\pi$.
Finally, increasing $K$ if necessary, the explicit initial gradients and their monotonicity
give, for every coordinate $j$,
\begin{equation}\label{eq:initial_accumulated_moment_bound}
    I_j(1)
    \leq(e^{c_2}-1)p_j(0)^2
    \leq K\rho^{2(L-1)}e^{-2\rho^L}
    \longrightarrow0.
\end{equation}
For $L>1$,
Lemma~\ref{lemma:second_moment_early_dominance} yields, uniformly in
$z\in[0,3\log M]$ and $\pi\in\Pi$ with $M$ fixed,
\begin{equation}\label{eq:weight_total_tail_small}
    \omega_j-I_j(1)
    =\smallo{I_j(1)}
    \qquad(\rho\to\infty).
\end{equation}
Therefore,
\begin{equation}\label{eq:weight_fixed_time_dominance}
    \omega_j=I_j(1)(1+\smallo{1}).
\end{equation}
Combining this with~\eqref{eq:weight_fixed_time_ratios}
proves~\eqref{eq:omega_layer_ratio_bounds}, after increasing
$\rho_1(M,\Pi)$. In particular,
\[
    \omega_{u_2}\leq2I_{u_2}(1)
\]
for all sufficiently large $\rho$.

For $L=1$, the gradient-scale constants for the $u_2$ coordinate
can be chosen independently of $M$. Indeed,
$p_{u_2}=\Lambda^{-1}e^{-q_1}$ and $q_2\geq q_1$, so
\[
    \frac{1}{2\Lambda}\Lcal(\thetab_t)
    \leq p_{u_2}(t)
    \leq\frac{1}{\Lambda}\Lcal(\thetab_t).
\]
Lemma~\ref{lemma:second_moment_early_dominance}, applied to
the $u_2$ coordinate, therefore implies that there exist $K,K_*>0$ such that
\[
    \frac{\dot I_{u_2}(t)}{I_{u_2}(t)}
    \leq K\eta(t)^{-2},
    \qquad t\geq1,
    \qquad
    \omega_{u_2}\leq K_*I_{u_2}(1),
\]
with $K,K_*$ independent of $M,\rho,z$.
Together with the preceding argument for $L>1$, we have proved,
in every depth, with $K_*>0$ enlarged if necessary,
\begin{equation}\label{eq:omega_u2_initial_moment_bound}
    \omega_{u_2}\leq K_*I_{u_2}(1),
\end{equation}
where $K_*$ is independent of $\Pi,M,\rho,z$ ($\Pi$ is void for $L=1$).

It remains to prove the enhancement when $z=0$. Set
\[
    \tau:=\rho^{1-L},
    \qquad
    p_0:=p_{u_2}(0)
         =\pi_0^{-1}\rho^{L-1}e^{-\rho^L}.
\]
We first bound $I_{u_2}(1)$ from above.
Since $p_{u_2}$ is nonincreasing, its bias-corrected moments satisfy
\(-\hat m_{u_2}(t)\geq p_{u_2}(t), \sqrt{\hat v_{u_2}(t)}\leq p_0,\), and bounding $\eta$ from below on $[0,1]$ implies that there exists $c>0$ such that
\[
    \dot u_2(t)\geq c\frac{p_{u_2}(t)}{p_0} =
c\frac{P(\bfb(t))}{P(\bfb(0))}
e^{-(q_1(t)-\rho^L)}
\ge ce^{-(q_1(t)-\rho^L)} .
\]
Since $P$ is nondecreasing and
$P(\bfb(0))/\Lambda=\pi_0^{-1}\tau^{-1}$, there exists $c_{\Pi}>0$ such that
\[
\begin{aligned}
    \dot q_1(t)
    &\geq\frac{P(\bfb(t))}{\Lambda}\dot u_2(t)\geq c_{\Pi}\tau^{-1}
       e^{-(q_1(t)-\rho^L)}.
\end{aligned}
\]
On the other hand, \eqref{eq:weight_fixed_time_prefactors} implies that there exists $C_{\Pi}>0$ such that
\[
    p_{u_2}(t)
    =p_0\frac{P(\bfb(t))}{P(\bfb(0))}
       e^{-(q_1(t)-\rho^L)}
    \leq C_{\Pi}p_0e^{-(q_1(t)-\rho^L)}.
\]
Consequently, for some $C_{\Pi}>0$,
\[
\begin{aligned}
    I_{u_2}(1)
    &\leq C_{\Pi}p_0^2
       \int_0^1 e^{-2(q_1(t)-\rho^L)}\,dt \leq C_{\Pi}\tau p_0^2
       \int_0^1\dot q_1(t)e^{-(q_1(t)-\rho^L)}\,dt \leq C_{\Pi}\tau p_0^2.
\end{aligned}
\]
By~\eqref{eq:omega_u2_initial_moment_bound}, for some $C_{\Pi}>0$, this implies
\begin{equation}\label{eq:omega_u2_upper_bound}
    \omega_{u_2}\leq C_{\Pi}\tau p_0^2.
\end{equation}

Now assume $z=0$, so that $q_2(0)=\rho^L$.
Since $q_2\leq(M/B)q_1$, for some $C_{\Pi}>0$, the fixed-time parameter estimates give
\[
    \frac{P(\bfb(t))}{\Lambda}\leq C_{\Pi}\tau^{-1},
    \qquad
    \frac{q_2(t)}{b_j(t)}
    \leq\frac MB\frac{q_1(t)}{b_j(t)}
    \leq C_{\Pi}M\tau^{-1}.
\]
For some $C_{\Pi}>0$, the universal velocity bound therefore yields
\[
    \dot q_2(t)
    =
    \frac{M}{\Lambda}P(\bfb(t))\dot u_1(t)
    +\sum_{j=1}^{L-1}\frac{q_2(t)}{b_j(t)}\dot b_j(t)
    \leq C_{\Pi}M\tau^{-1},
    \qquad 0\leq t\leq1.
\]
There exists $b_{\Pi}\in(0,1]$ sufficiently small such that
\[
    q_2(t)\leq\rho^L+1,
    \qquad 0\leq t\leq b_{\Pi}\tau/M.
\]
Since $P$ is nondecreasing, on this interval
\[
    p_{u_1}(t)
    \geq\frac{M}{\Lambda}P(\bfb(t))e^{-q_2(t)}
    \geq e^{-1}Mp_0.
\]
Hence
\[
    \omega_{u_1}
    \geq c_2\int_0^{b_{\Pi}\tau/M}p_{u_1}(t)^2\,dt
    \geq c_2b_{\Pi}e^{-2}M\tau p_0^2.
\]
Dividing by~\eqref{eq:omega_u2_upper_bound}
proves~\eqref{eq:omega_input_ratio_lower_z_zero}.
All constants are independent of $M$.
For $L=1$, the same argument applies with $\tau=1$ and with
the empty products and sums interpreted as $1$ and $0$,
respectively.
\end{proof}

\begin{lemma}[Continuity of Second Moment Weights]
\label{lemma:continuity_weights}
    Fix $M\geq M_0(B)$ and a compact
    $\Pi\subseteq(0,\infty)^{L-1}$. If
    $\rho\geq\rho_1(M,\Pi)$, then
    $\omega_{u_1},\omega_{u_2}$ and every $\omega_{b_j}$ are continuous
    in $(z,\pi)\in[0,3\log M]\times\Pi$.
\end{lemma}

\begin{proof}
Write $\sigma=(z,\pi)\in[0,3\log M]\times\Pi$.
We first show continuity of the finite-time weights, and then
show that they converge uniformly to $\omega_i$.

Fix $T<\infty$. The initialization depends continuously on $\sigma$,
and the universal Adam-ratio bound gives
\[
    0\leq\dot\theta_i(t;\sigma)\leq K_A\eta(t).
\]
Thus all trajectories on $[0,T]$ lie in one compact rectangle
$\mathcal R_T\subseteq(0,\infty)^{L+1}$.
On this rectangle the gradient coordinates $p_i$ are Lipschitz,
and there exist $a_T,b_T>0$ such that
\[
    0<a_T\leq p_i\leq b_T<\infty.
\]

The normalized EMA is a contraction in the supremum norm.
Moreover, if $p,\tilde p$ take values in $[a_T,b_T]$, then their
second-moment EMAs are at least $a_T^2$. Consequently, denoting $m = \emahat{p}{c_1}, v=\emahat{p^2}{c_2}$ and $\tilde m = \emahat{\tilde p}{c_1},\tilde v= \emahat{\tilde p^2}{c_2}$,
\begin{equation}\label{eq:adam_history_lipschitz}
\begin{aligned}
    &\left|
      \mathsf A_{c_1,c_2}[p](t)
      -\mathsf A_{c_1,c_2}[\tilde p](t)
    \right|\\
    & \leq \frac{\abs{m - \tilde m}}{\sqrt{v}} + \abs{\tilde m}\abs{\frac{1}{\sqrt{v}}-\frac{1}{\sqrt{\tilde v}}}\\
    &\qquad\leq
      \frac1{a_T}\|p-\tilde p\|_{\infty,[0,t]}
      +\frac{b_T}{2a_T^3}
       \|p^2-\tilde p^2\|_{\infty,[0,t]}\\
    &\qquad\leq
      \left(\frac1{a_T}+\frac{b_T^2}{a_T^3}\right)
      \|p-\tilde p\|_{\infty,[0,t]}.
\end{aligned}
\end{equation}
At $t=0$, both Adam ratios are defined to be $1$, so the same
estimate holds there.

Using the integral representation
\begin{equation}\label{eq:adam_integral_equation}
    \theta_i(t;\sigma)
    =\theta_i(0;\sigma)
     +\int_0^t\eta(s)\,
      \mathsf A_{c_1,c_2}
      [p_i(\cdot; \sigma)](s)\,ds,
\end{equation}
set
\[
    \varepsilon(t):=\sup_{0\leq s\leq t}
    \norm{\thetab(s;\sigma)-\thetab(s;\tilde\sigma)}_\infty.
\]
Subtracting the two integral equations and using
\eqref{eq:adam_history_lipschitz} and Lipschitzness of the $p_i$
shows that there exists $K_T>0$ such that
\[
    \varepsilon(t)\leq \varepsilon(0)+K_T\int_0^t\eta(s)\varepsilon(s)\,ds.
\]
Gronwall's inequality therefore yields
\[
    \varepsilon(T)\leq \varepsilon(0)
    \exp\left(K_T\int_0^T\eta(s)\,ds\right) \overset{\sigma \to \tilde \sigma}\longrightarrow 0.
\]
Thus the trajectories,
and hence the gradients, depend continuously on $\sigma$ in the
supremum norm on $[0,T]$. In particular, every map
\[
    \sigma\longmapsto
    I_i(T;\sigma)
    =c_2\int_0^T e^{c_2t}p_i(t;\sigma)^2\,dt
\]
is continuous.

For the tail, Lemma~\ref{lemma:second_moment_early_dominance}, whose
hypotheses were verified in the preceding proof, implies that there exists $K>0$ such that, uniformly
in $\sigma$, for every coordinate $i$,
\begin{equation}\label{eq:uniform_moment_log_derivative_bound}
    \frac{\dot I_i(t;\sigma)}{I_i(t;\sigma)}
    \leq K(1+t)^{-2\zeta},
    \qquad t\geq1.
\end{equation}

Finite-time continuity and compactness give
$\sup_{\sigma,i}I_i(1;\sigma)<\infty$.
The same lemma gives, for every coordinate $i$ and $T\geq1$,
\[
\begin{aligned}
    0\leq\omega_i(\sigma)-I_i(T;\sigma)
    &\leq KI_i(1;\sigma)(1+T)^{1-2\zeta}\\
    &\leq K\sup_{\sigma,i}I_i(1;\sigma)(1+T)^{1-2\zeta}.
\end{aligned}
\]
The right-hand side tends to zero independently of $\sigma$.
Therefore $\omega_i$ is a uniform limit of the continuous maps
$I_i(T;\cdot)$, and is continuous.
\end{proof}

\begin{lemma}[Frozen weights determine the limiting direction]\label{lemma:frozen_weights_determine_direction}
    Let $\delta_0 = B\sqrt{\frac{\omega_{u_2}}{\omega_{u_1}}}$ and let $M \geq M_0(B)$. Then
    $$\frac{\thetab_t}{u_2(t)} \toinfty{t} (\delta_0, 1, \delta_1,..., \delta_{L-1}), \quad \delta_j = \sqrt{1+B\delta_0}\left(\frac{\omega_{u_2}}{\omega_{b_j}}\right)^{1/4} \qquad (j \geq 1) \ .$$
    Therefore the margin-normalized limit is
    $$\frac{\thetab_t}{\qmin^{1/L}(t)} = \frac{\thetab_t}{q_1^{1/L}(t)} \toinfty{t}  \left(\prod_{j=1}^{L-1}\delta_j\frac{B\delta_0 +1}{\Lambda}\right)^{-1/L}\left(\delta_0,1,\delta_1,...,\delta_{L-1}\right) \ .$$
\end{lemma}
\begin{proof}
By Corollary~\ref{corr:construction_freezing_and_asymp},
$$\frac{p_{u_1}}{p_{u_2}} = B + Me^{-(q_2-q_1)} \toinfty{t}B.$$
Moreover, $q_1=c_2t/2+\smallo{t}$, so
Lemma~\ref{lemma:coordinate_comparability} and
$q_1=P(\bfb)(Bu_1+u_2)/\Lambda$ give
$$\theta_j=\Theta(t^{1/L}), \qquad j\in[L+1].$$
Since $p_{u_2}=P(\bfb)e^{-q_1}/\Lambda$ and
$P(\bfb)=\Theta(t^{1-1/L})$, we have
$$
    \log\left(e^{c_1t}p_{u_2}(t)\right)
    =\left(c_1-\frac{c_2}{2}\right)t+\smallo{t}.
$$
As $c_1>c_2/2$, this implies
$\int_0^\infty e^{c_1s}p_{u_2}(s)\,ds=\infty$.
It therefore holds that
$$\frac{m_{u_1}}{m_{u_2}} \toinfty{t} B.$$
And so
$$\frac{\dot u_1}{\dot u_2} \toinfty{t} B\sqrt{\frac{\omega_{u_2}}{\omega_{u_1}}} = \delta_0.$$
Recall $u_1,u_2 \toinfty{t} \infty$, implying by L'Hospital's rule that
$$\frac{u_1}{u_2} \toinfty{t} \delta_0.$$

For the parameters $b_j$, recall that $b_j=\Theta(t^{1/L})$ for each $j$. Moreover,
$$\frac{p_{b_j}}{p_{u_2}} = \frac{\Lambda\xi_1}{b_j}\left(1+\frac{q_2}{q_1}e^{-(q_2-q_1)}\right) = \frac{\Lambda\xi_1}{b_j}\left(1 + \smallo{1}\right).$$
Indeed $q_2=\bigO{t}$ by the parameter bounds while $q_1=\Theta(t)$,
so $q_2/q_1=\bigO{1}$; the extra term tends to zero because
$q_2-q_1\to\infty$.
We justify the use of Lemma~\ref{lemma:ema_with_multiplier} with
$g=p_{u_2}$ and $A=\frac{\Lambda\xi_1}{b_j}$.
The preceding logarithmic estimate verifies its first condition.
Also, since parameter time derivatives and parameters are respectively
$\bigO{t^{1/L-1}}, \Omega(t^{1/L})$,
$$\abs{\frac{d}{dt}\log \frac{\xi_1}{b_j}} \leq \abs{\frac{d}{dt} \log \xi_1} + \abs{\frac{d}{dt}\log b_j}\leq\bigO{\frac1t}.$$
Therefore, by Lemma~\ref{lemma:ema_with_multiplier},
$$\frac{m_{b_j}}{m_{u_2}} = \frac{\Lambda\xi_1}{b_j}\left(1 + \smallo{1}\right).$$
Thus,
$$\frac{\dot b_j}{\dot u_2} = \sqrt{\frac{\omega_{u_2}}{\omega_{b_j}}}\frac{\Lambda\xi_1}{b_j}(1+\smallo{1}).$$
Hence,
$$\frac{\frac{d}{dt}b_j^2}{\frac{d}{dt}u_2^2} = \frac{b_j \dot b_j}{u_2\dot u_2} =\sqrt{\frac{\omega_{u_2}}{\omega_{b_j}}}\frac{\Lambda\xi_1}{u_2}\left(1 + \smallo{1}\right) \toinfty{t} (B\delta_0+1)\sqrt{\frac{\omega_{u_2}}{\omega_{b_j}}}.$$
Both $b_j^2$ and $u_2^2$ tend to infinity, so again applying L'Hospital's rule and taking the root gives
$$\frac{b_j}{u_2} \toinfty{t} \sqrt{B\delta_0+1}\left(\frac{\omega_{u_2}}{\omega_{b_j}}\right)^{1/4} = \delta_j.$$

Finally, recalling that $\qmin=q_1$, the margin-normalized limit is
simply obtained by normalizing
$\bar\thetab=(\delta_0,1,\delta_1,\ldots,\delta_{L-1})$
by $\qmin^{1/L}(\bar\thetab)=q_1^{1/L}(\bar\thetab)$:
$$\hat\thetab=\lim_{t\to\infty}\frac{\thetab_t}{\qmin^{1/L}(\thetab_t)}
=\frac{\bar\thetab}{q_1^{1/L}(\bar\thetab)}
=\left(\frac1\Lambda(B\delta_0+1)\prod_{j=1}^{L-1}\delta_j\right)^{-1/L}
(\delta_0,1,\delta_1,\ldots,\delta_{L-1}).$$
\end{proof}
\subsubsection{Direction Reachability}
We now prove Proposition~\ref{prop:directional_convergence}. Note that we have already proved a dependence of the convergent direction on the initialization, but we are still missing an argument showing that all desired directions on $\Sigma_{B,\Lambda}$ are indeed reached. We use (without proof) the Poincare-Miranda Theorem, a generalization of the intermediate value theorem:

\begin{theorem}[Poincare-Miranda]\label{thm:poincare_miranda}
    Let $F = (F_1,...,F_n) : \mathcal B \to \bbR^n$ be continuous where $\mathcal B = \prod_{j=1}^n[a_j,b_j] \subseteq \bbR^n$ for $a_j < b_j, j \in [n]$. Let $\y \in \bbR^n$. Assume 
    \begin{align*}
      \forall j \in [n], \x = (x_1,...,x_n) \in \mathcal B: x_j &= a_j \Longrightarrow F_j(\x) \leq y_j \\
      x_j &= b_j \Longrightarrow F_j(\x) \geq y_j
    \end{align*}
    Then there exists $\x \in \mathcal B$ with $F(\x) = \y$.
\end{theorem}

\begin{proof}[Proof of Proposition~\ref{prop:directional_convergence}]

Fix $B>1$, $\Lambda>0$, and a nonempty compact
$K\subseteq\Sigma_{B,\Lambda}$.

For $\hat\thetab=(\hat u_1,\hat u_2,\hat b_1,\ldots,\hat b_{L-1})\in K$, set

\[
    \delta_0:=\frac{\hat u_1}{\hat u_2}\in(0,1),
    \qquad
    \delta_j:=\frac{\hat b_j}{\hat u_2}>0,
    \qquad j\in[L-1],
\]
and denote
\[
    \bar\thetab:=(\delta_0,1,\delta_1,\ldots,\delta_{L-1}).
\]

The ratios depend continuously on $\hat\thetab\in K$. Hence
\[
    0<\delta_-:=\min_K\delta_0
    \leq\max_K\delta_0=:\delta_+<1.
\]

We will choose an initialization whose limiting direction is
$\bar\thetab$, and then apply
Lemma~\ref{lemma:frozen_weights_determine_direction}
to obtain the desired margin-normalized limit.

We first restrict ourselves to a compact set of initialization parameters.
Use the coordinates
\[
    \iota_j:=\frac12\log\frac{\pi_j}{\pi_0},
    \qquad
    \pi_0=\frac{\Lambda}{\prod_{j=1}^{L-1}\pi_j}.
\]
This change of coordinates is invertible, with
\[
    \pi_0(\iota)
    =\Lambda^{1/L}\exp\left(-\frac2L\sum_{j=1}^{L-1}\iota_j\right),
    \qquad
    \pi_j(\iota)=\pi_0(\iota)e^{2\iota_j}.
\]
For each $j\in[L-1]$, set
\[
    \iota_j^*(\hat\thetab)
    :=\log\delta_j-\frac12\log(1+B\delta_0),
\]
and choose
\[
    A_j>
    \frac12\log(1+B)
    +\frac14\log8.
\]
Define
\[
    J_j:=\left[\min_K\iota_j^*-A_j,\max_K\iota_j^*+A_j\right],
    \qquad
    \Pi
    :=\left\{
        \pi(\iota)\;\middle|\;
        \iota\in\prod_{j=1}^{L-1}J_j
      \right\}.
\]
The set $\Pi$ is a compact subset of
$(0,\infty)^{L-1}$ and is independent of $M$.

Let $k_{\Pi}>0$ be the constant from
Lemma~\ref{lemma:second_moment_weights_bounds}.
Since this constant is independent of $M$, we may choose
$M_*>\max\{1,M_0(B)\}$ so that, for every $M\geq M_*$,

\[
    \frac{B}{\sqrt{k_{\Pi}M}}<\delta_-,
    \qquad
    \frac{1}{1+(BM^2)^{-1}}>\delta_+ \ .
\]

Fix any such $M$, and then fix any
$\rho\geq\rho_1(M,\Pi)$.

We now construct the continuous map to which we apply
Theorem~\ref{thm:poincare_miranda}.
On the box
\[
    \mathcal B
    :=[0,3\log M]\times\prod_{j=1}^{L-1}J_j,
\]
write $\omega_i(z,\iota)$ for the weight corresponding to
the initialization $(\rho,z,\pi(\iota))$, and define
\[
    d_0(z,\iota)
    :=B\sqrt{\frac{\omega_{u_2}(z,\iota)}
                   {\omega_{u_1}(z,\iota)}},
\]
\[
    d_j(z,\iota)
    :=\sqrt{1+Bd_0(z,\iota)}
      \left(
        \frac{\omega_{u_2}(z,\iota)}
             {\omega_{b_j}(z,\iota)}
      \right)^{1/4},
    \qquad j\in[L-1].
\]
All weights are positive and continuous by
Lemma~\ref{lemma:continuity_weights}, so
$d=(d_0,\ldots,d_{L-1})$ is continuous on $\mathcal B$.

Now fix any target $\hat\thetab\in K$, suppressing its dependence
in $\delta_j$ and $\iota_j^*$.
We first check the two faces in the $z$ coordinate.
Lemma~\ref{lemma:second_moment_weights_bounds} gives
\[
    d_0(0,\iota)
    \leq\frac{B}{\sqrt{k_{\Pi}M}}
    <\delta_0.
\]
At $z=3\log M$, the same lemma gives
\[
    \frac{\omega_{u_1}}{\omega_{u_2}}
    \leq(B+Me^{-3\log M})^2
    =(B+M^{-2})^2,
\]
and hence
\[
    d_0(3\log M,\iota)
    \geq\frac{B}{B+M^{-2}}
    =\frac{1}{1+(BM^2)^{-1}}
    >\delta_0.
\]
Moreover, the lower bound
$\omega_{u_1}/\omega_{u_2}\geq B^2$ shows that
$0<d_0\leq1$ throughout $\mathcal B$.

For the remaining faces, the definition of $\iota_j^*$ gives
\[
\begin{aligned}
    \log d_j-\log\delta_j
    &=
      \iota_j-\iota_j^*
      +\frac12\log\frac{1+Bd_0}{1+B\delta_0}
      +\frac14\log
       \frac{\pi_0^2\omega_{u_2}}
            {\pi_j^2\omega_{b_j}}.
\end{aligned}
\]
By Lemma~\ref{lemma:second_moment_weights_bounds},
\[
    \frac18
    \leq
    \frac{\pi_0^2\omega_{u_2}}
         {\pi_j^2\omega_{b_j}}
    \leq8.
\]
Together with $0<d_0\leq1$, this implies
\[
\begin{aligned}
    &\left|
      \frac12\log\frac{1+Bd_0}{1+B\delta_0}
      +\frac14\log
       \frac{\pi_0^2\omega_{u_2}}
            {\pi_j^2\omega_{b_j}}
    \right|\\
    &\qquad\leq
      \frac12\log(1+B)
      +\frac14\log8
      <A_j.
\end{aligned}
\]

Since $\min_K\iota_j^*\leq\iota_j^*\leq\max_K\iota_j^*$, the same
estimate gives
\[
\begin{aligned}
    \iota_j=\min_K\iota_j^*-A_j
    &\Longrightarrow d_j<\delta_j,\\
    \iota_j=\max_K\iota_j^*+A_j
    &\Longrightarrow d_j>\delta_j.
\end{aligned}
\]

All face inequalities required by
Theorem~\ref{thm:poincare_miranda} now hold.
Thus there exists a common point
$(z^\dagger,\iota^\dagger)\in\mathcal B$ such that
\[
    d_j(z^\dagger,\iota^\dagger)=\delta_j,
    \qquad j=0,\ldots,L-1.
\]
Use the initialization corresponding to
$\pi^\dagger=\pi(\iota^\dagger)$ and $z^\dagger$.
Lemma~\ref{lemma:frozen_weights_determine_direction} applies
and gives
\[
    \frac{\thetab_t}{u_2(t)}
    \toinfty{t}\bar\thetab,
    \qquad
    \frac{\thetab_t}{\qmin^{1/L}(\thetab_t)}
    \toinfty{t}
    \frac{\bar\thetab}
         {f(\x_1;\bar\thetab)^{1/L}}.
\]
Finally, $\bar\thetab=\hat\thetab/\hat u_2$ and
$f(\x_1;\hat\thetab)=1$. By positive homogeneity,
\[
    f(\x_1;\bar\thetab)=\hat u_2^{-L},
    \qquad
    \frac{\bar\thetab}
         {f(\x_1;\bar\thetab)^{1/L}}
    =\hat\thetab.
\]
This proves the desired convergence for every $M\geq M_*$. 

For continuous dependence, Lemma~\ref{lemma:frozen_weights_determine_direction} gives the
margin-normalized limit throughout $\mathcal B$ as
\[
    \hat \thetab(z, \iota)
    :=\frac{(d_0,1,d_1,\ldots,d_{L-1})}
            {f(\x_1;(d_0,1,d_1,\ldots,d_{L-1}))^{1/L}},
\]
which is continuous in $z, \iota$ and has image containing $K$. Finally, with $B,\Lambda,M,\rho$ fixed, $(z,\iota)$ can be recovered
continuously from $\thetab_0$ through~\eqref{eq:initialization}.
Hence the limiting direction depends continuously on initialization
within this family.

For $L=1$, the hidden-layer coordinates and their face conditions
are absent, $\pi_0=\Lambda$, and the same argument reduces to the
intermediate value theorem for $d_0$ on $[0,3\log M]$.
\end{proof}

\subsubsection{Signed Permutation Symmetries}\label{subsection:symmetries}

We now extend the construction from the positive sector to every sign-and-order
sector of parameter space.  In this section, the model~\eqref{eq:construction_model_appendix} remains
fixed; we transform only the dataset and the initialization.  All
other ingredients of training, including the loss, the learning-rate schedule,
and the Adam parameters, remain unchanged.

Recall the positive sector 
\begin{equation}\label{eq:positive_sector_for_symmetries}
    \Theta_+
    :=\left\{
       (\ub,\bfb)\in\bbR^{L+1}
       \;\middle|\;
       0<u_1<u_2,\quad b_j>0\ \text{for every }j\in[L-1]
     \right\}.
\end{equation}
Let
\begin{equation}\label{eq:signed_permutation_group}
    \mathcal S_2^\pm
    :=\left\{
       \Xi\in\{0,-1,1\}^{2\times2}
       \;\middle|\;
       \Xi^\top \Xi=I_2
     \right\}
\end{equation}
be the set of signed permutation matrices on the two input coordinates.  Thus,
$\Xi\in\mathcal S_2^\pm$ may exchange the two coordinates, change either of their
signs, or do both.  For
$\sigma=(\sigma_1,\ldots,\sigma_{L-1})
\in\{-1,1\}^{L-1}$, define
\begin{equation}\label{eq:symmetry_parity_and_map}
    \chi_\sigma
    :=\prod_{r=1}^{L-1}\sigma_r,
    \qquad
    T_{\Xi,\sigma}
    :=\operatorname{diag}
      \left(\Xi,\sigma_1,\ldots,\sigma_{L-1}\right).
\end{equation}
Equivalently,
\[
    T_{\Xi,\sigma}(\ub,\bfb)
    =(\Xi\ub,\sigma\odot\bfb).
\]
When $L=1$, the vector $\sigma$ is empty and the empty product is
understood to be $\chi_\sigma=1$.  Finally, let
\begin{equation}\label{eq:unsigned_symmetry_matrix}
    \Pi_{\Xi,\sigma}
    :=T_{\Xi,\sigma}\odot T_{\Xi,\sigma};
\end{equation}
this is the unsigned permutation matrix obtained by deleting the signs from
$T_{\Xi,\sigma}$.

Given a dataset $S=\{(\x_i,y_i)\}_{i=1}^m$, define 
\begin{equation}\label{eq:transformed_dataset_general}
    S^{\Xi,\sigma}
    :=\left\{
       \left(\Xi\x_i,\chi_\sigma y_i\right)
     \right\}_{i=1}^m.
\end{equation}
Thus the signs and order of the first-layer parameters are absorbed into the
input coordinates, while all labels are flipped exactly when an odd number of
the scalar-layer parameters are sign-flipped.

For $\Xi\in\mathcal S_2^\pm$ and
$\sigma\in\{-1,1\}^{L-1}$, define the corresponding sign-and-order
sector by
\begin{equation}\label{eq:transformed_sector}
    \Theta_{\Xi,\sigma}
    :=T_{\Xi,\sigma}\Theta_+.
\end{equation}
The sets $\Theta_{\Xi,\sigma}$ are exactly the open sectors obtained by
fixing the signs of all parameters and fixing which of $\abs{u_1}$ and
$\abs{u_2}$ is smaller.

\begin{lemma}[Dataset symmetries imply trajectory symmetries]
\label{lemma:adam_symmetries}
Let $(\thetab_t,\m_t,\vb_t)$ solve the bias-corrected Adam equations
for the fixed model~\eqref{eq:construction_model_appendix}
on a dataset $S$, with initialization $\thetab_0$ and zero moment buffers.
Fix $\Xi\in\mathcal S_2^\pm$ and
$\sigma\in\{-1,1\}^{L-1}$, and set
\begin{equation}\label{eq:transformed_adam_state}
    \widetilde\thetab_t
    :=T_{\Xi,\sigma}\thetab_t,
    \qquad
    \widetilde\m_t
    :=T_{\Xi,\sigma}\m_t,
    \qquad
    \widetilde\vb_t
    :=\Pi_{\Xi,\sigma}\vb_t.
\end{equation}
Then $(\widetilde\thetab_t,\widetilde\m_t,\widetilde\vb_t)$ solves the same
Adam equations for the same model $f$ on the transformed dataset
$S^{\Xi,\sigma}$, with initialization
$T_{\Xi,\sigma}\thetab_0$.

Moreover, for every $\thetab=(\ub,\bfb)$ and every $(\x_i,y_i)\in S$,
\begin{equation}\label{eq:signed_margin_invariance}
    (\chi_\sigma y_i)
    f\left(\Xi\x_i;T_{\Xi,\sigma}\thetab\right)
    =y_if(\x_i;\thetab).
\end{equation}
Consequently, if
\[
    \qmin^S(\thetab)
    :=\min_{i\in[m]}y_if(\x_i;\thetab),
\]
then
\begin{equation}\label{eq:qmin_symmetry}
    \qmin^{S^{\Xi,\sigma}}
       \left(T_{\Xi,\sigma}\thetab\right)
    =\qmin^S(\thetab).
\end{equation}
In particular, directional limits with respect to $\norm{\cdot}_2$ and margin-normalized limits are transported
by $T_{\Xi,\sigma}$, and each of the assumptions~\ref{traj_ass:positive_margin_interpolation},~\ref{traj_ass:loss_decay} holds for the trajectory $\thetab_t$ if and only if it holds for $\tilde \thetab_t$.
\end{lemma}

\begin{proof}
Since $\Xi^\top \Xi=I_2$ and
$P(\sigma\odot\bfb)=\chi_\sigma P(\bfb)$, we have
\begin{align*}
    (\chi_\sigma y_i)
    f\left(\Xi\x_i;T_{\Xi,\sigma}\thetab\right)
    &=(\chi_\sigma y_i)
      P(\sigma\odot\bfb)
      \innerprod{\Xi\ub}{\Xi\x_i}\\
    &=\chi_\sigma^2
      y_iP(\bfb)\innerprod{\ub}{\x_i}
      =y_if(\x_i;\thetab),
\end{align*}
which proves~\eqref{eq:signed_margin_invariance}.  Hence, if $\Lcal_S$ and
$\Lcal_{S^{\Xi,\sigma}}$ denote the empirical losses of the original and
transformed datasets, respectively, then
\begin{equation}\label{eq:loss_symmetry}
    \Lcal_{S^{\Xi,\sigma}}
      \left(T_{\Xi,\sigma}\thetab\right)
    =\Lcal_S(\thetab).
\end{equation}
Differentiating~\eqref{eq:loss_symmetry} gives
\[
    T_{\Xi,\sigma}^\top
    \nabla\Lcal_{S^{\Xi,\sigma}}
       \left(T_{\Xi,\sigma}\thetab\right)
    =\nabla\Lcal_S(\thetab).
\]
Because $T_{\Xi,\sigma}$ is orthogonal,
\begin{equation}\label{eq:gradient_symmetry}
    \nabla\Lcal_{S^{\Xi,\sigma}}
       \left(T_{\Xi,\sigma}\thetab\right)
    =T_{\Xi,\sigma}\nabla\Lcal_S(\thetab).
\end{equation}
Therefore, writing $\g_t=\nabla\Lcal_S(\thetab_t)$ and
$\widetilde\g_t
=\nabla\Lcal_{S^{\Xi,\sigma}}(\widetilde\thetab_t)$,
\begin{equation}\label{eq:gradient_square_symmetry}
    \widetilde\g_t=T_{\Xi,\sigma}\g_t,
    \qquad
    \widetilde\g_t^2=\Pi_{\Xi,\sigma}\g_t^2,
\end{equation}
where the square is coordinate-wise.  It follows that
\begin{align*}
    \dot{\widetilde\m}_t
    &=T_{\Xi,\sigma}\dot\m_t
      =c_1\left(\widetilde\g_t-\widetilde\m_t\right),\\
    \dot{\widetilde\vb}_t
    &=\Pi_{\Xi,\sigma}\dot\vb_t
      =c_2\left(\widetilde\g_t^2-\widetilde\vb_t\right).
\end{align*}
The zero initial conditions are preserved, and the scalar bias-correction
factors commute with the two transformations, so
\[
    \widehat{\widetilde\m}_t
    =T_{\Xi,\sigma}\hat\m_t,
    \qquad
    \widehat{\widetilde\vb}_t
    =\Pi_{\Xi,\sigma}\hat\vb_t.
\]
For every signed permutation matrix $T$ and every vector $\mathbf z$ and coordinate-wise positive
vector $\w$,
\[
    \frac{T\mathbf z}{\sqrt{(T\odot T)\w}}
    =T\frac{\mathbf z}{\sqrt \w}.
\]
Applying this identity with $T=T_{\Xi,\sigma}$ yields, for $t>0$,
\begin{equation}\label{eq:adam_ratio_symmetry}
    \frac{\widehat{\widetilde\m}_t}
         {\sqrt{\widehat{\widetilde\vb}_t}}
    =T_{\Xi,\sigma}
      \frac{\hat\m_t}{\sqrt{\hat\vb_t}}.
\end{equation}
If a second-moment coordinate is zero, its corresponding first moment is
also zero, and~\eqref{eq:adam_ratio_symmetry} holds under the
zero-denominator convention.
Thus
\[
    \dot{\widetilde\thetab}_t
    =T_{\Xi,\sigma}\dot\thetab_t
    =-\eta(t)
      \frac{\widehat{\widetilde\m}_t}
           {\sqrt{\widehat{\widetilde\vb}_t}},
\]
which is precisely the Adam parameter equation for the transformed dataset.

Equation~\eqref{eq:qmin_symmetry} follows immediately from
\eqref{eq:signed_margin_invariance}.  Since $T_{\Xi,\sigma}$ is orthogonal,
whenever $\thetab_t\neq\zero$,
\[
    \frac{\widetilde\thetab_t}{\norm{\widetilde\thetab_t}_2}
    =T_{\Xi,\sigma}
      \frac{\thetab_t}{\norm{\thetab_t}_2}.
\]
Whenever the minimum margin is positive, we likewise have
\begin{equation}\label{eq:margin_normalization_symmetry}
    \frac{\widetilde\thetab_t}
    {\qmin^{S^{\Xi,\sigma}}(\widetilde\thetab_t)^{1/L}}
    =T_{\Xi,\sigma}
      \frac{\thetab_t}{\qmin^S(\thetab_t)^{1/L}}.
\end{equation}
This proves the transportation of directional and margin-normalized limits. For equivalence of assumptions~\ref{traj_ass:positive_margin_interpolation},~\ref{traj_ass:loss_decay}, the losses agree by~\eqref{eq:loss_symmetry}, while
orthogonality and~\eqref{eq:qmin_symmetry} give, whenever
$\thetab_t\neq\zero$,
\[
    \frac{\qmin^{S^{\Xi,\sigma}}
        (\widetilde\thetab_t)}
         {\norm{\widetilde\thetab_t}_2^L}
    =
    \frac{\qmin^S(\thetab_t)}
         {\norm{\thetab_t}_2^L}.
\]
Thus the two trajectories have identical losses and normalized
margins, proving both equivalences.
\end{proof}

\begin{corollary}[Continuous directional limits on a fixed dataset]
\label{cor:compact_fixed_dataset_reachability}
Under the assumptions of Theorem~\ref{thm:no_regularizer_narrow_main_text},
let $p=L+1$ and define
\[
    U_{\Xi,\sigma}
    :=\Theta_{\Xi,\sigma}\cap\mathbb S^{p-1},
    \qquad
    \Xi\in\mathcal S_2^\pm,\quad
    \sigma\in\{-1,1\}^{L-1}.
\]
These $2^{L+2}$ sets are open in the sphere, and their union has full
surface measure. For every compact $A\subseteq U_{\Xi,\sigma}$,
there is one fixed two-point dataset $S_A$ such that every point of $A$
is the limit of $\thetab_t/\norm{\thetab_t}_2$ for a suitable
initialization. Moreover, the initializations can be taken from a compact
family on which this limiting direction exists and depends continuously
on the initialization.
\end{corollary}

\begin{proof}
The sectors fix the signs of all coordinates and the strict order of
$\abs{u_1},\abs{u_2}$, giving $2^{L+1}\cdot2=2^{L+2}$ open spherical
sets. Their complement has zero measure.

By Lemma~\ref{lemma:adam_symmetries}, it suffices to consider a nonempty
compact $A\subseteq\Theta_+\cap\mathbb S^{p-1}$.
Fix any $B>1$, $\Lambda>0$. Since $f(\x_1;\bar \thetab)>0$ on this sector,
\[
    K:=\left\{
       \frac{\bar\thetab}{f(\x_1;\bar\thetab)^{1/L}}
       \;\middle|\;\bar\thetab\in A
       \right\}
    \subseteq\Sigma_{B,\Lambda}
\]
is compact. Proposition~\ref{prop:directional_convergence} supplies one
fixed $M$ and a compact initialization family whose continuous
margin-normalized limits contain $K$.
The margin-normalized limits have minimum margin $1$, so they are nonzero,
and normalizing them to unit norm preserves continuity and reaches every
point of $A$.
\end{proof}

\begin{corollary}[Constancy on all sign-and-order sectors]
\label{cor:adam_symmetries_for_our_proof}
Assume $\alpha>\frac1L-\frac12$.
For every $\Xi\in\mathcal S_2^\pm$ and
$\sigma\in\{-1,1\}^{L-1}$, any implicit regularizer $R$ of Adam on the width-1 model with the exponential loss is constant on
$\Theta_{\Xi,\sigma}$.  All of these sector constants equal $R(\zero)$.
Consequently, $R$ is constant on $\bbR^{L+1}$.
\end{corollary}
\begin{proof}
Let $R$ be an implicit margin regularizer for $f$ with the exponential loss. Fix $\Xi\in\mathcal S_2^\pm$ and
$\sigma\in\{-1,1\}^{L-1}$, and write
$T:=T_{\Xi,\sigma}$. Define
\[
    R_T(\thetab):=R(T\thetab).
\]
We first verify that $R_T$ is an implicit margin regularizer
for $f$. Since $T$ is invertible, $R_T$ is locally
Lipschitz and admits a chain rule, with
\[
    \partial R_T(\thetab)
    =T^\top\partial R(T\thetab).
\]
Consider any Adam trajectory $\thetab_t$ on a dataset $S$
satisfying~\ref{traj_ass:positive_margin_interpolation},~\ref{traj_ass:loss_decay} in
Definition~\ref{def:implicit_regularizer}, with
margin-normalized limit $\hat\thetab$.
By Lemma~\ref{lemma:adam_symmetries},
$T\thetab_t$ is an Adam trajectory on $S^{\Xi,\sigma}$
with margin-normalized limit $T\hat\thetab$, and satisfying~\ref{traj_ass:positive_margin_interpolation},~\ref{traj_ass:loss_decay}.
Writing $q_i$ and $\widetilde q_i$ for the original and
transformed margins, respectively, we have
\[
    \widetilde q_i(T\thetab)=q_i(\thetab),
    \qquad
    T^\top\nabla\widetilde q_i(T\thetab)
    =\nabla q_i(\thetab).
\]
Because $R$ is an implicit regularizer, its KKT conditions
at $T\hat\thetab$ give
\[
\begin{gathered}
    \kb=\sum_i\lambda_i\nabla\widetilde q_i(T\hat\thetab),
    \qquad
    \kb\in\partial R(T\hat\thetab),\\
    \forall i:\quad
    \widetilde q_i(T\hat\thetab)\geq1,\qquad
    \lambda_i\geq0,\qquad
    \lambda_i\bigl(\widetilde q_i(T\hat\thetab)-1\bigr)=0.
\end{gathered}
\]
Multiplying stationarity by $T^\top$ and using
$\widetilde q_i(T\hat\thetab)=q_i(\hat\thetab)$ yields
\[
\begin{gathered}
    T^\top\kb=\sum_i\lambda_i\nabla q_i(\hat\thetab),
    \qquad
    T^\top\kb\in\partial R_T(\hat\thetab),\\
    \forall i:\quad
    q_i(\hat\thetab)\geq1,\qquad
    \lambda_i\geq0,\qquad
    \lambda_i\bigl(q_i(\hat\thetab)-1\bigr)=0.
\end{gathered}
\]
Thus $\hat\thetab$ is KKT for the
original margin constraints with objective $R_T$, proving
that $R_T$ is an implicit regularizer.

Lemma~\ref{lemma:constancy_on_sigma_BC}, applied to $R_T$,
makes $R_T$ constant on every $\Sigma_{B,\Lambda}$.
The positive-sector argument in the proof of
Theorem~\ref{thm:no_adam_reg}, applied to $R_T$, which uses only Lemma~\ref{lemma:constancy_on_sigma_BC}
and continuity of $R_T$, therefore makes $R_T$ constant on
$\Theta_+$. Hence $R$ is constant on
$T\Theta_+=\Theta_{\Xi,\sigma}$.

For any $\thetab\in\Theta_+$, the entire ray
$\{r\thetab:r>0\}$ lies in $\Theta_+$, so continuity gives
\[
    R(T\thetab)
    =R_T(\thetab)
    =\lim_{r\downarrow0}R_T(r\thetab)
    =R(0).
\]
Thus all sector constants equal $R(0)$.
The sectors have dense union in $\bbR^{L+1}$, and continuity
extends this equality to the whole space.
\end{proof}

\subsection{Arbitrary Widths and Input Dimension}\label{sec:arbitrary_widths}

In this section we generalize the width-$1$ construction to a scalar-output linear network of arbitrary width and input dimension.

Consider a depth-\(L\) scalar-output linear network. Let its input dimension
be \(d\), and let its hidden widths be \(n_1,\ldots,n_{L-1}\). Its parameter
space for \(L\geq2\) is
\[
 \mathcal V=\bbR^{n_1\times d}\times
 \prod_{\ell=2}^{L-1}\bbR^{n_\ell\times n_{\ell-1}}
 \times\bbR^{1\times n_{L-1}}.
\]
For \(L=1\), set \(\mathcal V=\bbR^{1\times d}\).
Assume throughout that \(L\geq1\), \(d\geq2\), and
\(n_\ell\in\mathbb N_{>0}\) for every \(1\leq\ell<L\).
Set once and for all
\[
 \kappa:=\prod_{r=1}^{L-1}n_r.
\]
Define the narrow parameter space and distinguish its parameter from the
wide one by
\[
 \Theta:=\bbR^2\times\bbR^{L-1}\cong\bbR^{L+1},
 \qquad
 \thetab=(\ub,\bfb)\in\Theta,
 \qquad
 \vartheta=(W_1,\ldots,W_L)\in\mathcal V.
\]
We retain the notation $f(\x; \thetab)=P(\bfb)\ub^\top\x$ for the width-$1$ model, and denote the wide model by
\(F(\xi;\vartheta)\):
\[
 F(\xi;\vartheta):=W_L\cdots W_1\xi\ .
\]

When \(L=1\), \(\bfb\) and the sign tuple \(\mathbf s\) below are empty, and the empty products give \(P=\kappa=1\).

The lifting is based on the following principle: choose a sign vector
\(\mathbf s_\ell\in\{\pm1\}^{n_\ell}\) for every hidden layer.  Its
\(i\)-th scalar entry is denoted by \((\mathbf s_\ell)_i\).  During
training, all neurons in a layer carry signed copies of the same narrow
scalar.  For the input-dimension lift, additionally choose
\(c\in[d]:=\{1,\ldots,d\}\) and, for \(\x,\ub\in\bbR^2\), define
\begin{equation}\label{eq:combined_input_maps}
 U_c,X_c:\bbR^2\to\bbR^d,
 \qquad
 U_c(\ub)=u_1e_c+\sum_{c'\neq c}u_2e_{c'},
 \qquad
 X_c(\x)=x_1e_c+\frac1{d-1}\sum_{c'\neq c}x_2e_{c'}.
\end{equation}
The two maps are paired so that
\begin{equation}\label{eq:input_copy_duality}
 U_c(\ub)^\top X_c(\x)=u_1x_1+u_2x_2=\ub^\top\x.
\end{equation}

Formally, for
\(\mathbf s=(\mathbf s_1,\ldots,\mathbf s_{L-1}) \in \{\pm 1\}^{n_1} \times ... \times \{\pm 1\}^{n_{L-1}}\), define the linear map
\(E_{\mathbf s,c}:\Theta\to\mathcal V\),
\(\thetab\mapsto\vartheta\), by the following formulas when \(L\geq2\):
\begin{align}
 W_1&=(E_{\mathbf s,c}\thetab)_1
      =\mathbf s_1U_c(\ub)^\top,\label{eq:lift_first_layer}\\
 W_\ell&=(E_{\mathbf s,c}\thetab)_\ell
      =b_{\ell-1}\mathbf s_\ell\mathbf s_{\ell-1}^\top,
      &&2\leq\ell\leq L-1,\label{eq:lift_middle_layers}\\
 W_L&=(E_{\mathbf s,c}\thetab)_L
      =b_{L-1}\mathbf s_{L-1}^\top.\label{eq:lift_output_layer}
\end{align}
For \(L=1\), define instead \(E_{\mathbf s,c}(\ub):=U_c(\ub)^\top\) (note $\mathbf s$ has no role here).
Denote the image of $E_{\mathbf s,c}$ by
\[
 V_{\mathbf s,c}:=\operatorname{im}E_{\mathbf s,c}.
\]

The lift is injective.  Indeed, for \(L\geq2\), its first block determines \(U_c(\ub)\),
hence \(\ub\), every middle block determines the corresponding scalar
\(b_r\), and the final block determines \(b_{L-1}\).  For \(L=1\), this follows from injectivity of \(U_c\). Thus
\(E_{\mathbf s,c}:\Theta\to V_{\mathbf s,c}\) is a linear bijection.

\begin{lemma}[Spanning of the combined copy subspaces]
\label{lem:common-spanning-copies}
The subspaces of the lift span the full wide parameter space:
\begin{equation}
 \operatorname{span}\!\left(
   \bigcup_{c=1}^d\ \bigcup_{\mathbf s}V_{\mathbf s,c}
 \right)=\mathcal V.
\label{eq:copy_subspaces_span}
\end{equation}
\end{lemma}

\begin{proof}
For \(L=1\), \(E_{\varnothing,c}((1,0))=e_c^\top\), so varying \(c\) spans \(\mathcal V\). Assume henceforth that \(L\geq2\).
Since \(U_c((1,0))=e_c\),
\[
 E_{\mathbf s,c}((1,0),0,\ldots,0)
 =(\mathbf s_1e_c^\top,0,\ldots,0).
\]
For fixed \(c\), varying \(\mathbf s_1\) spans the \(c\)-th input column,
and varying \(c\) spans the whole first matrix.  For
\(1\leq \ell \leq L-2\), setting only \(b_\ell\) nonzero isolates
\(b_\ell\mathbf s_{\ell+1}\mathbf s_\ell^\top\) in matrix \(\ell+1\); setting only
\(b_{L-1}\) nonzero isolates
\(b_{L-1}\mathbf s_{L-1}^\top\) in the output matrix.

Sign vectors span every coordinate space because, for a basis vector $\e_i^{(n)} \in \bbR^n$,
\[
\e_i^{(n)}=\frac12\left(\mathbf1_n-
   (\mathbf1_n-2\e_i^{(n)})\right),
\]
and both vectors on the right are sign vectors.  Products of these
representations express every middle-layer matrix unit $\e_i^{(\ell)}{\e_{j}^{(\ell-1)}}^\top$ as a linear
combination of four signed outer products.  Hence every matrix unit in every
block belongs to the span in \eqref{eq:copy_subspaces_span}.
\end{proof}

\begin{lemma}[Convex propagation]
\label{lem:common-convex-propagation}
Let \(R:\mathcal V\to\bbR\) be finite and convex.  If
\[
 R|_{V_{\mathbf s,c}}\equiv R(0)
 \qquad\text{for every \(\mathbf s\) and \(c\)},
\]
then \(R\) is constant on \(\mathcal V\).
\end{lemma}

\begin{proof}
By Lemma~\ref{lem:common-spanning-copies}, write any
\(X\in\mathcal V\) as \(X=\sum_{a=1}^N X_a\), with each \(X_a\) in one
lift subspace.  Since those subspaces are linear, convexity gives
\[
 R(X)=R\!\left(\frac1N\sum_{a=1}^N NX_a\right)
 \leq\frac1N\sum_{a=1}^NR(NX_a)=R(0).
\]
The same inequality holds for \(-X\).  Convexity at the midpoint of
\(X\) and \(-X\) then forces equality, so \(R(X)=R(0)\).
\end{proof}

The next lemma identifies the wide gradient by combining the chain rule
with the fact that the gradient lies in the copy subspace.

\begin{lemma}[Model and gradient identities]
\label{lem:common-model-gradient-identities}
Fix \(\mathbf s\) and \(c\in[d]\), let \(E:=E_{\mathbf s,c}\), and put
\[
 E_{\mathbf s,c}^\sharp:=E^\sharp
 :=E(E^\top E)^{-1}.
\]

Let $S^{\mathrm n},S^{c,\mathrm w}$ denote the narrow and wide datasets,
\[
 S^{\mathrm n}:=\{(\x_i,y_i)\}_{i=1}^{m}
 \subseteq\bbR^2\times\{\pm1\},
 \qquad
 S^{c,\mathrm w}:=
 \left\{\left(X_c(\x_i)/\kappa,y_i\right)\right\}_{i=1}^{m},
\]
Define
\begin{align*}
 \mathcal L^{\mathrm n}_S(\thetab)
 &:=\sum_{i=1}^{m}\ell\bigl(y_if(\x_i;\thetab)\bigr),\\
 \mathcal L^{\mathrm w}_{S^{c,\mathrm w}}(\vartheta)
 &:=\sum_{i=1}^{m}
   \ell\!\left(
     y_iF\!\left(X_c(\x_i)/\kappa;\vartheta\right)\right).
\end{align*}
Then
\begin{equation}
 F\!\left(\frac{X_c(\x)}{\kappa};E_{\mathbf s,c}\thetab\right)
 =f(\x;\thetab)
 \qquad\text{for every }\x\in\bbR^2,
 \label{eq:combined_wide_model_agrees}
\end{equation}
and, at corresponding parameters, the gradients
\[
 \g^{\mathrm w}:=\nabla_\vartheta\mathcal L^{\mathrm w}_{S^{c,\mathrm w}}
       (E_{\mathbf s,c}\thetab),
 \qquad
 \g^{\mathrm n}:=\nabla_{\thetab}\mathcal L^{\mathrm n}_S(\thetab)
\]
satisfy
\begin{equation}
 \g^{\mathrm w}=E_{\mathbf s,c}^\sharp\g^{\mathrm n}.
 \label{eq:gradient_lift}
\end{equation}
Equivalently, for \(L\geq2\), their blocks are
\begin{align}
 \g_{W_1}^{\mathrm w}
 &=\frac1{n_1}\mathbf s_1
             X_c(\g_{\ub}^{\mathrm n})^\top,
 \label{eq:combined_first_gradient_copy}\\
 \g_{W_\ell}^{\mathrm w}
 &=\frac{g_{b_{\ell-1}}^{\mathrm n}}{n_\ell n_{\ell-1}}
   \mathbf s_\ell\mathbf s_{\ell-1}^\top,
 &&2\leq\ell\leq L-1,
 \label{eq:combined_middle_gradient_copy}\\
 \g_{W_L}^{\mathrm w}
 &=\frac{g_{b_{L-1}}^{\mathrm n}}{n_{L-1}}
   \mathbf s_{L-1}^\top.
 \label{eq:combined_last_gradient_copy}
\end{align}
Here
\(\g_{\ub}^{\mathrm n}:=
(g_{u_1}^{\mathrm n},g_{u_2}^{\mathrm n})^\top\), and the middle identity is
vacuous when \(L=2\).  For \(L=1\), the single block is \(\g_{W_1}^{\mathrm w}=X_c(\g_{\ub}^{\mathrm n})^\top\). In particular, every such wide gradient belongs to
\(V_{\mathbf s,c}\).  Thus parameters lift by \(E\), whereas gradients and
linear momentum buffers lift by \(E^\sharp\).
\end{lemma}

\begin{proof}
Set \(D:=E^\top E\).  For \(L=1\), \(D=\operatorname{diag}(1,d-1)\) and \(E^\sharp\ub=X_c(\ub)^\top\), so both identities follow from \eqref{eq:input_copy_duality} and direct differentiation. The remaining computations assume \(L\geq2\). Contracting adjacent sign vectors in the copied
weights and using \eqref{eq:input_copy_duality} gives
\begin{align*}
 W_L\cdots W_1&=\kappa P(\bfb)U_c(\ub)^\top,\\
 F\!\left(X_c(\x)/\kappa;E\thetab\right)
 &=P(\bfb)U_c(\ub)^\top X_c(\x)=f(\x;\thetab).
\end{align*}
Consequently, the losses agree under the lift, i.e. $\mathcal L^{\mathrm n}_S
 =\mathcal L^{\mathrm w}_{S^{c,\mathrm w}}\circ E$, and the chain rule gives
\begin{equation}\label{eq:g_n_w_chain_rule}
 \g^{\mathrm n}=E^\top\g^{\mathrm w}.   
\end{equation}
The wide gradient lies in \(\operatorname{im}E\).  Indeed, for
\(\xi=X_c(\x)/\kappa\), the standard layer derivative is
\[
 \nabla_{W_\ell}F(\xi;\vartheta)
 =(W_L\cdots W_{\ell+1})^\top
   (W_{\ell-1}\cdots W_1\xi)^\top,
\]
with empty matrix products interpreted as identities.  At \(\vartheta = E\thetab\),
every hidden forward or backward vector is a scalar multiple of the
corresponding sign vector $\mathbf s_\ell$, and \(\xi\in\operatorname{im}U_c\).  Thus the
gradient blocks have the copy forms in \eqref{eq:lift_first_layer}--\eqref{eq:lift_output_layer},
so each model gradient belongs to \(\operatorname{im}E\).  The same
holds for \(\g^{\mathrm w}\), which is a weighted sum of these gradients.
Writing \(\g^{\mathrm w}=E\h\), multiplying by $E^\top$ and using~\eqref{eq:g_n_w_chain_rule} now yields $E^{\top}E\h=\g^{\mathrm n}$, therefore
\[
\g ^{\mathrm w}=E(E^{\top}E)^{-1}\g^{\mathrm n}=E^\sharp\g^{\mathrm n},
\]
where \(E^{\top}E\) is invertible because \(E\) is injective.

For the explicit block formulas, we compute the adjoint
\(E^\top:\mathcal V\to\Theta\).  Put
\(\mathbf r_c:=\sum_{k\neq c}e_k\), and use the standard Euclidean inner
product on \(\Theta\) and the sum of Frobenius inner products on
\(\mathcal V\).  For any tuple \(\vartheta=(W_1,\ldots,W_L)\), the adjoint is characterized by
\[
 \innerprod{E^\top\vartheta}{\thetab}_{\Theta}
 =\innerprod{\vartheta}{E\thetab}_{\mathcal V}
 \qquad\text{for every }\thetab\in\Theta.
\]
We find its coordinates by expanding the right-hand side and comparing
the coefficients of the coordinates of \(\thetab\).

For a matrix \(W\) and column vectors \(a,b\) of compatible dimensions,
\[
 \innerprod{W}{ab^\top}_F
 =\sum_{i,j}W_{ij}a_i b_j=a^\top Wb.
\]
Write \(\thetab=(\ub,b_1,\ldots,b_{L-1})\).
Since \(U_c(\ub)=u_1\e_c+u_2\mathbf r_c\), the first
block of the lift is
\[
 (E\thetab)_1
 =u_1\mathbf s_1\e_c^\top
  +u_2\mathbf s_1\mathbf r_c^\top.
\]
Its contribution to the inner product is therefore
\begin{align*}
 \innerprod{W_1}{(E\thetab)_1}_F
 &=u_1\innerprod{W_1}{\mathbf s_1\e_c^\top}_F
   +u_2\innerprod{W_1}{\mathbf s_1\mathbf r_c^\top}_F\\
 &=u_1\mathbf s_1^\top W_1\e_c
   +u_2\mathbf s_1^\top W_1\mathbf r_c.
\end{align*}
Applying the same matrix identity to each middle block and to the output
block gives the complete expansion
\begin{align*}
 \innerprod{\vartheta}{E\thetab}_{\mathcal V}
 &=u_1\mathbf s_1^\top W_1e_c
   +u_2\mathbf s_1^\top W_1\mathbf r_c\\
 &\quad+\sum_{r=1}^{L-2}
   b_r\mathbf s_{r+1}^\top W_{r+1}\mathbf s_r
   +b_{L-1}W_L\mathbf s_{L-1}.
\end{align*}
Because the defining adjoint identity holds for every \(\thetab\),
the coefficients of \(u_1,u_2,b_1,\ldots,b_{L-1}\)
are exactly the coordinates of \(E^\top\vartheta\).
Thus
\begin{equation}
 E^\top\vartheta
 =\left(
 \mathbf s_1^\top W_1e_c,\,
 \mathbf s_1^\top W_1\mathbf r_c,\,
 \bigl(\mathbf s_{r+1}^\top W_{r+1}\mathbf s_r\bigr)_{r=1}^{L-2},\,
 W_L\mathbf s_{L-1}
 \right).
 \label{eq:lift_adjoint_coordinates}
\end{equation}

Since \(\|\mathbf s_\ell\|^2=n_\ell\) and
\(\|\mathbf r_c\|^2=d-1\), applying this to \(E\thetab\) gives
\begin{equation}
 E^\top E(\ub,b_1,\ldots,b_{L-1})
 =\left(n_1u_1,n_1(d-1)u_2,
        (n_{r+1}n_rb_r)_{r=1}^{L-2},
        n_{L-1}b_{L-1}\right).
 \label{eq:lift_gram_map}
\end{equation}
Substituting this diagonal map into \(E(E^\top E)^{-1}\g^{\mathrm n}\) gives
\eqref{eq:combined_first_gradient_copy}--\eqref{eq:combined_last_gradient_copy}.
The indexed middle tuple is empty when \(L=2\).
\end{proof}

\begin{corollary}[The Adam trajectory is exactly lifted]
\label{prop:adam-exact-lift}
Fix \(\mathbf s\), \(c\in[d]\), a differentiable margin loss \(\ell\),
and a finite dataset
\[
 S^{\mathrm n}=\{(\x_i,y_i)\}_{i=1}^{m}
 \subseteq\bbR^2\times\{\pm1\}.
\]
Use the corresponding wide dataset
\[
 S^{c,\mathrm w}
 :=\left\{\left(X_c(\x_i)/\kappa,y_i\right)\right\}_{i=1}^{m}.
\]

If \(\thetab_t\) is the Adam trajectory on $S^{\mathrm n}$ from \(\thetab_0\),
then \(E_{\mathbf s,c}\thetab_t\) is the Adam trajectory on $S^{c,\mathrm w}$ from
\(E_{\mathbf s,c}\thetab_0\).
\end{corollary}

\begin{proof}
Let
\[
 \g^{\mathrm n}(t):=
 \nabla_{\thetab}\mathcal L^{\mathrm n}_S(\thetab_t),
 \qquad
 \g^{\mathrm w}(t):=
 \nabla_\vartheta\mathcal L^{\mathrm w}_{S^{c,\mathrm w}}
   (E_{\mathbf s,c}\thetab_t).
\]
The definition of $E_{\mathbf s,c}$ and Equations~\eqref{eq:combined_first_gradient_copy}--\eqref{eq:combined_last_gradient_copy} (together with the $L=1$ formula in
Lemma~\ref{lem:common-model-gradient-identities}) give for each wide
coordinate \(e\) a narrow coordinate \(r(e)\) and a fixed nonzero
constant \(a_e\) such that
\begin{equation}
 g_e^{\mathrm w}(t)=a_e g_{r(e)}^{\mathrm n}(t),
 \label{eq:adam_gradient_scalar_copy}
\end{equation}
and
\begin{equation}\label{eq:adam_parameter_scalar_copy}
        (E_{\mathbf s,c}\thetab)_e
    =\sign{a_e}\thetab_{r(e)} \ .
\end{equation}

Let \(m_r^{\mathrm n},v_r^{\mathrm n}\) be the narrow moments. Define wide
moments along the candidate trajectory by
\begin{equation}
 m_e^{\mathrm w}(t):=a_e m_{r(e)}^{\mathrm n}(t),
 \qquad
 v_e^{\mathrm w}(t):=a_e^2v_{r(e)}^{\mathrm n}(t).
 \label{eq:adam_moment_lift}
\end{equation}
By linearity of the EMA operators, these are the moments associated
with \(\g^{\mathrm w}\). The bias corrections give
\begin{equation}
 \frac{\widehat m_e^{\mathrm w}}{\sqrt{\widehat v_e^{\mathrm w}}}
 =\sign{a_e}\,
  \frac{\widehat m_{r(e)}^{\mathrm n}}
       {\sqrt{\widehat v_{r(e)}^{\mathrm n}}}.
 \label{eq:adam_ratio_scalar_copy}
\end{equation}
This identity also holds at zero second-moment coordinates under
the zero-denominator convention.

Hence, combining~\eqref{eq:adam_parameter_scalar_copy},~\eqref{eq:adam_ratio_scalar_copy},
\[
 \frac{d}{dt}(E_{\mathbf s,c}\thetab_t)
 =E_{\mathbf s,c}\dot\thetab_t
 =E_{\mathbf s,c}\left(
   -\eta(t)\frac{\widehat{\m}^{\mathrm n}_t}
                    {\sqrt{\widehat{\vb}^{\mathrm n}_t}}
   \right)
 =-\eta(t)\frac{\widehat{\m}^{\mathrm w}_t}
                   {\sqrt{\widehat{\vb}^{\mathrm w}_t}}.
\]
So $E_{\mathbf s,c}\thetab_t$ solves the Adam equations.
\end{proof}

\begin{proposition}[Wide KKT points pull back]
\label{prop:full-wide-kkt-pulls-back}
Fix \(\mathbf s\) and \(c\in[d]\), and abbreviate
\(E:=E_{\mathbf s,c}\).  For a function
\(R:\mathcal V\to\bbR\), set \(R_{\mathbf s,c}:=R\circ E\).
For a dataset \(S=\{(\x_i,y_i)\}_{i=1}^{m}\), write
\[
 q_i^{\mathrm n}(\thetab):=y_if(\x_i;\thetab),
 \qquad
 q_i^{\mathrm w}(\vartheta)
 :=y_iF\!\left(X_c(\x_i)/\kappa;\vartheta\right).
\]
Consider the narrow problem
\begin{equation}
 \min_{\thetab\in\Theta} R_{\mathbf s,c}(\thetab)
 \qquad\text{subject to}\qquad
 q_i^{\mathrm n}(\thetab)\geq1,
 \quad 1\leq i\leq m.
\label{eq:kkt_narrow_problem}
\end{equation}
If $R$ is convex and
\(\widehat\vartheta=E\widehat\thetab\) is a KKT point of
\[
 \min_{\vartheta\in\mathcal V}R(\vartheta)
 \qquad\text{subject to}\qquad
 q_i^{\mathrm w}(\vartheta)\geq1,
 \quad 1\leq i\leq m,
\]
then \(\widehat\thetab\) is a KKT point of
\eqref{eq:kkt_narrow_problem}.
\end{proposition}

\begin{proof}
The model and gradient identities give
\begin{equation}
 q_i^{\mathrm w}(E\thetab)=q_i^{\mathrm n}(\thetab),
 \qquad
 E^\top\nabla q_i^{\mathrm w}(E\thetab)
 =\nabla q_i^{\mathrm n}(\thetab).
\label{eq:kkt_copy_margin_gradient}
\end{equation}
By the wide KKT conditions, there are
\(\zeta\in\partial R(E\widehat\thetab)\) and
\(\lambda_i\geq0\) such that
\begin{equation}\label{eq:wide_KKT_conditions}
 q_i^{\mathrm w}(E\widehat\thetab)\geq1,
 \qquad
 \zeta-\sum_{i=1}^{m}\lambda_i
   \nabla q_i^{\mathrm w}(E\widehat\thetab)=0,
 \qquad
 \lambda_i\bigl(q_i^{\mathrm w}(E\widehat\thetab)-1\bigr)=0.   
\end{equation}
For every \(\thetab'\in\Theta\), the subgradient inequality for a convex $R$ gives
\begin{align*}
 R_{\mathbf s,c}(\thetab')
 &=R(E\thetab')\\
 &\geq R(E\widehat\thetab)
   +\innerprod{\zeta}{E(\thetab'-\widehat\thetab)}\\
 &=R_{\mathbf s,c}(\widehat\thetab)
   +\innerprod{E^\top\zeta}{\thetab'-\widehat\thetab}.
\end{align*}
Thus
\(E^\top\zeta\in\partial R_{\mathbf s,c}(\widehat\thetab)\).
Applying \(E^\top\) to the middle equation in ~\eqref{eq:wide_KKT_conditions} and using
\eqref{eq:kkt_copy_margin_gradient} gives
\[
 E^\top\zeta-\sum_{i=1}^{m}\lambda_i
   \nabla q_i^{\mathrm n}(\widehat\thetab)=0.
\]
The margin identity in \eqref{eq:kkt_copy_margin_gradient} transfers
feasibility and complementarity.  Hence \(\widehat\thetab\) satisfies the
KKT conditions for \eqref{eq:kkt_narrow_problem}.
\end{proof}

\begin{lemma}[Convex regularizer triviality for the wide model]
\label{lemma:abstract-convex-transfer}
Let $\mathsf O_{\mathcal V},\mathsf O_{\Theta}$ be optimizers on the wide and narrow parameter spaces respectively. Let $\ell : \bbR \to \bbR$.
Assume the following two properties.
\begin{enumerate}
\item Trajectory lifting: for every trajectory \(\thetab_t\) of
      \(\mathsf O_{\Theta}\) on $f$ and a dataset $S^{\mathrm n}$, and every sign pattern \(\mathbf s\) and \(c\in[d]\), \(E_{\mathbf s,c}\thetab_t\) is a trajectory of
      \(\mathsf O_{\mathcal V}\) on $F$ and the dataset $S^{c,\mathrm w}$.
\item Triviality of the narrow regularizer: every convex implicit         regularizer of $f$ for the
      narrow optimizer \(\mathsf O_{\Theta}\) with the loss $\ell$ is constant on
      \(\Theta\).
\end{enumerate}
Then every convex implicit regularizer of $F$ for
\(\mathsf O_{\mathcal V}\) with the loss $\ell$ is constant on \(\mathcal V\).
\end{lemma}

\begin{proof}
Let \(R:\mathcal V\to\bbR\) be such a wide regularizer.  Fix
\(\mathbf s,c\), abbreviate \(E:=E_{\mathbf s,c}\), and set
\(R_{\mathbf s,c}:=R\circ E\). Since $R$ is convex, so is $R_{\mathbf s,c}$; we
show that $R_{\mathbf s,c}$ is an implicit regularizer for $\mathsf O_{\Theta}$.

Let \(\thetab_t\) be a trajectory of $\mathsf O_{\Theta}$ satisfying Assumptions~\ref{traj_ass:positive_margin_interpolation},~\ref{traj_ass:loss_decay}, and suppose 
\[
 \frac{\thetab_t}
      {\norm{\thetab_t}_2}
 \longrightarrow\bar\thetab\neq0,
 \qquad
 \qmin^{\mathrm n}(\bar\thetab)>0.
\]
By the first assumption, \(\vartheta_t=E\thetab_t\) is a legal wide
trajectory on $S^{c,\mathrm w}$.  Injectivity gives \(E\bar\thetab\neq0\), and linearity of $E$ gives
\begin{align*}
 \frac{\vartheta_t}{\norm{\vartheta_t}_{2}}
 &=\frac{E\bigl(\thetab_t/
       \norm{\thetab_t}_{2}\bigr)}
 {\norm{E\bigl(\thetab_t/
       \norm{\thetab_t}_{2}\bigr)}
       _{2}} \toinfty{t}\frac{E\bar\thetab}{\norm{E\bar\thetab}_{2}} := \bar \vartheta\\
\end{align*}
Homogeneity and the model equivalence therefore give
\[
 \qmin^{\mathrm w}(\bar\vartheta)
 =\frac{\qmin^{\mathrm n}(\bar \thetab)}{\norm{E\bar\thetab}_{2}^L}>0.
\]
Consequently, the wide margin-normalized limit is
\begin{align*}
 \widehat\vartheta
 &:=\frac{\bar\vartheta}
 {\bigl(\min_iq_i^{\mathrm w}(\bar\vartheta)\bigr)^{1/L}}=E\widehat\thetab \ .
\end{align*}
Assumptions~\ref{traj_ass:positive_margin_interpolation},~\ref{traj_ass:loss_decay} hold for $\vartheta_t$: $\Lcal^{\mathrm w}(\vartheta_t) = \Lcal^{\mathrm n}(\thetab_t) \toinfty{t} 0$ and $\gamma^{\mathrm w}(\vartheta_t) = \gamma^{\mathrm n}(\thetab_t) \cdot \left(\frac{\norm{\thetab_t}_2}{\norm{\vartheta_t}_2} \right)^L \geq \frac{\gamma_0}{\norm{E}_{2 \to 2}^L} > 0$. Therefore, 
By definition of $R$, \(E\widehat\thetab\) is a KKT point for the
wide problem.  Proposition~\ref{prop:full-wide-kkt-pulls-back}
then makes \(\widehat\thetab\) KKT for the narrow problem with objective
\(R_{\mathbf s,c}\).  Hence \(R_{\mathbf s,c}\) is a regularizer for the narrow model.

The second assumption makes \(R_{\mathbf s,c}\) constant on \(\Theta\).
Its value is \(R_{\mathbf s,c}(0)=R(0)\), and \(E\) maps \(\Theta\) onto
\(V_{\mathbf s,c}\).  Therefore
\[
 R|_{V_{\mathbf s,c}}\equiv R(0)
 \qquad\text{for every \(\mathbf s\) and \(c\)}.
\]
Lemma~\ref{lem:common-convex-propagation} now makes \(R\) constant on all
of \(\mathcal V\).
\end{proof}

\begin{corollary}[No nonconstant convex wide Adam regularizer]
\label{cor:adam-wide-constancy}
Let $\alpha > \frac 1L - \frac12$. For any scalar-output linear network $F$ of depth $L$, input dimension \(d\geq2\), any choice of positive hidden widths and a total of $p$ parameters, every
convex implicit regularizer of Adam on $F$ with the exponential loss is constant on $\bbR^p$.
\end{corollary}

\begin{proof}
This is precisely the statement of Lemma~\ref{lemma:abstract-convex-transfer}, following from Corollary~\ref{prop:adam-exact-lift} and the narrow (width-1) construction.
\end{proof}

\subsection{General Homogeneous Models}\label{sec:general_homogeneous_models}

\subsubsection{Preconditioned Momentum GD}
\begin{proposition}\label{prop:preconditioned_momentum_gd}
    Let $\alpha > \frac 1L - \frac12$, $f$ satisfying~\ref{model_ass:1_locally_lipschitz},~\ref{model_ass:2_homogeneous} and $\ell \in \mathrm L_{\e}$. Under Assumptions~\ref{traj_ass:positive_margin_interpolation},~\ref{traj_ass:loss_decay}, for every $j \in [p]$,
    \begin{equation}\label{eq:second_moment_freezing_rate}
       \omega_j:=c_2\int_0^\infty e^{c_2s}\g_s[j]^2ds < \infty, \qquad \omega_j - c_2\int_0^te^{c_2s}\g_s[j]^2\,ds \leq \bigO{t^{-2\left(\alpha - \left(\frac1L - \frac12\right)\right)}} 
    \end{equation}
    Therefore Adam asymptotically follows a preconditioned momentum gradient descent trajectory. Namely, it holds that
    \[
    \dtheta{t} = -D_\omega \eta(t)e^{\frac{c_2}2t}\m_t\left(1+\smallo{1}\right), \qquad (D_\omega)_{j,j} = \begin{cases}
        \omega_j^{-1/2}  &:\omega_j > 0, \\
        1  &:\omega_j = 0
    \end{cases} \ .
    \]
\end{proposition}
\begin{proof}
Equation~\eqref{eq:second_moment_freezing_rate} is exactly the result of Corollary~\ref{cor:second_moment_freezing}. Set
\[
    I_j(t):= c_2\int_0^t e^{c_2s}\g_s[j]^2\,ds.
\]
For every $j$ with $\omega_j>0$, it holds that
$I_j(t)>0$ for all sufficiently large $t$.  Since
$v_j(t)=e^{-c_2t}I_j(t)$ and $\widehat b(t) = \frac{\sqrt{1-e^{-c_2t}}}{1-e^{-c_1t}}\to1$, define
\[
    (E_t)_{j,j}
    := \begin{cases}
    \widehat b(t)
       I_j(t)^{-1/2}-(D_\omega)_{j,j}, \quad & :\omega_j > 0, \\
    0 \quad & :\omega_j = 0 \ .
    \end{cases}
\]
Then $\norm{E_t}_{\mathrm{op}}\to0$. Note that if $\omega_j=0$ then $\m_t[j]=\vb_t[j]=0$ for all $t$, so the Adam equation is exactly
\[
    \dot\thetab_t
    =
    -\eta(t)e^{c_2t/2}(D_\omega+E_t)\m_t.
\]
Equivalently, for each coordinate, either $(D_\omega + E_t)_{j,j} = (D_\omega)_{j,j}$ when $\omega _j = 0$ or
\[
    (D_\omega+E_t)_{jj}
    =
    (D_\omega)_{jj}
    \left[
      1+\widehat b(t)\sqrt{\frac{\omega_j}{I_j(t)}}-1
    \right],
\]
when $\omega_j > 0$, where the expression in square brackets is $1+\smallo{1}$.
\end{proof}

\subsubsection{Trajectory-dependent KKT}\label{sec:trajectory_dependent_kkt}
We show that every directional limit point satisfies the
trajectory-dependent KKT conditions.

\begin{theorem}[Every directional limit point is KKT]
\label{thm:adam_fast_trajectory_dependent_kkt_appendix}
Let $f$ satisfy~\ref{model_ass:1_locally_lipschitz},~\ref{model_ass:2_homogeneous} and let $\alpha>1/L-1/2, \ell \in \mathrm L_{\e}$. Let $\thetab_t$ be an Adam trajectory under assumptions~\ref{traj_ass:positive_margin_interpolation},~\ref{traj_ass:loss_decay} and let
\(\bar\thetab\) be any limit point of
\(\thetab_t/\norm{\thetab_t}_2\) as \(t\to\infty\).
Then \(\qmin(\bar\thetab)>0\), and
\(\hat\thetab:=\bar\thetab/\qmin(\bar\thetab)^{1/L}\)
is a Clarke--KKT point of
\begin{equation}\label{eq:preconditioned_kkt_appendix}
    \min_{\thetab}
    \frac12\thetab^\top D_\omega^{-1}\thetab
    \qquad\text{s.t.}\qquad
    q_i(\thetab)=y_i f(\x_i;\thetab)\geq1
    \quad(i\in[m]).
\end{equation}
\end{theorem}

For brevity, write \(D=D_\omega\). The second-moment freezing
result makes \(D\) a fixed positive definite diagonal matrix, including
when some \(\omega_j=0\). We use the weighted inner product and norm
\[
    \innerprod{\mathbf u}{\mathbf v}_{D^{-1}}
    :=\mathbf u^\top D^{-1}\mathbf v,
    \qquad
    \norm{\mathbf u}_{D^{-1}}^2
    :=\innerprod{\mathbf u}{\mathbf u}_{D^{-1}}.
\]

For the remainder, work under Assumptions~\ref{traj_ass:positive_margin_interpolation},~\ref{traj_ass:loss_decay} and
\(\alpha>1/L-1/2\), and set
\[
    \nu:=1-\frac1L,\qquad
    \zeta:=\alpha+\nu>\frac12,\qquad
    \lambda:=c_1-\frac{c_2}{2}>0.
\]
Let
\[
    I_j(t):=c_2\int_0^t e^{c_2s}\g_s[j]^2\,ds,\qquad
    a(t):=\eta_0 t^\alpha e^{\frac{c_2}{2}t},
    \qquad
    (B_t)_{j,j}:=
    \begin{cases}
        \sqrt{\omega_j/I_j(t)},&\omega_j>0,\\
        1,&\omega_j=0.
    \end{cases}
\]
These definitions are valid on a sufficiently late tail. Set
\[
    A_t:=\frac{\hat\eta(t)}{\eta_0 t^\alpha}B_t.
\]
The exact frozen dynamics and the estimates established in the
second-moment freezing give for some $C > 0$
\begin{align}
    \dot\thetab_t&=-a(t)A_tD\m_t,
    &\dot\m_t&=c_1(\g_t-\m_t),
    \label{eq:adam_frozen_preconditioner}\\
    \norm{\g_t}_2&\leq Ce^{-\frac{c_2}{2}t}t^{-\zeta},
    &\norm{B_t-I}_{\mathrm{op}}
        &\leq Ct^{-(2\zeta-1)}.
    \label{eq:all_limit_gradient_and_preconditioner_bounds}
\end{align}
If \(\omega_j=0\), then
\(\g_t[j]=0\) almost everywhere, \(\m_t[j]=\vb_t[j]=0\),
and \(\thetab_t[j]\) is constant; hence the same equations apply. We also use the selected decomposition and chain rule
\begin{align}
    -\g_t&=\sum_{i=1}^m \varphi'(q_i(\thetab_t))\ell(q_i(\thetab_t))\h_i(t),
    &\h_i(t)&\in\partial q_i(\thetab_t),
    \label{eq:all_limit_gradient_decomposition}\\
    \frac{d}{dt}\Lcal(\thetab_t)
        &=\innerprod{\g_t}{\dot\thetab_t}.
    \label{eq:all_limit_selected_chain_rule}
\end{align}

Define the discrepancy from purely radial growth and the error in the
frozen dynamics by
\begin{equation}\label{eq:all_limit_discrepancy_and_error}
    \mathcal E_t:=-a(t)D\m_t-\frac{\thetab_t}{Lt},
    \qquad
    \mathbf e_t:=-a(t)(A_t-I)D\m_t.
\end{equation}
Thus
\begin{equation}\label{eq:all_limit_theta_equation}
    \dot\thetab_t=-a(t)D\m_t+\mathbf e_t
       =\frac{\thetab_t}{Lt}+\mathcal E_t+\mathbf e_t.
\end{equation}

\begin{lemma}[Bounds for the dynamics]
\label{lemma:adam_fast_rescaled_bounds}
There exist \(T,c,C>0\) such that, for all \(t\geq T\),
\begin{equation}\label{eq:all_limit_scale_bounds}
    ct^{1/L}\leq\norm{\thetab_t}_2\leq Ct^{1/L},\qquad
    \qmin(\thetab_t)\geq ct,\qquad
    \norm{\m_t}_2\leq Ce^{-\frac{c_2}{2}t}t^{-\zeta},
\end{equation}
\[
    \Lcal(\thetab_t)\leq Ce^{-\frac{c_2}{2}t}t^{-(\zeta+\nu)},
    \qquad
    \norm{\mathcal E_t}_2\leq Ct^{-\nu},
\]
and
\begin{equation}\label{eq:all_limit_error_rate}
    \norm{\mathbf e_t}_2\leq Ct^{-\nu-\delta},\qquad
    \delta:=\min\{1,2\zeta-1\}>0.
\end{equation}
Almost everywhere on this tail,
\begin{equation}\label{eq:all_limit_gradient_and_speed_bounds}
    \norm{\g_t}_2\leq Ce^{-\frac{c_2}{2}t}t^{-\zeta},\qquad
    \norm{\dot\thetab_t}_2\leq Ct^{-\nu}.
\end{equation}
\end{lemma}
\begin{proof}
Since $\zeta>1/2$, Theorem~\ref{thm:loss_decay_rates_appendix}
gives for some $C>0$, with $U(t):=-\log\Lcal(\thetab_t)$,
\begin{equation}\label{eq:all_limit_loss_asymptotics}
\begin{aligned}
    U(t)
    &=\frac{c_2}{2}t+(\zeta+\nu)\log t+\bigO{1},\\
    \Lcal(\thetab_t)
    &\leq Ce^{-\frac{c_2}{2}t}t^{-(\zeta+\nu)}.
\end{aligned}
\end{equation}
Combining this with Lemma~\ref{lemma:freezing-tail-scales} yields for some $C > 0$
\[
    \norm{\thetab_t}_2\asymp t^{1/L},
    \qquad
    \qmin(\thetab_t)\asymp t,
    \qquad
    \norm{\g_t}_2\leq Ce^{-\frac{c_2}{2}t}t^{-\zeta},
\]
where the gradient estimate holds almost everywhere.

Corollary~\ref{cor:tracking_and_speed_estimates} and
$\vb_t[j]=e^{-c_2t}I_j(t)\leq\omega_j e^{-c_2t}$ imply for some $C > 0$
\[
    \norm{\m_t}_2
    \leq Ct^{-\zeta}
         \left(\sum_j\vb_t[j]\right)^{1/2}
    \leq Ce^{-\frac{c_2}{2}t}t^{-\zeta}.
\]
Coordinates with $\omega_j=0$ contribute zero.
Boundedness of $A_t,D$ and~\eqref{eq:adam_frozen_preconditioner}
therefore give for some $C > 0$, almost everywhere,
\[
    \norm{\dot\thetab_t}_2
    \leq Ca(t)\norm{\m_t}_2
    \leq Ct^{\alpha-\zeta}=Ct^{-\nu}.
\]

Finally, since $\hat\eta(t)/(\eta_0 t^\alpha)=1+\bigO{t^{-1}}$,
the freezing estimate in~\eqref{eq:all_limit_gradient_and_preconditioner_bounds}
gives $\norm{A_t-I}_{\mathrm{op}}\leq Ct^{-\delta}$.
The definitions in~\eqref{eq:all_limit_discrepancy_and_error} now yield for some $C > 0$
\begin{align*}
    \norm{\mathbf e_t}_2
    &\leq a(t)\norm{A_t-I}_{\mathrm{op}}\norm{D\m_t}_2
      \leq Ct^{-\nu-\delta},\\
    \norm{\mathcal E_t}_2
    &\leq a(t)\norm{D\m_t}_2
          +\frac{\norm{\thetab_t}_2}{Lt}
      \leq Ct^{-\nu}.
\end{align*}
All bounds hold on a common sufficiently late tail.
\end{proof}

Although pointwise we only have the bound $\norm{\mathcal E_t}_2 \lesssim t^{-\nu}$, in fact we can show summability of $t^{2\nu-1}\norm{\mathcal E_t}^2$.
\begin{lemma}[Finite energy of the radial discrepancy]
\label{lemma:adam_fast_finite_energy}
For sufficiently large \(T>0\),
\begin{equation}\label{eq:all_limit_finite_energy}
    \int_T^\infty
       t^{2\nu}\norm{\mathcal E_t}_{D^{-1}}^2\,\frac{dt}{t}
       <\infty.
\end{equation}
\end{lemma}

\begin{proof}
Fix a sufficiently large \(T\) as in
Lemma~\ref{lemma:adam_fast_rescaled_bounds},
and let \(T'\geq T\). All constants below are independent of \(T'\).
All differential identities and inequalities hold almost everywhere.
Define the nonnegative, locally absolutely continuous function
\[
    \mathcal C(t)
    :=\frac{\norm{\thetab_t}_{D^{-1}}^2}{2t^{2/L}}
      +\frac{L}{2\lambda}t^{2\nu-1}
          \norm{\mathcal E_t}_{D^{-1}}^2
      +\frac{c_1L}{\lambda}t^{2\nu-1}
          a(t)\Lcal(\thetab_t).
\]

We show that for some $C > 0$,
\begin{equation}\label{eq:C_derivative_bound}
    \dot {\mathcal C}(t) \leq -Lt^{2\nu -1}\norm{\mathcal E_t}^2_{D^{-1}} + Ct^{-1-\delta},
\end{equation}
which implies the lemma's claim: integrating and using \(\mathcal C(T')\geq0\), we conclude that
\[
    L\int_T^{T'}t^{2\nu}
       \norm{\mathcal E_t}_{D^{-1}}^2\,\frac{dt}{t}
       \leq\mathcal C(T)-\mathcal C(T')
          +C\int_T^{T'}t^{-1-\delta}\,dt
       \leq\mathcal C(T)+\frac{C}{\delta}T^{-\delta}.
\]
Letting \(T'\to\infty\)
proves~\eqref{eq:all_limit_finite_energy}.

We proceed to prove~\eqref{eq:C_derivative_bound}. First, Equation~\eqref{eq:all_limit_theta_equation} and the identity
\(\thetab_t=Lt(-a(t)D\m_t-\mathcal E_t)\) give
\begin{align*}
    \frac{d}{dt}\frac{\norm{\thetab_t}_{D^{-1}}^2}{2t^{2/L}}
       &=t^{-2/L}\innerprod{\thetab_t}{\dot\thetab_t}_{D^{-1}}
          -\frac1L t^{-2/L-1}\norm{\thetab_t}_{D^{-1}}^2\\
       &=t^{-2/L}
          \innerprod{\thetab_t}
                    {\dot\thetab_t-\thetab_t/(Lt)}_{D^{-1}}\\
       &=t^{-2/L}\innerprod{\thetab_t}{\mathcal E_t}_{D^{-1}}
          +t^{-2/L}\innerprod{\thetab_t}{\mathbf e_t}_{D^{-1}}\\
       &=Lt^{1-2/L}
          \innerprod{-a(t)D\m_t-\mathcal E_t}
                    {\mathcal E_t}_{D^{-1}}
          +t^{-2/L}\innerprod{\thetab_t}{\mathbf e_t}_{D^{-1}}\\
       &=Lt^{2\nu-1}
          \innerprod{-a(t)D\m_t}{\mathcal E_t}_{D^{-1}}
          -Lt^{2\nu-1}\norm{\mathcal E_t}_{D^{-1}}^2
          +t^{-2/L}\innerprod{\thetab_t}{\mathbf e_t}_{D^{-1}},
\end{align*}
where \(1-2/L=2\nu-1\).

For the second term, the momentum equation
in~\eqref{eq:adam_frozen_preconditioner} and
\(a'(t)/a(t)=c_2/2+\alpha/t\) give
\begin{equation}\label{eq:all_limit_scaled_momentum_equation}
    \frac{d}{dt}(-a(t)D\m_t)
       =-c_1a(t)D\g_t-\lambda(-a(t)D\m_t)
          +\frac{\alpha}{t}(-a(t)D\m_t).
\end{equation}
Differentiating~\eqref{eq:all_limit_discrepancy_and_error} and using
\eqref{eq:all_limit_scaled_momentum_equation} gives
\begin{align*}
    \dot{\mathcal E}_t
       &=\frac{d}{dt}(-a(t)D\m_t)
          -\frac{\dot\thetab_t}{Lt}
          +\frac{\thetab_t}{Lt^2}\\
       &=-c_1a(t)D\g_t-\lambda(-a(t)D\m_t)
          +\frac{\alpha}{t}(-a(t)D\m_t)
          -\frac1{Lt}
             \left(\dot\thetab_t-\frac{\thetab_t}{t}\right).
\end{align*}
Thus, by the product rule,
\begin{equation}\label{eq:weighted_E_norm_derivative}
\begin{aligned}
    \frac{d}{dt}
       \left(\frac{L}{2\lambda}t^{2\nu-1}
          \norm{\mathcal E_t}_{D^{-1}}^2\right)
       &=\frac{L}{\lambda}t^{2\nu-1}
          \innerprod{\dot{\mathcal E}_t}{\mathcal E_t}_{D^{-1}}
          +\frac{L(2\nu-1)}{2\lambda}t^{2\nu-2}
             \norm{\mathcal E_t}_{D^{-1}}^2\\
       &=\frac{c_1L}{\lambda}t^{2\nu-1}
          \innerprod{-a(t)D\g_t}{\mathcal E_t}_{D^{-1}}\\
       &\quad-Lt^{2\nu-1}
          \innerprod{-a(t)D\m_t}{\mathcal E_t}_{D^{-1}}\\
       &\quad+\frac{\alpha L}{\lambda}t^{2\nu-2}
          \innerprod{-a(t)D\m_t}{\mathcal E_t}_{D^{-1}}\\
       &\quad-\frac1\lambda t^{2\nu-2}
          \innerprod{\dot\thetab_t-\thetab_t/t}
                    {\mathcal E_t}_{D^{-1}}\\
       &\quad+\frac{L(2\nu-1)}{2\lambda}t^{2\nu-2}
          \norm{\mathcal E_t}_{D^{-1}}^2.
\end{aligned}
\end{equation}

For the third term, the product rule,
\(a'(t)/a(t)=c_2/2+\alpha/t\), and
the chain rule~\eqref{eq:all_limit_selected_chain_rule} give
\begin{equation}\label{eq:weighted_loss_derivative}
\begin{aligned}
    \frac{d}{dt}
       \left(\frac{c_1L}{\lambda}t^{2\nu-1}
          a(t)\Lcal(\thetab_t)\right)
       &=\frac{c_1L(2\nu-1)}{\lambda}t^{2\nu-2}
          a(t)\Lcal(\thetab_t)\\
       &\quad+\frac{c_1L}{\lambda}t^{2\nu-1}
          a'(t)\Lcal(\thetab_t)\\
       &\quad+\frac{c_1L}{\lambda}t^{2\nu-1}
          a(t)\frac{d}{dt}\Lcal(\thetab_t)\\
       &=\frac{c_1L}{\lambda}t^{2\nu-1}a(t)
          \innerprod{\g_t}{\dot\thetab_t}\\
       &\quad+\frac{c_1Lc_2}{2\lambda}t^{2\nu-1}
          a(t)\Lcal(\thetab_t)\\
       &\quad+\frac{c_1L(\alpha+2\nu-1)}{\lambda}t^{2\nu-2}
          a(t)\Lcal(\thetab_t).
\end{aligned}
\end{equation}

Adding the three derivatives gives
\[
\begin{aligned}
    \dot{\mathcal C}(t)
    ={}&-Lt^{2\nu-1}\norm{\mathcal E_t}_{D^{-1}}^2\\
    &+\underbrace{
       Lt^{2\nu-1}\left[
          \innerprod{-a(t)D\m_t}{\mathcal E_t}_{D^{-1}}
          -\innerprod{-a(t)D\m_t}{\mathcal E_t}_{D^{-1}}
       \right]}_{(*)}\\
    &+\underbrace{
       \frac{c_1L}{\lambda}t^{2\nu-1}
       \left[
       \begin{aligned}
          &\innerprod{-a(t)D\g_t}{\mathcal E_t}_{D^{-1}}
             +a(t)\innerprod{\g_t}{\dot\thetab_t}\\
          &\qquad+\frac{c_2}{2}a(t)\Lcal(\thetab_t)
       \end{aligned}
       \right]}_{(**)}\\
    &+\underbrace{
       \begin{aligned}[t]
          &t^{-2/L}\innerprod{\thetab_t}{\mathbf e_t}_{D^{-1}}+t^{2\nu-2}\left[
          \begin{aligned}
             &\frac{\alpha L}{\lambda}
                \innerprod{-a(t)D\m_t}{\mathcal E_t}_{D^{-1}}
                -\frac1\lambda
                \innerprod{\dot\thetab_t-\thetab_t/t}
                          {\mathcal E_t}_{D^{-1}}\\
             &+\frac{L(2\nu-1)}{2\lambda}
                \norm{\mathcal E_t}_{D^{-1}}^2
                +\frac{c_1L(\alpha+2\nu-1)}{\lambda}
                   a(t)\Lcal(\thetab_t)
          \end{aligned}
          \right]
       \end{aligned}}_{(***)}.
\end{aligned}
\]

The two terms in \((*)\), coming from the first and second
derivatives, cancel exactly.

For \((**)\), Equation~\eqref{eq:all_limit_theta_equation} gives
\begin{align*}
    (**)
       &=\frac{c_1L}{\lambda}t^{2\nu-1}a(t)
          \left[
             -\innerprod{\g_t}{\mathcal E_t}
             +\innerprod{\g_t}{\dot\thetab_t}
             +\frac{c_2}{2}\Lcal(\thetab_t)
          \right]\\
       &=\frac{c_1L}{\lambda}t^{2\nu-1}a(t)
          \left[
             \frac1{Lt}\innerprod{\g_t}{\thetab_t}
             +\frac{c_2}{2}\Lcal(\thetab_t)
             +\innerprod{\g_t}{\mathbf e_t}
          \right].
\end{align*}
Equations~\eqref{eq:radial_gradient_loss_scale}
and~\eqref{eq:all_limit_loss_asymptotics} imply, after enlarging \(T\)
if necessary,
\begin{equation}\label{eq:all_limit_radial_force}
    -\frac1L\innerprod{\g_t}{\thetab_t}
    \geq\bigl(U(t)-\varphi(0)\bigr)\Lcal(\thetab_t)
    \geq\frac{c_2}{2}t\Lcal(\thetab_t)
    \qquad(t\geq T).
\end{equation}
Thus the first two terms in the bracket have nonpositive sum.
Equations~\eqref{eq:all_limit_gradient_and_speed_bounds} and
\eqref{eq:all_limit_error_rate}, together with
\(\alpha-\zeta=-\nu\), therefore give, for some \(C>0\),
\begin{align*}
    (**)
       &\leq\frac{c_1L}{\lambda}t^{2\nu-1}a(t)
          \innerprod{\g_t}{\mathbf e_t}\leq Ct^{2\nu-1}a(t)
          \norm{\g_t}_2\norm{\mathbf e_t}_2\leq Ct^{2\nu-1+\alpha-\zeta-\nu-\delta}
        =Ct^{-1-\delta}.
\end{align*}

For \((***)\),
Equations~\eqref{eq:all_limit_scale_bounds},
\eqref{eq:all_limit_gradient_and_speed_bounds},
\eqref{eq:all_limit_discrepancy_and_error}, and
\eqref{eq:all_limit_loss_asymptotics} imply, for some \(C>0\),
\begin{align*}
    \norm{a(t)D\m_t}_{D^{-1}}
       &\leq Ca(t)\norm{\m_t}_2\leq Ct^\alpha e^{c_2t/2}
                  e^{-c_2t/2}t^{-\zeta}
        =Ct^{-\nu},\\
    \norm{\mathcal E_t}_{D^{-1}}
       &\leq\norm{a(t)D\m_t}_{D^{-1}}
          +\frac{\norm{\thetab_t}_{D^{-1}}}{Lt}\leq Ct^{-\nu}+Ct^{1/L-1}
        \leq Ct^{-\nu},\\
    \norm{\dot\thetab_t-\thetab_t/t}_{D^{-1}}
       &\leq\norm{\dot\thetab_t}_{D^{-1}}
          +\frac{\norm{\thetab_t}_{D^{-1}}}{t}\leq Ct^{-\nu}+Ct^{1/L-1}
        \leq Ct^{-\nu},\\
    a(t)\Lcal(\thetab_t)
       &\leq Ct^{\alpha-\zeta-\nu}
        =Ct^{-2\nu}.
\end{align*}
Here we used \(\alpha-\zeta=-\nu\) and \(1/L-1=-\nu\).
By Cauchy--Schwarz and~\eqref{eq:all_limit_error_rate}, for some \(C>0\),
\begin{align*}
    \left|(***)\right|
       &\leq Ct^{-2/L}\norm{\thetab_t}_2\norm{\mathbf e_t}_2\\
       &\quad+Ct^{2\nu-2}\left[
          \norm{a(t)D\m_t}_{D^{-1}}
             \norm{\mathcal E_t}_{D^{-1}}
          +\norm{\dot\thetab_t-\thetab_t/t}_{D^{-1}}
             \norm{\mathcal E_t}_{D^{-1}}
          +\norm{\mathcal E_t}_{D^{-1}}^2
          +a(t)\Lcal(\thetab_t)
       \right]\\
       &\leq Ct^{-2/L}t^{1/L}t^{-\nu-\delta}
          +Ct^{2\nu-2}t^{-2\nu}\\
       &=C\left(t^{-1-\delta}+t^{-2}\right).
\end{align*}

Combining the estimates for \((*)\), \((**)\), and \((***)\),
and using \(0<\delta\leq1\), we obtain, for some \(C>0\),
\[
    \dot{\mathcal C}(t)
       \leq-Lt^{2\nu-1}\norm{\mathcal E_t}_{D^{-1}}^2
          +C\left(t^{-2}+t^{-1-\delta}\right)
       \leq-Lt^{2\nu-1}\norm{\mathcal E_t}_{D^{-1}}^2
          +Ct^{-1-\delta},
\]
which proves~\eqref{eq:C_derivative_bound}.
\end{proof}

\begin{lemma}[Finite energy implies KKT]
\label{lemma:adam_fast_cluster_point_stationarity}
Assume~\ref{traj_ass:positive_margin_interpolation},~\ref{traj_ass:loss_decay}.
Let \(\bar\thetab\) be any limit point of
\(\thetab_t/\norm{\thetab_t}_2\), and define
\[
    \gamma_*:=\qmin(\bar\thetab)>0,\qquad
    \kappa_*:=\norm{\bar\thetab}_{D^{-1}}^2>0,
\]
\[
    I_*:=\{i\in[m]:q_i(\bar\thetab)=\gamma_*\},\qquad
    K_*:=\operatorname{conv}
          \bigcup_{i\in I_*}\partial q_i(\bar\thetab).
\]
Then
\begin{equation}\label{eq:all_limit_cluster_membership}
    \frac{L\gamma_*}{\kappa_*}D^{-1}\bar\thetab\in K_*.
\end{equation}
\end{lemma}

\begin{proof}
Choose \(t_n\to\infty\) such that
\(\thetab_{t_n}/\norm{\thetab_{t_n}}_2\to\bar\thetab\), and set
\(J_n:=[t_n,2t_n]\). We show the following in succession:
\[
    \frac{\thetab_{2t_n}-\thetab_{t_n}}{\norm{\thetab_{t_n}}_2}
       \longrightarrow(2^{1/L}-1)\bar\thetab \ ,
\]
\[
    -\frac1{\norm{\thetab_{t_n}}_2}\int_{J_n}a(t)\g_t\,dt
       \longrightarrow\frac{\lambda(2^{1/L}-1)}{c_1}D^{-1}\bar\thetab \ .
\]
The latter, after suppressing the inactive margins, will supply the desired inclusion.

For every \(s\in J_n\),
\[
    \thetab_s-\left(\frac{s}{t_n}\right)^{1/L}\thetab_{t_n}
       =s^{1/L}\int_{t_n}^s t^{-1/L}
          \left(\mathcal E_t+\mathbf e_t\right)dt.
\]
The lower bound $\norm{\thetab_t}_2 \geq ct^{1/L}$ implies $\frac{s^{1/L}}{\norm{\thetab_{t_n}}_2} \leq \frac{(2t_n)^{1/L}}{ct_n^{1/L}} = \frac{2^{1/L}}{c}$. This, ~\eqref{eq:all_limit_error_rate}, and
Cauchy--Schwarz therefore imply that there exists a constant
\(C>0\) such that
\begin{align*}
    \frac{\norm{\thetab_s-(s/t_n)^{1/L}\thetab_{t_n}}_2}
         {\norm{\thetab_{t_n}}_2}
       &\leq C\int_{J_n}t^{-1/L}
          \left(\norm{\mathcal E_t}_{D^{-1}}
                +\norm{\mathbf e_t}_{D^{-1}}\right)dt\\
       &\leq C\int_{J_n}t^\nu
          \norm{\mathcal E_t}_{D^{-1}}\,\frac{dt}{t}
          +C\int_{t_n}^{2t_n}t^{-1/L-\nu-\delta}\,dt\\
       &\leq C\left(\int_{J_n}t^{2\nu}
          \norm{\mathcal E_t}_{D^{-1}}^2\,\frac{dt}{t}
          \right)^{1/2}
          \left(\int_{J_n}1^2\,\frac{dt}{t}\right)^{1/2}
          +C\int_{t_n}^{2t_n}t^{-1-\delta}\,dt\\
       &\leq C\sqrt{\log2}\left(\int_{J_n}t^{2\nu}
          \norm{\mathcal E_t}_{D^{-1}}^2\,\frac{dt}{t}
          \right)^{1/2}
          +Ct_n^{-\delta}\longrightarrow0.
\end{align*}
Here the integral over \(J_n\) tends to zero by
Lemma~\ref{lemma:adam_fast_finite_energy}.
Thus \(\thetab_t/\norm{\thetab_t}_2\to\bar\thetab\) uniformly on \(J_n\),
and
\[
    \frac{\thetab_{2t_n}-\thetab_{t_n}}{\norm{\thetab_{t_n}}_2}
       \longrightarrow(2^{1/L}-1)\bar\thetab.
\]
Integrating~\eqref{eq:all_limit_scaled_momentum_equation} over \(J_n\), and using
\(-a(t)D\m_t=\dot\thetab_t-\mathbf e_t\), gives
\begin{align*}
    -\frac{c_1}{\norm{\thetab_{t_n}}_2}\int_{J_n}a(t)D\g_t\,dt
       &=\lambda\frac{\thetab_{2t_n}-\thetab_{t_n}}
                            {\norm{\thetab_{t_n}}_2}
          -\frac{[a(t)D\m_t]_{t_n}^{2t_n}}
                 {\norm{\thetab_{t_n}}_2}\\
       &\quad+\frac{\alpha}{\norm{\thetab_{t_n}}_2}
                    \int_{J_n}\frac{a(t)D\m_t}{t}\,dt
          -\frac{\lambda}{\norm{\thetab_{t_n}}_2}
                    \int_{J_n}\mathbf e_t\,dt.
\end{align*}
The momentum and parameter bounds make the boundary term and the
integral containing \(\alpha\) of order \(\bigO{t_n^{-1}}\).
The error bound makes the last term \(\bigO{t_n^{-\delta}}\).
Therefore
\begin{equation}\label{eq:all_limit_averaged_force}
    -\frac1{\norm{\thetab_{t_n}}_2}\int_{J_n}a(t)\g_t\,dt
       \longrightarrow\frac{\lambda(2^{1/L}-1)}{c_1}D^{-1}\bar\thetab.
\end{equation}

Next we show that we can focus only on the active margins. Homogeneity and \eqref{eq:all_limit_gradient_decomposition} give
\[
    \frac{\h_i(t)}{\norm{\thetab_t}_2^{L-1}}
       \in\partial q_i\!\left(\frac{\thetab_t}{\norm{\thetab_t}_2}\right),
    \qquad
    -a(t)\g_t=\sum_{i=1}^m b_i(t)
                    \frac{\h_i(t)}{\norm{\thetab_t}_2^{L-1}},
\]
where
\[
    b_i(t):=a(t)\norm{\thetab_t}_2^{L-1}\varphi'(q_i(\thetab_t))\ell(q_i(\thetab_t))\geq0.
\]
The upper bound \(\varphi' \leq \Phi'_M\), the bounds in
\eqref{eq:all_limit_scale_bounds} and $1 - \nu = \frac 1L$ imply that there exist constants \(B,C>0\) with
\begin{equation}\label{eq:all_limit_coefficient_bound}
\begin{aligned}
    \sum_i b_i(t)
       &\leq\Phi_M'a(t)\norm{\thetab_t}_2^{L-1}\Lcal(\thetab_t)
       \leq Ct^{-\nu}, \\
    \frac1{\norm{\thetab_{t_n}}_2}
       \int_{J_n}\sum_i b_i(t)\,dt&\leq \frac{C}{c(1-\nu)}t_n^{-1/L}\left((2t_n)^{1-\nu}-t_n^{1-\nu}\right)\leq B.
\end{aligned}
\end{equation}   

For any \(i\notin I_*\), fix \(j\in I_*\). Uniform convergence
of the directions on \(J_n\) and continuity of the margins imply that
there exists \(d_i>0\) such that, for all large \(n\),
\[
    q_i\!\left(\frac{\thetab_t}{\norm{\thetab_t}_2}\right)
       -q_j\!\left(\frac{\thetab_t}{\norm{\thetab_t}_2}\right)
       \geq d_i>0\qquad(t\in J_n).
\]
Since \(q_i(\thetab_t)=\norm{\thetab_t}_2^L
q_i(\thetab_t/\norm{\thetab_t}_2)\) and
\(\varphi'\) has a positive lower bound on the attained positive
margin range, there exists a constant \(c>0\) such that, for some \(C>0\),
\[
    \frac{b_i(t)}{b_j(t)}
       =\frac{\varphi'(q_i(\thetab_t))}
                {\varphi'(q_j(\thetab_t))}
          e^{-\varphi(q_i(\thetab_t))+\varphi(q_j(\thetab_t))}
       \leq Ce^{-cd_i t}\qquad(t\in J_n).
\]
Together with \eqref{eq:all_limit_coefficient_bound}, this gives
\[
    \frac1{\norm{\thetab_{t_n}}_2}
       \int_{J_n}\sum_{i\notin I_*}b_i(t)\,dt\longrightarrow0.
\]

For each \(i\in I_*\), let \(\h_i^*(t)\) be the Euclidean projection
of \(\h_i(t)/\norm{\thetab_t}_2^{L-1}\) onto
\(\partial q_i(\bar\thetab)\). Upper semicontinuity of \(\partial q_i\)
and uniform convergence of the directions on \(J_n\) give
\(\varepsilon_n\to0\) such that
\[
    \h_i^*(t)\in\partial q_i(\bar\thetab),\qquad
    \norm{\frac{\h_i(t)}{\norm{\thetab_t}_2^{L-1}}-\h_i^*(t)}_2
       \leq\varepsilon_n\quad(i\in I_*)
\]
for almost every \(t\in J_n\).
Equation~\eqref{eq:all_limit_coefficient_bound} and local boundedness
of the normalized subgradients therefore give, for some \(C>0\),
\begin{align*}
    &\frac1{\norm{\thetab_{t_n}}_2}\int_{J_n}
       \norm{-a(t)\g_t-\sum_{i\in I_*}b_i(t)\h_i^*(t)}_2dt\\
    &\quad\leq B\varepsilon_n
       +\frac{C}{\norm{\thetab_{t_n}}_2}
          \int_{J_n}\sum_{i\notin I_*}b_i(t)\,dt\longrightarrow0.
\end{align*}
Equation~\eqref{eq:all_limit_averaged_force} therefore gives
\[
    \frac1{\norm{\thetab_{t_n}}_2}\int_{J_n}
        \sum_{i\in I_*}b_i(t)\h_i^*(t)\,dt
       \longrightarrow
       \frac{\lambda(2^{1/L}-1)}{c_1}D^{-1}\bar\thetab.
\]

It remains to normalize the coefficients. Define
\[
    a_n:=\frac1{\norm{\thetab_{t_n}}_2}
             \int_{J_n}\sum_{i\in I_*}b_i(t)\,dt.
\]
Euler's identity gives
\(\innerprod{\bar\thetab}{\h_i^*(t)}=L\gamma_*\)
for every active index. Taking the inner product of the preceding
limit with \(\bar\thetab\), we obtain
\[
    L\gamma_*a_n
       =\innerprod{\bar\thetab}
          {\frac1{\norm{\thetab_{t_n}}_2}\int_{J_n}
             \sum_{i\in I_*}b_i(t)\h_i^*(t)\,dt}
       \longrightarrow\frac{\lambda(2^{1/L}-1)\kappa_*}{c_1}.
\]
Thus
\[
    a_n\longrightarrow
       \frac{\lambda(2^{1/L}-1)\kappa_*}{c_1L\gamma_*}>0.
\]
For all sufficiently large \(n\), the normalized integral
\[
    \frac1{\norm{\thetab_{t_n}}_2a_n}\int_{J_n}
        \sum_{i\in I_*}b_i(t)\h_i^*(t)\,dt
\]
is a weighted average of elements of \(K_*\), with nonnegative
weights of total mass one, and hence belongs to \(K_*\).
Since \(K_*\) is closed, taking the limit yields
\[
    \frac{L\gamma_*}{\kappa_*}D^{-1}\bar\thetab\in K_*.
\]
This proves \eqref{eq:all_limit_cluster_membership}.
\end{proof}

The inclusion proved in Lemma~\ref{lemma:adam_fast_cluster_point_stationarity} provides, up to bookkeeping, the KKT conditions.
\begin{proof}[Proof of Theorem~\ref{thm:adam_fast_trajectory_dependent_kkt_appendix}]
    Let $\bar\thetab$ be a limit point of
$\thetab_t/\norm{\thetab_t}_2$.
    By Lemma~\ref{lemma:adam_fast_cluster_point_stationarity},
    \[
        \frac{L\gamma_*}{\kappa_*}D^{-1}\bar\thetab\in \operatorname{conv}
          \bigcup_{i\in I_*}\partial q_i(\bar\thetab), \qquad \gamma_*:=\qmin(\bar\thetab)>0,\qquad
    \kappa_*:=\norm{\bar\thetab}_{D^{-1}}^2>0 \ .
    \]
Convexity of each active Clarke subdifferential gives
\[
    \frac{L\gamma_*}{\kappa_*}D^{-1}\bar\thetab
       =\sum_{i=1}^m\tilde \lambda_i\tilde\h_i,
    \qquad
    \tilde\h_i\in\partial q_i(\bar\thetab),\quad
    \tilde \lambda_i\geq0,\quad\sum_i\tilde \lambda_i=1,
\]
with \(\tilde \lambda_i=0\) for \(i\notin I_*\).
Set
\[
    \hat\thetab:=\gamma_*^{-1/L}\bar\thetab,\qquad
    \hat\h_i:=\gamma_*^{-(L-1)/L}\tilde\h_i
                    \in\partial q_i(\hat\thetab),\qquad
    \lambda_i:=\frac{\kappa_*\tilde \lambda_i}{L\gamma_*^{2/L}}.
\]
The preceding representation and homogeneity give
\[
    D^{-1}\hat\thetab=\sum_{i=1}^m\lambda_i\hat\h_i,
    \qquad
    q_i(\hat\thetab)\geq1,\quad\lambda_i\geq0,\quad
    \lambda_i\bigl(q_i(\hat\thetab)-1\bigr)=0.
\]
These are the Clarke--KKT conditions for
\eqref{eq:preconditioned_kkt_appendix}.

Since \(\gamma_*=\qmin(\bar\thetab)\), we have
\(\hat\thetab=\bar\thetab/\qmin(\bar\thetab)^{1/L}\).
So \(\frac{\bar\thetab}{\qmin(\bar\thetab)^{1/L}}\) is a KKT point as required.
\end{proof}

\section{Adam - Intermediate Regime}
The main theorem of this section in fact also holds for the slow regime under the same assumptions; compared to \citet{gronich2026implicitbiasadammuon}, it assumes gradient directional convergence with no zero coordinate (which implies parameter directional convergence), but does not assume convergence of subgradients (Assumption (T3) in \citet{gronich2026implicitbiasadammuon}).

The key distinction between this regime and the fast regime is the non-freezing of the second moments, which implies also non-freezing of the first moments (Proposition~\ref{prop:no-scalar-freezing}).

\begin{theorem}[Intermediate-and-slow regime $\ell_\infty$ implicit bias]
\label{thm:adam_intermediate_kkt_appendix}
Let $L\geq1$ and $-1 < \alpha \leq \frac 1L - \frac 12$. Let $f$ satisfy~\ref{model_ass:1_locally_lipschitz},~\ref{model_ass:2_homogeneous}, and let $\ell \in \mathrm L_{\e}$.
Assume that $\thetab_t:[0,\infty)\to\bbR^p$ is an Adam trajectory satisfying assumptions~\ref{traj_ass:positive_margin_interpolation},~\ref{traj_ass:loss_decay}, and the assumption
    \begin{equation}
        \frac{\g_t}{\norm{\g_t}_1}\longrightarrow\bar\g,
        \qquad
        \norm{\bar\g}_1=1,
        \qquad
        \bar \g_j\neq0\quad\text{for every }j\in[p].
        \label{eq:gradient_direction}
    \end{equation}
Then
    \begin{equation}
        \frac{\thetab_t}{\norm{\thetab_t}_\infty}
        \longrightarrow
        \bar\thetab:=-\sign{\bar\g}
        \in\{-1,1\}^p;
        \label{eq:parameter_direction}
    \end{equation}
And $\hat \thetab := \frac{\bar \thetab}{\qmin^{1/L}(\bar \thetab)}$ is a KKT point of the $\ell_\infty$ max-margin problem.

Thus, by Definition~\ref{def:implicit_regularizer}, $R(\thetab)=\frac12\norm{\thetab}_\infty^2$ is an implicit regularizer of Adam when $-1 < \alpha \leq \frac 1L - \frac12$ for any model $f$ satisfying~\ref{model_ass:1_locally_lipschitz},~\ref{model_ass:2_homogeneous} and with any $\ell \in \mathrm L_{\e}$, under trajectory assumption \eqref{eq:gradient_direction}. Moreover, under~\ref{traj_ass:positive_margin_interpolation},~\ref{traj_ass:loss_decay} and for any $j \in[p]$ with $\frac{\abs{\g_t[j]}}{\norm{\g_t}_1} \geq \Omega(1)$, it holds that
\begin{equation}\label{eq:intermediate_adam_ratio_bound_appendix}
    \frac{\abs{\hat \m_t[j]}}{\sqrt{\vb_t[j]}} = \bigO{t^{-\zeta}} \ .
\end{equation}
\end{theorem}
\begin{proof}[Proof of Theorem~\ref{thm:adam_intermediate_kkt_appendix}]
    Denote for $j \in [p]$,
    \[
    \lim_{t \to \infty}\frac{\g_t[j]}{\norm{\g_t}_1}
    = \bar \g[j] \neq 0 \ .
    \]
    Noting that $\hat \m_t[j] = \emahat{\g_t[j]}{c_1}
    = \emahat{\frac{\g_t[j]}{\norm{\g_t}_1}\norm{\g_t}_1}{c_1}$
    (and similarly for $\hat \vb_t[j]$),
    Proposition~\ref{prop:no-scalar-freezing} and
    Lemma~\ref{lemma:weighted_ema} imply
    \[
    \frac{\hat \m_t[j]}{\emahat{\norm{\g_t}_1}{c_1}}
    \toinfty{t} \bar \g[j],
    \qquad
    \frac{\hat \vb_t[j]}{\emahat{\norm{\g_t}_1^2}{c_2}}
    \toinfty{t} \bar \g[j]^2 \ .
    \]
    Thus,
    \begin{equation}\label{eq:adam_ratio_sign_with_envelope}
    \frac{\hat \m_t[j]}{\sqrt{\hat \vb_t[j]}}
    = \adam{\norm{\g_t}_1}{c_1}{c_2}
      \sign{\bar \g[j]}\left(1+\smallo{1}\right) \ .
    \end{equation}
    Denote $\nu(t) := \eta(t)\adam{\norm{\g_t}_1}{c_1}{c_2}$.
    We first prove
    \[
        \int_0^\infty \nu(s)\,ds = \infty \ .
    \]
    Note that $\norm{\dtheta{t}}_\infty \leq 2\nu(t)$ for large
    enough $t$, so if towards a contradiction
    $\int_0^\infty \nu < \infty$ then
    $\int_0^\infty \norm{\dtheta{t}}\,dt < \infty$,
    implying the trajectory converges to a finite point,
    contradicting $\Lcal \to 0$.

    Taking $T$ so that for any $j \in [p], t \geq T$
    Equation~\eqref{eq:adam_ratio_sign_with_envelope} holds up to
    a residual $e_j(t)$ with
    $\sup_{j,t\geq T}\abs{e_j(t)}<\frac12,
    e_j(t)\toinfty{t}0$, we have
    \[
    \thetab_t[j]
    = \thetab_T[j]
      - \sign{\bar \g[j]}
        \int_T^t \nu(s)\left(1+e_j(s)\right)\,ds \ .
    \]
    By Lemma~\ref{lemma:weighted-cesaro} we get
    \[
    \frac{\thetab_t[j]}{\int_T^t \nu(s)\,ds}
    \toinfty{t} -\sign{\bar \g[j]} \ .
    \]
    Since $\int_0^\infty \nu(s)\,ds = \infty$, the contribution
    on $[0,T]$ is negligible, proving
    \[
    \frac{\norm{\thetab_t}_\infty}{\int_0^t \nu(s)\,ds}
    \toinfty{t} 1,
    \qquad
    \frac{\thetab_t}{\norm{\thetab_t}_\infty}
    \toinfty{t} -\sign{\bar \g} \ .
    \]
    Consequently,
    \[
    \angles{
        \frac{\thetab_t}{\norm{\thetab_t}_\infty}
    }{
        -\frac{\g_t}{\norm{\g_t}_1}
    }
    \toinfty{t}
    \angles{\sign{\bar\g}}{\bar\g}
    = \norm{\bar\g}_1 = 1 \ .
    \]
    Assumption~\ref{traj_ass:positive_margin_interpolation}
    gives $\qmin(\bar\thetab)>0$. Together with
    Assumption~\ref{traj_ass:loss_decay} and the preceding
    alignment, this proves the KKT claim by the argument of
    Theorem C.8 in \citet{gronich2026implicitbiasadammuon},
    with $\norm{\cdot}=\norm{\cdot}_\infty$.
    Although that theorem is stated under Whitney
    $C^1$-stratifiability, its proof applies under our assumptions:
    the loss-subgradient decomposition, homogeneous subgradient
    identities, local boundedness, and closed-graph property of
    the Clarke subdifferentials used there all hold here.

    Equation~\eqref{eq:intermediate_adam_ratio_bound_appendix}
    is exactly the result of
    Corollary~\ref{cor:tracking_and_speed_estimates}
    when $\zeta \geq 0$; when $\zeta < 0$,
    Equation~\eqref{eq:intermediate_adam_ratio_bound_appendix}
    follows from the universal Adam
    bound~\ref{lemma:adam_universal_ratio}.
\end{proof}

\section{Experimental Details}\label{sec:experiment_details}
\subsection{Experimental Validation of Width-$1$ Construction}
The parameter directions appearing in Figure~\ref{fig:adam-sphere} are obtained by training the width-$1$ model~\eqref{eq:construction_model_main_text} with $L=2$ on the dataset~\eqref{eq:construction_dataset_main_text}, with $B=2$, $\Lambda=1$, and $M=35$. We use initialization~\eqref{eq:initialization} with $\rho=1.01>\rho_0(35,2)=1$. We use 142 pairs $(z,\pi_1)$ with $z\in[0,9.528876]\subset[0,3\log35]$ and $\pi_1\in[0.091242,14.412188]$, and $\pi_0=1/\pi_1$. These were selected using preliminary experiments to obtain broad, well-spaced coverage of the attainable parameter directions. Exchanging the two input coordinates and $u_1(0),u_2(0)$ gives the corresponding trajectories on the symmetric dataset, yielding 284 directions in total.

We minimize the summed exponential loss $\mathcal{L}(\thetab)=\sum_{i=1}^{2}\exp(-y_i f(\mathbf{x}_i;\thetab))$ using full-batch, bias-corrected Adam with $\beta_1=0.9$, $\beta_2=0.999$, $\varepsilon=0$, zero initial moments, and learning rate $\eta_k=10^{-3}k^{0.15}$ for update $k\geq1$, that is $\alpha = 0.15 > \frac1L - \frac12 = 0$. All trajectories run for 600,000 updates in float64 arithmetic. Trajectory length was chosen to ensure stability of the parameter direction: each trajectory's parameter direction is stable up to an error of $0.012$ throughout the entire $100,000$ last steps, that is:
\[
    \max_{500,000\leq s,t \leq 600,000} \norm{\frac{\thetab_s}{\norm{\thetab_s}_2}-\frac{\thetab_t}{\norm{\thetab_t}_2}}_2 < 0.012 \ .
\]

\subsection{Margin Comparisons on MNIST}
We conduct two experiments on binary classification of MNIST: one long-horizon experiment with $\varepsilon = 0$ only, using a bias-free network and small initialization scale, and a shorter experiment comparing $\varepsilon = 0$ and $\varepsilon=10^{-8}$ on a more practical setting: a network with biases and standard initialization.

For both experiments we use 1,024 images from the MNIST training set: the first 512 even-digit and first 512 odd-digit examples in the original dataset order, labeled \(-1\) and \(+1\), respectively. Pixels are divided by \(255\), centered by the per-pixel mean of these selected examples, and flattened. The network is a multilayer perceptron with widths \(784 \to 512 \to 512 \to 1\), with ReLU activations after the two hidden layers. For the $\varepsilon$ comparison experiment only, we incorporate biases in every layer for ease of training (even though such networks are not precisely homogeneous, they are nearly homogeneous, see~\citet{cai2025_nearlyhomogeneous}). We minimize the full-batch mean logistic loss, \(\mathcal{L}(\thetab)=1024^{-1}\sum_i\log(1+\exp(-y_i f(\x_i;\thetab)))\), using float64 arithmetic. Each layer’s weights and biases are initialized independently from \(\mathcal{U}(-1/\sqrt{n_{\mathrm{in}}},\,1/\sqrt{n_{\mathrm{in}}})\) and then multiplied by a common factor of $10^{-4}$ in the long-horizon $\varepsilon = 0$ experiment (a small initialization is known to accelerate arrival to the asymptotic regime \citep{moroshko2020implicit,woodworth2020kernelrichregimesoverparametrized}). In both experiments, when running Adam with \(\varepsilon=0\), we assign zero updates to coordinates whose gradients have remained identically zero. 

In all runs we use \(\beta_1=0.9\) and \(\beta_2=0.999\) (PyTorch \citep{paszke2019pytorch} defaults). At update \(k\geq1\), the learning rate is \(\eta_k=3\times10^{-4}k^\alpha\), where \(\alpha\in\{0,-0.1,-0.3,-0.5,-0.7\}\). We repeat each schedule with five initialization seeds, \(2026100901,\ldots,2026100905\), using identical initial weights across schedules for each seed. For the long-horizon $\varepsilon=0$ experiment, every run terminates at the first iterate satisfying \(\mathcal{L}\leq10^{-20}\), and for the comparison experiment, every run terminates at the first iterate satisfying \(\mathcal{L}\leq10^{-10}\) or at 1 million steps (the $\alpha=-0.7,-0.5$ trajectories finish at a slightly larger loss than $10^{-10}$). Experiments use Python 3.12.14 and PyTorch 2.13.0+cu130 on NVIDIA GPUs.

Below, we present supplementary figures to the main text.

\par\vspace{0.5em}
\textbf{Long-horizon, small-initialization Adam with no stability constant ($\varepsilon = 0$).}

\begin{figure}[!htb]
  \centering
  \includegraphics[width=0.5\columnwidth]{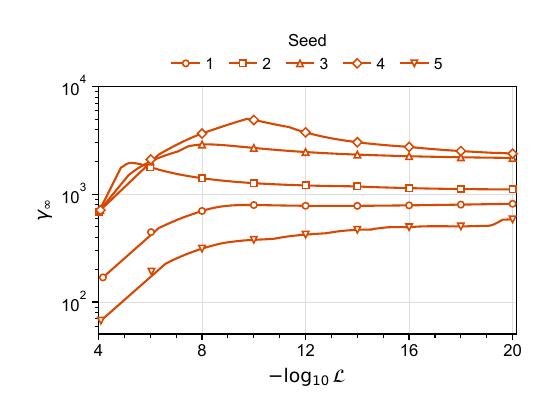}
  \caption{\textbf{$\ell_\infty$ margins for $\alpha=-0.3$, for five seeds.} }
  \label{fig:appendix-alpha-m03-seeds}
\end{figure}

In Figure~\ref{fig:appendix-alpha-m03-seeds} we focus on the $\ell_\infty$ margin for $\alpha=-0.3$ by seed. This schedule is contained in the intermediate regime and hence we expect $\ell_\infty$ margin maximization, but this bias could be expressed late in the trajectory. In the experiment, some seeds show monotone increasing margin, others have a slight decrease and recovery of the margin, and yet others show no recovery  during the trajectory after achieving a peak value. 

\begin{figure}[p]
  \centering
  \includegraphics[width=\linewidth,height=0.86\textheight,keepaspectratio]{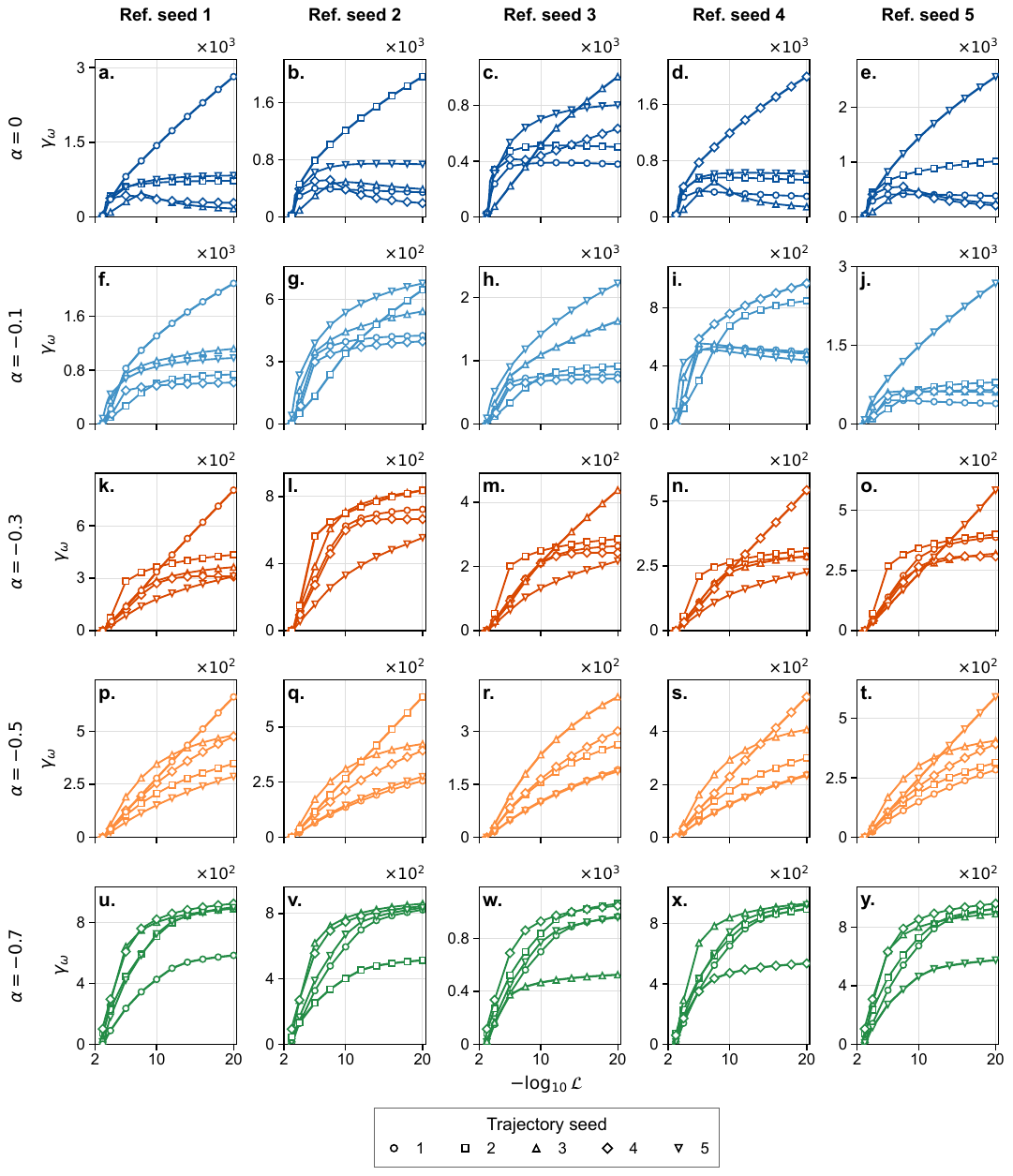}
  \caption{\textbf{Endpoint-weighted margins for $\varepsilon=0$ in the long-horizon experiment.} Rows correspond to learning-rate schedules $\eta_t\propto t^\alpha$; columns identify the reference seed whose endpoint metric is held fixed. Each panel shows all five trajectories under that schedule.}
  \label{fig:endpoint-grid-epsilon0}
\end{figure}

In Figure~\ref{fig:endpoint-grid-epsilon0} we show the weighted-$\ell_2$ margin of each seed and schedule, evaluated on all seeds. The fast regime $\alpha=0$ shows the strongest trajectory-dependent weighted-$\ell_2$ margin increase, with this property becoming less pronounced as the schedule becomes slower. 

\par\vspace{0.5em}
\textbf{Comparison of $\varepsilon=0,10^{-8}$ in a practical setting.}

For $\varepsilon=10^{-8}$, to identify the onset of the effect of $\varepsilon$, we measure the relative error of the update with and without $\varepsilon$,
\[
e_{\varepsilon}(t) =  \frac{\norm{\frac{\hat \m_t}{\sqrt{\hat \vb_t}} - \frac{\hat \m_t}{\sqrt{\hat \vb_t}+\varepsilon}}_2}{\norm{\frac{\hat \m_t}{\sqrt{\hat \vb_t}}}_2} \ ,
\]
which tends to 1 when $t\longrightarrow \infty$, since for every $j \in [p]$ for which $\vb_t[j] > 0$ eventually, it holds that $\vb_t[j] \toinfty{t} 0$, and so
\[
    \frac{\hat \m_t[j]}{\sqrt{\hat \vb_t[j]}+\varepsilon} \sim \varepsilon^{-1}\hat \m_t[j] = \smallo{\frac{\hat \m_t[j]}{\sqrt{\hat \vb_t[j]}}} \ .
\]

\begin{figure}[!htb]
  \centering
  \includegraphics[width=0.65\linewidth]{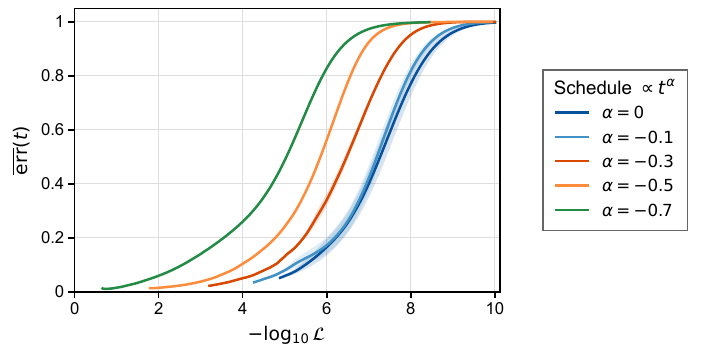}
  \caption{
    \textbf{Relative update error for Adam with $\varepsilon=10^{-8}$.}
    The quantity $\overline{e_{\varepsilon}}(t)$ denotes the trailing
    1,000-update moving average of $e_{\varepsilon}(t)$.
    Curves show means across five seeds; shading denotes one
    standard error.}
  \label{fig:epsilon-relative-error}
\end{figure}

To avoid momentary effects, we observe in Figure~\ref{fig:epsilon-relative-error} a moving average 
\[
\overline{e_{\varepsilon}}(t) = \frac{1}{1000}\sum_{i=0}^{999}e_{\varepsilon}(t - i), \qquad t \geq 1,000  \ ,
\]
for all five schedules in the $\varepsilon=10^{-8}$ experiment. Note that some curves start at an advanced loss value at update 1,000. As expected, the curves begin near $0$ and tend to $1$ towards the end of training.

\begin{figure}[p]
  \centering
  \includegraphics[width=\linewidth,height=0.86\textheight,keepaspectratio]{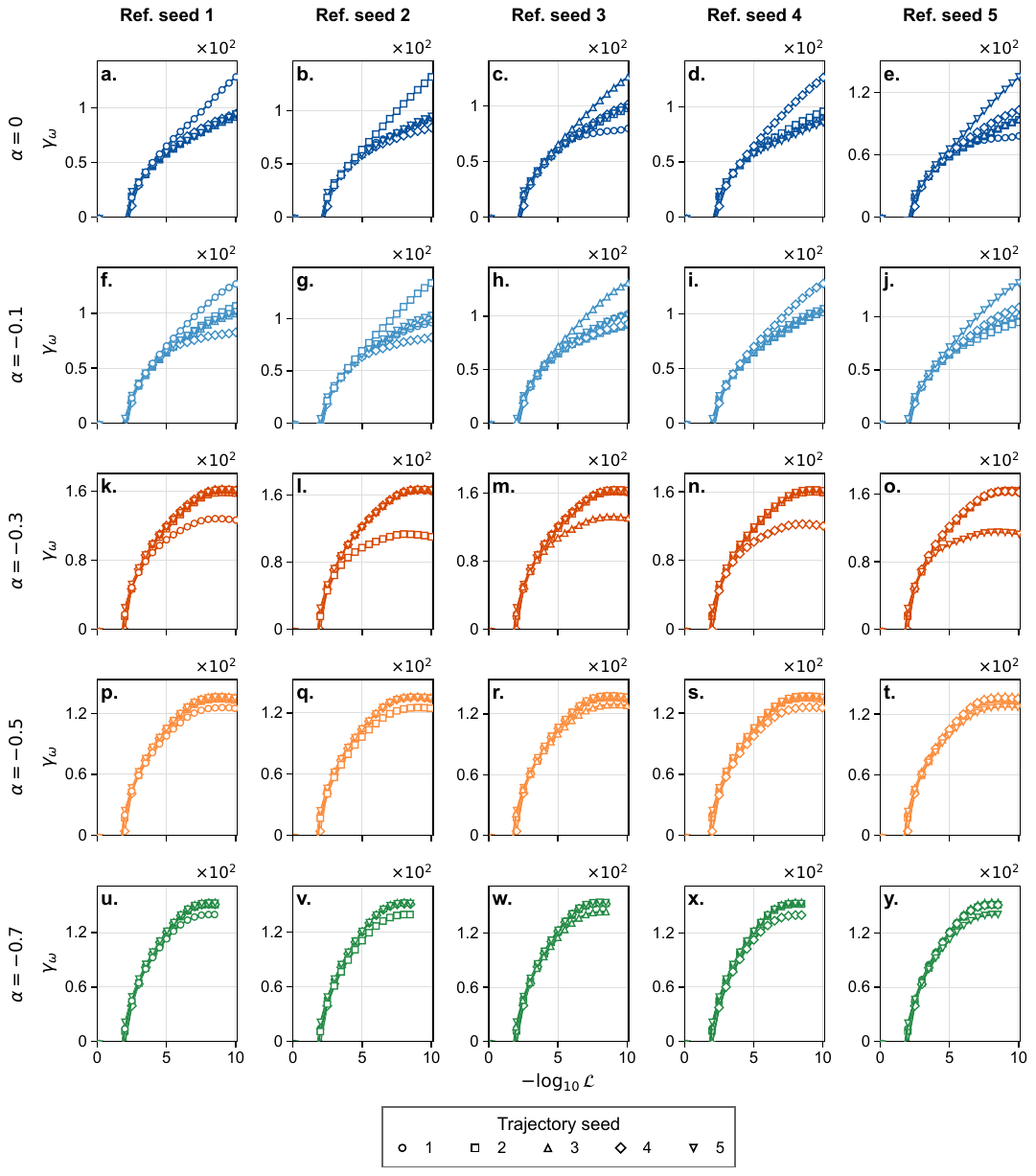}
  \caption{\textbf{Endpoint-weighted margins for $\varepsilon=10^{-8}$ in the comparison experiment.} Rows correspond to learning-rate schedules $\eta_t\propto t^\alpha$; columns identify the reference seed whose endpoint metric is held fixed. Each panel shows all five trajectories under that schedule.}
  \label{fig:endpoint-grid-comparison-epsilon1e8}
\end{figure}

\begin{figure}[p]
  \centering
  \includegraphics[width=\linewidth,height=0.86\textheight,keepaspectratio]{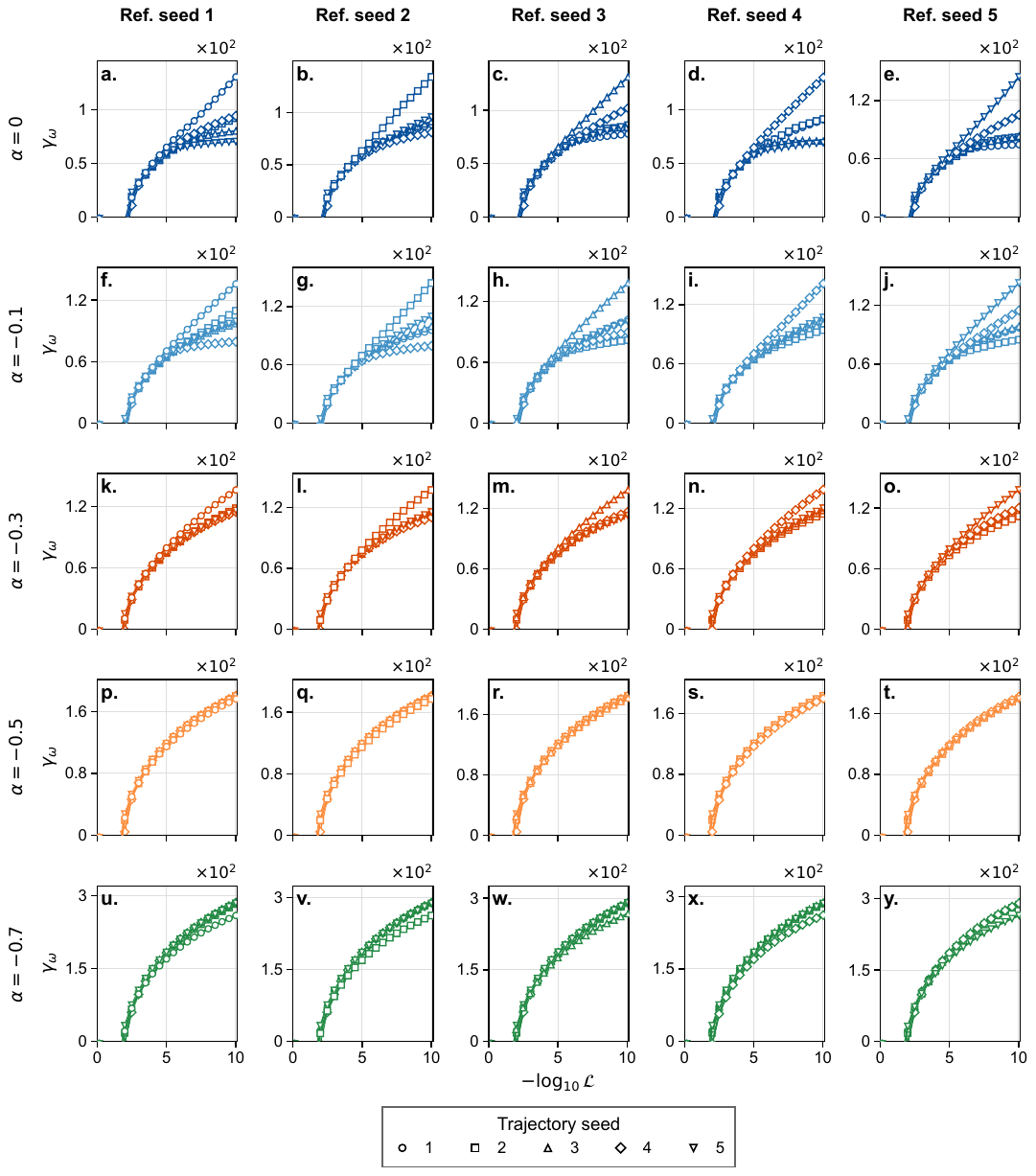}
  \caption{\textbf{Endpoint-weighted margins for $\varepsilon=0$ in the comparison experiment.} Rows correspond to learning-rate schedules $\eta_t\propto t^\alpha$; columns identify the reference seed whose endpoint metric is held fixed. Each panel shows all five trajectories under that schedule.}
  \label{fig:endpoint-grid-comparison-epsilon0}
\end{figure}

In Figures~\ref{fig:endpoint-grid-comparison-epsilon1e8},~\ref{fig:endpoint-grid-comparison-epsilon0} we show the weighted-$\ell_2$ margin of each seed and schedule, evaluated on all seeds, for $\varepsilon=10^{-8}$ and $\varepsilon=0$ respectively. The fast regime $\alpha=0$ shows the strongest trajectory-dependent weighted-$\ell_2$ margin increase, with this property becoming less pronounced as the schedule becomes slower. 

\end{document}